\documentclass[nonacm,acmsmall,screen]{acmart}

\AtBeginDocument{%
  }

\setcopyright{acmlicensed}
\copyrightyear{2018}
\acmYear{2018}
\acmDOI{XXXXXXX.XXXXXXX}

\acmConference[Conference acronym 'XX]{Make sure to enter the correct
  conference title from your rights confirmation emai}{June 03--05,
  2018}{Woodstock, NY}
\acmISBN{978-1-4503-XXXX-X/18/06}

\usepackage{enumitem}
\usepackage{comment}
\usepackage[utf8]{inputenc} % allow utf-8 input
\usepackage{hyperref}       % hyperlinks
\usepackage{url}            % simple URL typesetting
\usepackage{booktabs}       % professional-quality tables
\usepackage{amsfonts}       % blackboard math symbols
\usepackage{nicefrac}       % compact symbols for 1/2, etc.
\usepackage{microtype}      % microtypography
\usepackage{xcolor}         % colors
\usepackage{amsmath}

\usepackage{xcolor}         % colors

\usepackage{graphicx}
\usepackage{subcaption}
\usepackage{algorithm}
\usepackage{algorithmic}

\theoremstyle{plain}
\newtheorem{theorem}{Theorem}[section]
\newtheorem{proposition}[theorem]{Proposition}
\newtheorem{lemma}[theorem]{Lemma}

\theoremstyle{definition}

\newtheorem{assumption}[theorem]{Assumption}

\theoremstyle{remark}
\newtheorem{remark}[theorem]{\bf Remark}

\newcommand{\abs}[1]{\left|#1\right|}
\newcommand{\norm}[1]{\lVert#1\rVert}

\newcommand\numberthis{\addtocounter{equation}{1}\tag{\theequation}}
\newcommand{\indicator}{\mathbf{1}}

\begin{document}

\title{Spectral Clustering for Crowdsourcing with Inherently Distinct Task Types}

\author{Saptarshi Mandal}
\authornote{Both authors contributed equally to this research.}
\email{smandal4@illinois.edu}
\author{Seo Taek Kong}
\authornotemark[1]
\email{skong10@illinois.edu}
\author{Dimitrios Katselis}
\email{katselis@illinois.edu}
\author{R. Srikant}
\email{rsrikant@illinois.edu}
\affiliation{%
  \institution{University of Illinois, Urbana-Champaign}
  %\city{}
  %\state{}
  \country{USA}
}


\begin{abstract}
 The Dawid-Skene model is the most widely assumed model in the analysis of crowdsourcing algorithms that estimate ground-truth labels from noisy worker responses.
In this work, we are motivated by crowdsourcing applications where workers have distinct skill sets and their accuracy additionally depends on a task's type.
While weighted majority vote (WMV) with a single weight vector for each worker achieves the optimal label estimation error in the Dawid-Skene model, we show that different weights for different types are necessary for a multi-type model. Focusing on the case where there are two types of tasks, we propose a spectral method to partition tasks into two groups that cluster tasks by type.
Our analysis reveals that task types can be perfectly recovered if the number of workers $n$ scales logarithmically with the number of tasks $d$.
Any algorithm designed for the Dawid-Skene model can then be applied independently to each type to infer the labels.
Numerical experiments show how clustering tasks by type before estimating ground-truth labels enhances the performance of crowdsourcing algorithms in practical applications.
% when applying crowdsourcing
% We also prove that when task type information is not available and the clustering step is sufficiently accurate, our algorithm outperforms type-agnostic algorithms based on weighted-majority vote for accurately clustered tasks.
% We show through experiments that clustering followed by label estimation for each type to infer the truth values outperforms task-agnostic algorithms...
\end{abstract}




\begin{comment}

\begin{teaserfigure}
  \includegraphics[width=\textwidth]{sampleteaser}
  \caption{Seattle Mariners at Spring Training, 2010.}
  \Description{Enjoying the baseball game from the third-base
  seats. Ichiro Suzuki preparing to bat.}
  \label{fig:teaser}
\end{teaserfigure}

\received{20 February 2007}
\received[revised]{12 March 2009}
\received[accepted]{5 June 2009}
\end{comment}
\maketitle

\section{Introduction}\label{introduction}
Labeled datasets are required in many machine learning applications to either train classifiers using supervised learning or to evaluate their performance. 
Crowdsourcing is a popular way to label large datasets by collecting labels from a large number of workers at a low cost. The collected labels are often noisy due to many reasons including the difficulty of some labeling tasks, differing skill sets of the workers, etc.(\citet{BonaldCombes, Gao16}). The crowdsourced labels are then used to infer ground-truth labels by aggregating the responses of the workers. 
To analyze the quality of the inferred labels, a statistical model for the workers' responses is often assumed.

% Because the truth values of crowdsourced tasks are never observed, a common alternative to the majority vote 
% While weighing the responses of more reliable workers is ideal, it is often ...

A widely-studied model for crowdsourcing was first proposed by \citet{DawidSkene79}.
Their one-coin model assumes that workers have distinct skill sets, and each worker submits responses to a task independently of all other tasks and workers.
Formally, each worker $i$ is assumed to submit a response $X_{ij}$ to a task $j$ that correctly reflects the label $y_j$ with an unknown but fixed probability $p_i$.
Although the true labels are never observed, it is possible to estimate the unknown accuracy parameters $p = (p_1, \dots, p_n)$ by assuming that workers respond according to this statistical model.
Once the accuracy parameters are estimated, labels can be estimated using the Nitzan-Paroush estimate(\citet{NitzanParoush}).
% The truth values can then be estimated using the Nitzan-Paroush estimate \citet{NitzanParoush} or with expectation maximization \citet{ProjectedEM}.
% The estimated accuracy parameters are then used to infer the truth values as the Nitzan-Paroush estimate \citet{NitzanParoush} or by iteratively applying the expectation maximization algorithm \citet{ProjectedEM}.
Despite the simplicity of this Dawid-Skene model, the optimal error rates of label estimation algorithms have only been understood relatively recently \citet{berend2014consistency,Gao16}.

% {\color{red}Start here. March 10}
In this paper, we are interested in modeling worker responses when crowdsourced tasks demand different levels of expertise.
The considered model is motivated by expert behavior in radiology when labeling the presence of thoracic nodules can be more difficult because of their shape and size, or when they are imaged with different resolutions, resulting in labels that are more reliable for tasks with one type than the other \citet{JSRT,ResolutionRadiology}. The contributions of the paper are the following:
\begin{enumerate}[leftmargin=*]
    \item We consider a model for crowdsourcing that describes settings when workers label tasks that require different levels of expertise. To motivate the need for clustering tasks into different types, we first examine a scenario where the task types are known. For this model, we demonstrate that the type-agnostic weighted majority vote (WMV) algorithm, designed for Dawid-Skene models, performs worse than a WMV algorithm designed for each task type separately.
    \item Next, we consider our main model where the task types are unknown. For this model, we design a spectral clustering algorithm to cluster the tasks into two types. Our main contribution is a result that the clustering algorithm correctly classifies all tasks if the number of workers is of the order of the log of the number of tasks. Specifically, we first show that the observation matrix whose spectral properties are used for clustering has a special structure, namely, a low-rank part plus a perturbation. We show that the perturbation is small and then adapt the ideas from \citet{fan2018eigenvector} to show that perfect clustering is possible under conditions that are natural in crowdsourcing applications. The most common perturbation result used in the clustering literature is the Davis-Kahan theorem \cite{DavisKahan}, which characterizes the perturbation in eigenvectors as a function of the perturbation in a matrix. However, the Davis-Kahan theorem does not exploit any special structure of the matrix that is being perturbed while the result in \cite{fan2018eigenvector} allows us to exploit a low-rank structure that we have identified in the crowdsourcing observation matrix.
    \item Once the tasks are clustered into two different types, any crowdsourcing algorithm designed for the DS model can then be applied to estimate labels corresponding to tasks of each type independently of the other.
For concreteness, we focus on estimating the workers' reliabilities using the Triangular Estimation (TE) algorithm proposed by \citet{BonaldCombes}, followed by the Nitzan-Paroush (NP) decision rule(\citet{NitzanParoush}) to estimate the labels. For this algorithm, we obtain an upper bound on the probability of task labeling error, and using this bound, we show that the probability of labeling error goes to zero exponentially fast in the number of workers. Further, we show that for the type-agnostic algorithm, this is not the case, i.e., there are problem parameters for which the asymptotic probability of labeling error does not go to zero exponentially fast. 
\item Finally, we conduct experiments using publicly available datasets. We compared two classes of algorithms: one where we first performed task clustering by type and then applied an algorithm designed for the traditional DS model to label tasks separately for each type and the other where the labeling algorithm is directly applied to the dataset without any clustering. Our experimental results show that clustering followed by labeling outperforms direct labeling in most of the datasets we considered except in one dataset. We also compared our algorithm with other algorithms which also divide tasks into types. Again, we found that our algorithm outperforms other task type-dependent algorithms except in the case of one dataset. We also explain why we believe this to be the case.
\end{enumerate}

\section{Background}\label{sec:RelatedWork}
\subsection{Problem Setting}
Throughout this paper, the notation $\log$ refers to the natural logarithm.
For any positive integer $m$, denote by $[m]$ the set $\{1, \dots, m\}$.
Let $n \geq 3$ be the number of workers labeling $d > 4$ tasks.
Each task $j \in [d]$ is associated with deterministic but unknown ground-truth labels $y_1, y_2, \dots, y_d \in \{-1, +1\}$ following \citet{Gao16}.
Each worker $i \in [n]$ independently submits a response $X_{ij} \in \{-1, +1\}$ to each task $j$ with $X_{ij}$ being independent across task index $j$. The goal is to estimate the true label $y_j \in \{-1, +1\}$ for every task $j \in [d]$.
%When ground-truth labels are considered random variables \citet{BonaldCombes}, the worker responses are assumed to be conditionally independent given the true labels.

In our model, each task $j$ is further associated with a type $k_j \in \{e, h\}$ indicating ``easy'' and ``hard'' types, respectively.
The task types are also deterministic but unknown, and a task's type $k_j$ determines the accuracy parameter $p_{k_j i} = \mathbb{P}(X_{ij} = y_j) $ as the probability of worker $i$ correctly labeling a task $j$ for all workers $i \in [n]$. Using the accuracy vectors, we can define the reliability vectors $r_e, r_h \in [-1, 1]^n$ as $r_k = 2p_k-1,$ where we denote the $i^{\rm th}$ element of $p_k$ by $p_{ki}$ for all $k \in \{e,h\}.$
Finally, we let the number of tasks of type $k$ be $d_k$; clearly, $d_e + d_h = d$. We assume that $d_k$ is unknown. We consider the case when all workers label all tasks.

This \emph{hard-easy model} is motivated by applications where certain tasks can inherently be more difficult than others. 
We characterize the collective potential of the crowd of workers for type $k$ by the Euclidean norm of the corresponding reliability vectors $\norm{r_k}$, where $r_k = (r_{k1}, \dots, r_{kn})$.
In keeping with the motivation of studying problems with hard and easy tasks, we assume the following:
\begin{assumption}\label{assumption:easyhard}
    The reliability vectors satisfy
    \begin{equation}
        \norm{r_e} > \norm{r_h} .
    \end{equation}
\end{assumption}
We also assume that the reliabilities are bounded away from $-1$ and $+1$:
\begin{assumption} \label{assumption:rho}
    For some $\rho \in (0, 1/2)$ independent of $(n, d, k)$, the reliability vectors $r_e, r_h$ satisfy
    \begin{equation}
        \rho \leq \frac{1 \pm r_{ki}}{2} \leq 1 - \rho
    \end{equation}
    for all $i \in [n]$ and type $k \in \{e, h\}$.
\end{assumption}

Our hard-easy model can be considered an extension of the one-coin Dawid-Skene (DS) model to two types of tasks. Henceforth, when we refer to the DS model, we mean the one-coin DS model unless explicitly stated otherwise.

\subsection{Related Work: Dawid-Skene Model}\label{sec:DS algo}
Crowdsourcing models differ in the assumed structure for the accuracy matrix $P$, where
\begin{equation}
    P_{ij} = \mathbb{P}\left(X_{ij} = y_j\right) .
\end{equation}
In the one-coin DS model, $P$ is a matrix with $d$ identical columns. There is a vast literature on inferring labels from data under this model. These include the original EM algorithm proposed in \citet{DawidSkene79}, spectral-EM algorithm in \citet{SpectralEM}, message passing algorithm in \citet{karger_sigmetrics,karger_OR}, label estimation from the principal eigenvector of the worker-similarity matrix studied in \citet{Dalvi} to name a few. 

For our experiments, we will use the following DS algorithm: first estimate reliabilities and then use a weighted majority vote algorithm for estimating task tasks. We will review the WMV algorithm first. 
Consider the Dawid-Skene model so that the distribution of the binary worker response matrix $X \in \{-1, +1\}^{n \times d}$ is determined by a single reliability vector $r \in [-1, +1]^n$, i.e. all tasks are of same type.
Given \emph{known} reliabilities $r$ and focusing on a single task with worker responses $x = (x_1, \dots, x_n)$, the maximum likelihood decision rule for a given task $j$ is then given by the map
\begin{equation}\label{eq:wmv}
    g^* (x) = \mathrm{sgn} \left(\sum_{i=1}^n w_i x_i\right) ,
\end{equation}
with (possibly infinite) weights 
\begin{equation}\label{eq:wmv_weights}
    w_i = \log \frac{1 + r_i}{1 - r_i} .
\end{equation}
Based on this observation, a common approach is to estimate the reliability vector $r$ from the responses $X$, denoted as $\hat{r}$, and use the Nitzan-Paroush decision rule(\citet{NitzanParoush}) to infer the labels as
\begin{equation}\label{eq:NitzanParoush}
    \hat{y}_j^{NP} = \mathrm{sgn}\left(
        \sum_{i=1}^n \log \frac{1+ \hat{r}_i}{1 - \hat{r}_i} X_{ij}
    \right), \forall j \in [d] .
\end{equation}
Equation \eqref{eq:NitzanParoush} corresponds to a \emph{weighted majority vote} of the form \eqref{eq:wmv} with weights $w_i = \log \frac{1+\hat{r}_i}{1-\hat{r}_i}$.

Next, we review the TE algorithm for estimating reliabilities proposed in \citet{BonaldCombes}, which we will use in our theoretical results. The reason we focus on this algorithm is that it has been compared to other algorithms and shown to perform better in real datasets. Additionally, by comparing the probability of labeling error expression derived from \citet{BonaldCombes} with the lower bounds in \citet{Gao16}, it can be seen that the algorithm is provably asymptotically optimal. We give a brief description of the TE algorithm for completeness.
The TE algorithm designed for estimating a reliability vector for the DS model first computes the worker-covariance matrix
\begin{equation}
    W_{ab} = \frac{1}{d} \sum_{j = 1}^d X_{aj} X_{bj} , \forall a, b \in [n].
\end{equation}
For every worker $i \in [n]$, the most informative pair of co-workers $\arg\max_{a, b \in [n]: a\neq b \neq i} \abs{W_{ab}}$ denoted by $(a_i, b_i)$ is computed, and the magnitude of the $i$th worker's reliability is estimated as
\begin{equation}\label{eq:est_rel_abs}
    \abs{\hat{r}_{i}} 
    = \left\{\begin{array}{cc}\left[\sqrt{\abs{\frac{W_{a_i i} W_{b_i i}}{W_{a_i b_i}}}}\right]_{[2\rho-1,1-2\rho]},& \text{if}\ \  |W_{a_i b_i}| > 0\\
    0, & \text{else}
    \end{array} \right. .
\end{equation}
The sign of $\hat{r}_{i}$ is estimated by letting 
\begin{equation*}
i^{*}=\arg\max_{i \in [n]}\left|\hat{r}_{ i}^2 +\sum_{j\in [n] : j\neq i}W_{ji}\right|.
\end{equation*}
and by setting the sign of $\hat{r}$ according to
\begin{equation*}
    \mathrm{sgn}(\hat{r}_{ i })=\left\{\begin{array}{cc}
    \mathrm{sgn}\left(\hat{r}_{i^{*}}^2 +\sum_{j\in [n]: j\neq i^{*}}W_{ji^{*}}\right), & \text{if}\ \ i=i^{*}\\
    \mathrm{sgn}\left(\hat{r}_{ i^{*}}W_{ii^{*}}\right), & \text{else}
\end{array}\right. .
\end{equation*}
This concludes our discussion of the TE algorithm.

\subsection{Related Work: Task-Specific Reliability Models}

Having reviewed the DS model and associated algorithms in the previous subsection, we note that the key feature of all these models is that a worker has the same reliability for all tasks although different workers may have different reliabilities. The basic DS model has been extended to consider the case where the same worker can have different reliabilities for different tasks, we review these models in the next subsection. 

A rank-1 model studied in \citet{KhetanOh} assumes that $P$ is an outer product of the accuracy of the workers and a vector parametrizing the easiness of all tasks.
A more general model was studied in \citet{OBIWAN}, where $P$ is assumed to satisfy strong stochastic transitivity (\citet{SST}).
In the context of crowdsourcing, this assumption implies that workers can be ranked from most to least accurate, and that this ranking does not change across tasks. The $P$ that they consider can be associated with a rank as large as $\min(n,d).$ 
% Such cases may arise, for example, when tasks require different specialties, and the workers have different areas of expertise.
Lastly, the model in \citet{Graphon,Kaist} assumes an accuracy matrix $P$ that exhibits a low-rank structure with a fixed number of distinct entries. they call it a $k$-type specialization model which is close to a stochastic block model with $k$ communities. In their model, each worker and task can be of $k$ different types with type assignment being independent among tasks and workers. The accuracy parameter $P_{ij}$ with worker $i$ and task $j$ having a matched type is relatively higher than any other mismatched pair. The algorithms designed for this model in \citet{Graphon,Kaist} have a two-step approach. The first step involves clustering workers according to their types. The second step is estimating labels for each task $j$ using a weighted majority vote where significant weight is given to workers that match the type of task $j$ and negligible weight is given to all other workers.   

We now compare our model to the above models. As pointed out in \citet{Kaist}, both \citet{KhetanOh} and \citet{OBIWAN} consider the following: if worker A is better than worker B for any task, then this same ordering holds for all other tasks. Such a monotonicity is not assumed in our model. The $k$-type specialization model in \citet{Graphon,Kaist} somewhat similar in spirit to our model in the sense it attempts to cluster tasks according to types. However, they also cluster workers according to types and their algorithm uses a simple majority vote or a majority vote with two weights. Such a voting scheme is not optimal when different workers have different reliabilities \cite{NitzanParoush}. 

It is worth noting that our model assumes all workers respond to all tasks, as it is motivated by applications where an institution contracts professionals to label a dataset.
This does not model applications that use platforms such as Amazon Mechanical Turk, in which workers independently select a sparse subset of tasks to label. 
Our model can be extended to accommodate such a sparsity in the dataset, where a sparsity parameter would be integrated in the performance bounds. 
But given our motivation, we have chosen not to do so in this paper. 

\section{Main Results}
\subsection{Limitations of Type-Agnostic Weighted Majority Vote}

In our hard-easy model, the response matrix $X$ is drawn from a distribution depending on two reliability vectors $r_e$ and $r_h$.
Suppose these parameters are known, while the task types $k_1, \dots, k_d$ are unknown.
This motivates the study of all algorithms of the form \eqref{eq:wmv}, which we call type agnostic weighted majority vote (TA-WMV), that use a single weight vector $w$ as a function of $r_e$ and $r_h$.
The average error rate for this label estimate $\hat{y}^{WMV}(w)$ is defined as
\begin{equation}\label{eq:wmv_error}
    \mathbb{P}_{av}(w) = \frac{1}{d} \sum_{j=1}^d \mathbb{P}\left(\hat{y}_j^{WMV}(w) \neq y_j\right) ,
\end{equation}
where the dependence on weights $w$ is emphasized. 
For the case of one-coin model DS model, the error rate \eqref{eq:wmv_error} is invariant to the true labels $y_1, \dots, y_d$, and can therefore be equivalently expressed as
% as justified in Appendix \ref{appendix:y_invariance}.
% {\color{red}Move the corresponding lines to appendix}.
% As a consequence of label-invariance, \eqref{eq:wmv_error} can be equivalently expressed as
\begin{equation}
    \mathbb{P}_{av} (w) = \frac{d_e}{d} \mathbb{P}_e \left(\hat{y}^{WMV}(w) \neq y\right) + \frac{d_h}{d} \mathbb{P}_h \left(\hat{y}^{WMV}(w) \neq y\right) ,
\end{equation}
where $\mathbb{P}_k$ is the probability measure parametrized by task type $k \in \{e, h\}$.
The following two propositions characterize the error rate of all TA-WMV algorithms with access to the reliability vectors but use a single weight vector for both task types. The proof of the Proposition \ref{prop:WMV UB} is a straightforward application of Chernoff bound and is given in the Appendix \ref{proof: prop: WMV UB} for completeness.
\begin{proposition}[upper-bound on expected labeling error: TA-WMV]\label{prop:WMV UB} 
    Suppose $X$ is drawn from the hard-easy model, and that the reliability vectors $r_e, r_h$ are known.
    For any weight vector $w = w(r_e, r_h)$, the probability of error on task $j$ of type $k \in \{e, h\}$ satisfies
    \begin{equation}
        \mathbb{P}_k \left(\hat{y}_j^{WMV}(w) \neq y_j \right) \leq \exp \left(-n \varphi_{n}(w, r_k)\right) ,
    \end{equation}
    where the error exponent $\varphi_{n}(w,r_k)$ is given by
    \begin{equation}\label{eq:varphi}
       \varphi_n(w,r_k)= - \inf_{t \geq 0} \frac{1}{n} \sum_{i=1}^n \log \left(
            e^{t w_i} \frac{1 - r_{ki}}{2} + e^{-t w_i} \frac{1+r_{ki}}{2}
        \right) .
    \end{equation}
\end{proposition}
Next, the Proposition \ref{thm:WMV} shows that the error rate is tight up to multiplicative factors:
\begin{proposition}[lower-bound on expected labeling error: TA-WMV]\label{thm:WMV}
    Let $0< w_l \leq w_u < \infty$ be positive constants such that the weights of a TA-WMV algorithm satisfy  $w_l \leq w_i \leq w_u$ for all workers $i$.
    For any $y \in \{-1, +1\}^d$, the average error rate $\mathbb{P}_{av}(w)$ of any TA-WMV with that uses a single weight vector across all tasks for label estimation satisfies
    \begin{equation}
        \liminf_{n \to \infty} 
            \frac{1}{n} \log \min_{w} P_{av} (w) 
        \geq -\limsup_{n \to \infty}\max_{w}\min_{k}\varphi_n(w,r_k),
    \end{equation}
 for any ground-truth vector $y \in \{-1, +1\}^d$.
\end{proposition}
The above result is similar to Theorem 5.1 in \citet{Gao16}; however, our proposition uses weighted majority voting for arbitrary weights for a type $k$, whereas their result is for majority voting. While the techniques are similar, the generalization is important to study the role of having two task types as compared to the single task-type DS model.
The requirement that the norm of the weights $w$ is bounded excludes pathological constructions where, for example, weights are all zeros. The proof of Proposition \ref{thm:WMV} is given in Appendix \ref{appendix:proof thm:WMV }.

To understand the limitation of TA-WMV algorithms, it is instructive to compare the error rates in Proposition \ref{thm:WMV} with the achievable rates by an algorithm that accounts for type difference among different tasks under the setting when task types are known but the reliability vectors $(r_e,r_h)$ are unknown. 
For this purpose, we assume the following condition, which is also used in \citet{BonaldCombes}:
\begin{assumption}\label{assumption:V}
    There exists a positive constant $\bar{r}$ such that 
    $\frac{1}{n} \sum_{i} r_{ki} > \bar{r}$ for all types $k \in \{e, h\}$.
    Further, the co-reliability of workers is non-zero: For every $k$,
    \begin{equation}
        V_k = \min_i \max_{a, b \neq i} \sqrt{\abs{r_{ka} r_{kb}}} > 0 .
    \end{equation}
\end{assumption}

\begin{proposition}\label{prop:achievable_perfect}
    Suppose the number of workers $n$ satisfies 
    \begin{equation}\label{n requirement: achievable_perfect}
        n \geq \sqrt{3\rho / \bar{r}},
    \end{equation} the number of tasks per type satisfies
    \begin{equation}\label{d requirement: achievable_perfect}
        d_k \geq
            C_1\frac{n^2}{V_k^4 \min(\rho^2,\bar{r}^2)}\left(n \Phi(r_k)  + \log (6 n^2)\right) 
        .
    \end{equation}
    for some universal constant $C_1.$
    Then, the TE algorithm to estimate the reliability vectors followed by NP-WMV for label estimation separately for each type achieves a label error rate
    \begin{equation}
      \frac{1}{d}\sum_j P(\hat{y}_j\neq y_j) \leq 3\sum_{k\in \{e,h\}}\frac{d_k}{d} \exp \left(-n \Phi(r_k) \right),
    \end{equation}
    where $\hat{y}_j$ and $y_j$ are the estimated and true labels of task $j$, respectively, and
\begin{equation}\label{eq:averageRenyi}
    \Phi(r_k)= - \frac{1}{n} \sum_{i=1}^n \log \left(\sqrt{(1+r_{ki}) (1-r_{ki})}\right) .
\end{equation}
    % where $C_6 > 0$ is a universal constant.
\end{proposition}
The error exponent \eqref{eq:averageRenyi}\footnote{This error exponent also serves as the asymptotic lower bound for the labeling error for a one-coin DS model corresponding to a reliability type $r_k$(\citet{Gao16}).} for type-dependent weighted majority voting can be related to the error exponent for the type-agnostic weighted majoring voting in \eqref{eq:varphi} through the identity
\begin{equation}
    \Phi(r_k)  = \max_w \varphi_n (w, r_k) , 
\end{equation}
where the maximizing weights are given by the maximum likelihood weights $w$ in \eqref{eq:wmv_weights}.
Recall from Proposition~\ref{thm:WMV}, the lower bound on the error exponent for type-agnostic Weighted Majority Vote is $\max_{w}\min_{k}\varphi(w,r_k)$ and from the definition of $\Phi(r_k)$, it is clear that $\max_{w}\min_{k}\varphi(w,r_k)\leq  \Phi(r_k), \forall k$. This suggests that knowing the type of a task can be helpful. In fact, it is easy to come up with examples where the error rates for the known and unknown task types are dramatically different. 

Consider the case that for $r_{ei} = \beta$ for all $i$ and that $r_{hi} = -\beta$ for all $i$ for some $\beta : 0 < \beta <1$.
Then, 
\begin{align*}
    \max_w \min_k (\varphi (w, r_k)) =  \min_k (\varphi (0, r_k)) = 0,
\end{align*}
whereas $\Phi(r_k) = -\log \left(\sqrt{\beta(1-\beta)}\right) > 0$. In fact, $\Phi(r_k)$ is always positive for a non-zero reliability vector $r_k$. This shows that, while separating tasks by type always leads to an exponentially decaying probability of error, not doing so does not. 

Our analysis in this section suggests that if we have an algorithm that can perfectly cluster tasks by type, then it can be useful to cluster tasks before the label estimation. With this motivation, we now study a spectral clustering algorithm for separating tasks into different types.

\subsection{Spectral Clustering}\label{sec:clustering}

When task types are not known, we propose a spectral algorithm that clusters tasks into two groups.
Crowdsourcing algorithms designed for the Dawid-Skene model can then be applied separately to each group.
We assume that the number of tasks per type is in the same order.
Otherwise, the task types are almost homogeneous, and it is difficult to cluster tasks by type. 
\begin{assumption}\label{assumption:number_of_tasks_per_k}
    There exists $\alpha \in (0, 1)$ such that $d_e = \alpha d$ and $d_h = (1-\alpha) d$.
\end{assumption}

% {\color{blue}Notation: $\hat{k}_j \in \{1, 2\}$ or $\{e, h\}$ where, if we use the latter we mention that this may be the opposite permutation.
% Then reserve $\alpha_1, \alpha_2$ to be the distinct magnitudes in the principal eigenvector. 
% }

\begin{algorithm}[tb]
   \caption{Clustering tasks into hard and easy types}
   \label{alg:cluster}
    \begin{algorithmic}
   \STATE {\bfseries Input:} Worker responses $X \in \{-1, +1\}^{n \times d}$.
   \STATE Compute the principal eigenvector $\hat{v}$ of the task-similarity matrix $T = n^{-1} X^T X$.
   \STATE Set threshold $\hat{\mu} = \frac{1}{d} \sum_j \abs{\hat{v}_j}$.
   \STATE Classify task types by thresholding: $\hat{k}_j = \begin{cases}
           e & \text{if } \abs{\hat{v}_j} \geq \hat{\mu} \\ 
           h & \text{if } \abs{\hat{v}_j} < \hat{\mu}.
       \end{cases} $
   \STATE \textbf{Return:} Task type estimates $\hat{k}_1, \dots, \hat{k}_d$.
\end{algorithmic}
\end{algorithm}
The proposed spectral algorithm is described in Algorithm \ref{alg:cluster}. We adopt the convention that any eigenvector has unit norm in this paper. 
%separates tasks into two disjoint subsets or groups by employing the threshold rule
We note that our clustering algorithm only needs to classify tasks into two groups, as long as all the easy tasks fall into one group and all the hard tasks fall into the other group. Later, we will apply the TE-WMV algoithm separately to each cluster and hence, it does not matter which group we call hard and which group we call easy. Therefore, the clustering error associated with Algorithm \ref{alg:cluster} can be defined as
\begin{equation}\label{eq:eta}
    \eta \coloneqq \min_{\pi: \{e, h\} \to \{e, h\}} \frac{1}{d} \sum_j \mathbf{1}\left\{\pi(\hat{k}_j) \neq k_j\right\}
\end{equation}
 We show that the probability of the event of perfect clustering, that is, $\{\eta = 0\}$ goes to 1 with a rate exponentially fast in $n$. This is precisely stated and shown in Theorem \ref{thm:cluster_crowdsourcing}.

We give an outline of the key ideas involved in proving Theorem \ref{thm:cluster_crowdsourcing} here:
\begin{enumerate}[leftmargin=*]
\item Let the expected task similarity matrix be $\mathbb{E}[T] \coloneqq \mathbb{E}[n^{-1}X^{\top}X].$ We first note that $\mathbb{E}[T]$ can be written in the form $n^{-1}R_y+S$ where $n^{-1}R_y$ is a low-rank matrix and $S$ is a diagonal matrix. The expressions for these matrices are provided in Lemma \ref{lemma: Decomposition of ET}.
\item Then, we show that the principal eigenvector $v(n^{-1}R_y)$ of $n^{-1}R_y$ has a special structure: all the elements corresponding to easy tasks take on the same value in magnitude and all the elements corresponding to hard tasks take on the same value in magnitude. Further, these magnitudes are sufficiently separated under our model for hard and easy tasks under an additional assumption (see Assumption \ref{assumption:spectral_properties_2}) which holds outside a set of measure zero. The precise version of this result is stated in Lemma \ref{lemma: spectral properties of R}. This lemma suggests that, if we had access to $v(n^{-1}R_y),$ then we can cluster tasks by using a threshold to differentiate the magnitudes of the elements of $v(n^{-1}R_y)$. But we do not have access to this eigenvector, therefore the rest of the proof shows that the eigenvector we have access to is a small perturbation of $v(n^{-1}R_y).$
\item Next, we note that
$$T=n^{-1}R_y+S+N,$$
where $N$ is a random matrix noise term given by $N=T-E(T).$ We use matrix Hoeffding inequality to show that this noise term is small in the infinity-norm sense. This is shown in Lemma \ref{lemma: noise matrix concentration}. 
\item Since $S$ is a diagonal matrix, it can be easily shown that its spectral norm is sufficiently small when the number of tasks is large, which is the case in crowdsourcing models. This implies that the spectral norm of $S+N$ is sufficiently small with high probability. Then, using the result of \citet{fan2018eigenvector}, we show that the principal eigenvector of the matrix $T$ which is denoted as $\hat{v}$ has a structure similar to that of $v(n^{-1}R_y),$ i.e., $\hat{v}$ is a perturbed version of $v(n^{-1}R_y),$ in the $l_{\infty}$ norm sense, where the perturbation is small under our hard-easy model. This is shown in Lemma \ref{lemma :l inf concentration} under the assumption that the vectors $r_e$ and $r_h$ are not collinear, which again holds outside a set of measure zero.
\item Now, putting these results together yields our main result in Theorem \ref{thm:cluster_crowdsourcing}, which shows that our spectral clustering algorithm works due to the structure of $v(n^{-1}R_y).$ In other words, if an appropriate threshold is chosen to differentiate the magnitudes of the elements of $\hat{v},$ then this would leads to perfect clustering with high probability.
\end{enumerate}

Now, we will present the lemmas leading up to the main result in the same order as in the outline above. To do this, we suppose for analysis that the tasks are arranged so that easy tasks are in the first $d_e$ columns of $X$ and hard tasks are in the remaining columns.
Knowing the arrangement of columns implies knowledge of task types, but we only use this to simplify exposition and note that this is not used by our algorithm and does not affect our analysis. 

The following Lemma \ref{lemma: Decomposition of ET} presents the decomposition of the expected task similarity matrix $\mathbb{E}[T].$ The proof is straightforward and can be obtained by expressing the expectation of the entries of the task-similarity matrix $T$.

\begin{lemma}\label{lemma: Decomposition of ET}
  The expected task-similarity matrix $\mathbb{E}[T]$ can be written as a perturbation of a low-rank signal $n^{-1} R_y:= n^{-1} \mathrm{diag}(y)\begin{pmatrix}         \|r_e\|_2^2 1_{d_e \times d_e} & r_e^T r_h 1_{d_e \times d_h} \\ r_h^T r_e 1_{d_h \times d_e} & \|r_h\|_2^2 1_{d_h \times d_h}      \end{pmatrix}\mathrm{diag}(y).$
% with entries
% \begin{equation}\label{eq:E_tilde_T_ij}
%     \mathbb{E}[T_{ij}] = \frac{1}{n} \mathbb{E}\left[\sum_{l=1}^n X_{li} X_{lj} \right]
%  \end{equation}
%  for $(i, j) \in [d] \times [d]$ can be written as a perturbation
\begin{align}\label{eq:ET_decomposition_main_part}
   \mathbb{E}[T] = \underbrace{\frac{1}{n} 
   \mathrm{diag}(y) \begin{pmatrix}         \|r_e\|_2^2 1_{d_e \times d_e} & r_e^T r_h 1_{d_e \times d_h} \\ r_h^T r_e 1_{d_h \times d_e} & \|r_h\|_2^2 1_{d_h \times d_h}      \end{pmatrix}\mathrm{diag}(y)}_{n^{-1}R_y}
   % R \odot y y^T 
    + \underbrace{I_d- \frac{1}{n}\mathrm{diag}\left([\|r_e\|_2^2 1_{1\times d_e}, \|r_h\|_2^2 1_{1\times d_h}]^T\right)}_{S}
\end{align}

% Here, $\odot$ denotes the Hadamard product and $I_d$ is the $d\times d$ identity matrix. 
Here, $I_d$ is the $d \times d$ identity matrix and for any natural numbers $a$ and $b,$ $1_{a \times b}$ is the all ones matrix of size $a \times b$. The perturbation matrix $S$ in this context is a diagonal matrix.  
\end{lemma}

The spectral properties of the signal matrix $n^{-1}R_y$ are presented in the next lemma.
 As previously mentioned in the outline, we observe in Lemma \ref{lemma: spectral properties of R} that the magnitude of the entries of $v(n^{-1}R_y)$ are the same for tasks of the same type but are different for different types of tasks. The proof of Lemma \ref{lemma: spectral properties of R} follows from the eigendecomposition of matrix $n^{-1}R_y$ and is presented in Appendix \ref{appendix:spectral_properties}.

\begin{lemma}\label{lemma: spectral properties of R}
     Consider $d_e \geq 1$ and $d_h \geq 1$. 
     The principal eigenvector of the matrix $n^{-1}R_y$ has the following form:

\begin{equation} \label{eq: v(R) structure}
    v(n^{-1}R_y) = \begin{cases}
        diag(y)\begin{bmatrix}
     \frac{s}{\sqrt{s^2d_e+d_h}} 1_{d_e \times 1}\\
     \frac{1}{\sqrt{s^2d_e+d_h}} 1_{d_h \times 1}
\end{bmatrix}, & \text{when $r_e^{\top}r_h \neq 0$}\\
\mathrm{diag}(y) \left[\begin{array}{c}
    \frac{1}{\sqrt{d_e}}1_{d_e\times 1}\\
    0_{d_h\times 1}\end{array}\right] & \text{when $r_e^{\top}r_h = 0$}
    \end{cases} 
\end{equation}

where
\begin{equation}\label{eq: s R}
    s = \frac{d_e\|r_e\|_2^2-d_h\|r_h\|_2^2 + \sqrt{\left[d_e\|r_e\|_2^2-d_h\|r_h\|_2^2\right]^2+4d_ed_h(r_e^Tr_h)^2}}{2d_er_e^Tr_h}.
\end{equation}

  The matrix $n^{-1}R_y$ is a rank-$\ell$ matrix with $\ell \leq 2.$ Let the two largest eigenvalues of the matrix $n^{-1}R_y$ be $\lambda_1(n^{-1}R_y)$ and $\lambda_2(n^{-1}R_y)$ defined in non-decreasing order. They can be expressed as below: 
     \begin{equation}\label{eq: 1st eigenvalue of R}
 \lambda_1(n^{-1}R_y) = \frac{d_e\|r_e\|_2^2+d_h\|r_h\|_2^2 + \sqrt{\left[d_e\|r_e\|_2^2-d_h\|r_h\|_2^2\right]^2+4d_ed_h(r_e^Tr_h)^2}}{2n}  . 
\end{equation}
\begin{equation}\label{eq: 2nd eigenvalue of R}
  \lambda_2(n^{-1}R_y) = \frac{d_e\|r_e\|_2^2+d_h\|r_h\|_2^2 - \sqrt{\left[d_e\|r_e\|_2^2-d_h\|r_h\|_2^2\right]^2+4d_ed_h(r_e^Tr_h)^2}}{2n}  . 
\end{equation}
Clearly when $r_e$ and $r_h$ are collinear, $\lambda_2(n^{-1}R_y)=0$ from definition.\\

\end{lemma}

Let the magnitude of the entries of $v(n^{-1}R_y)$ corresponding to the easy and hard tasks be denoted as $\mu_e(n^{-1}R_y)$ and $\mu_h(n^{-1}R_y)$ respectively. Clearly, from Lemma \ref{lemma: spectral properties of R}, we can write $\mu_e(n^{-1}R_y)$ and $\mu_h(n^{-1}R_y)$ as :
\begin{equation}
    \mu_e(n^{-1}R_y) = \begin{cases}
        \left|\frac{s}{\sqrt{s^2d_e+d_h}}\right| , & \text{when $r_e^{\top}r_h \neq 0$} \\ 
        \frac{1}{\sqrt{d_e}}, & \text{when $r_e^{\top}r_h = 0$}
    \end{cases}
\end{equation}
\begin{equation}
    \mu_h(n^{-1}R_y) = \begin{cases}
       \left| \frac{1}{\sqrt{s^2d_e+d_h}}\right|, & \text{when $r_e^{\top}r_h \neq 0$} \\ 
        0, & \text{when $r_e^{\top}r_h = 0$}
    \end{cases}
\end{equation}
We see when the following assumption is satisfied, $\mu_e(n^{-1}R_y)$ and $\mu_h(n^{-1}R_y)$ are distinct, that is $\mu_e(n^{-1}R_y) \neq \mu_h(n^{-1}R_y)$.
\begin{assumption}\label{assumption:spectral_properties_2}
       When $d_e \neq d_h$, 
    \begin{align*}
        \lvert r_e^T r_h \rvert \neq \left\lvert\frac{ d_e \lVert r_e \rVert_2^2 - d_h \lVert r_h\rVert_2^2 }{d_e - d_h}\right\rvert .
    \end{align*}
  
\end{assumption}
Hence, under the Assumption \ref{assumption:spectral_properties_2}, if we have access to the signal matrix $n^{-1}R_y$, we can differentiate tasks of one type from another using the entries of the vector $v(n^{-1}R_y).$ Specifically, by using the average of the magnitude of the entries of $v(n^{-1}R_y)$ as a threshold, we can separate the elements of this eigenvector into two clusters. Clearly, the threshold is given by 
$$
\mu(n^{-1}R_y)=\frac{d_e}{d}\mu_e(n^{-1}R_y) + \frac{d_h}{d} \mu_h(n^{-1}R_y).
$$
The rest of the lemmas show that such a clustering can be performed with just access to $\hat{v}.$ In particular, we show that each entry of $\hat{v}$ is a perturbed version of the corresponding entry of $v(n^{-1}R_y)$ where the magnitude of the perturbation is, with high probability, at most $\frac{1}{2}\min(m_e(n^{-1}R_y),m_h(n^{-1}R_y)),$
where $m_e(n^{-1}R_y)=|\mu_e(n^{-1}R_y)-\mu|$ and $m_h(n^{-1}R_y)=|\mu-\mu_h(n^{-1}R_y)|.$ A little thought shows that this would imply that all tasks are clustered perfectly.
\begin{comment}
    In the following discussion, we assume that the Assumption \ref{assumption:spectral_properties_2} is satisfied. A natural way to cluster the entries of $v(n^{-1}R_y)$ by their magnitude is to use a threshold $\mu$ that lies between $\mu_e(n^{-1}R_y)$ and $\mu_h(n^{-1}R_y)$. One such threshold that motivates our Algorithm \ref{alg:cluster} is the average $\mu = d^{-1} \sum_{j=1}^d \lvert v_j(n^{-1}R_y) \rvert$ of magnitudes in $v(n^{-1}R_y)$ given by
\begin{equation}
    \mu = \frac{d_e}{d}\mu_e(n^{-1}R_y) + \frac{d_h}{d} \mu_h(n^{-1}R_y) .
\end{equation} 
We define the absolute margin $m_k$ from the threshold $\mu$ associated with a type $k$ task as
\begin{equation}
    m_k = \lvert \mu - \mu_k \rvert .    
\end{equation}
Now the algorithm has access to the principal eigenvector of the task similarity matrix $T$ denoted by $\hat{v}.$ The Algorithm \ref{alg:cluster} uses $\hat{\mu} = d^{-1} \sum_{j=1}^d \lvert \hat{v}_j \rvert$ as a threshold for clustering takes based on the magnitude of entries of $\hat{v}.$ We show in Lemma \ref{lemma :l inf concentration} that the $\hat{v}$ is a perturbation of $v(n^{-1}R_y)$ where the perturbation is small with high probability in the $l_{\infty}$ norm sense when we have a sufficiently large number of workers $n.$ 
\end{comment}

Before proceeding further, we define the normalized spectral gap of $n^{-1} R_y$ as
\begin{equation}\label{eq: spectral gap}
    \nu (n^{-1} R_y) =  d^{-1}\min \left\{
        \lambda_1 (n^{-1}R_y) - \lambda_2 (n^{-1} R_y), \lambda_2 (n^{-1} R_y)
    \right\} .
\end{equation}
$\nu(n^{-1}R_y)$ would be an important quantity to characterize the performance of our clustering algorithm later. For using the matrix perturbation result in \citet{fan2018eigenvector} in the following parts, it is required that the normalized spectral gap $\nu (n^{-1} R_y) \neq 0$ which holds when the following assumption is satisfied (see the last statement of Lemma \ref{lemma: spectral properties of R}):

\begin{assumption}\label{assumption: not collinear}
    The reliability vectors $r_e$ and $r_h$ are not collinear.
\end{assumption}

 From Lemma \ref{lemma: Decomposition of ET}, we can view the matrix $T$ as a perturbation of the signal matrix $n^{-1}R_y$ as follows:
\begin{equation}\label{eq: decomposition of T}
    T = \mathbb{E}[T] + N = n^{-1}R_y + S +N.
\end{equation}
Let the infinity norm of a square matrix $M$ be $\lVert M\rVert_\infty = \max_i \sum_{j} \lvert M_{ij} \rvert$. The following matrix concentration ineqaulity shows how the noise matrix $N$ is small in the infinity-norm sense for sufficiently large $n$; the proof can be found in subsection \ref{appendix: proof of concentration of N}. 
\begin{lemma} \label{lemma: noise matrix concentration} For any $t > 0$ and any positive values of $n$ and $d$, the task-similarity matrix $T$ concentrates around its expectation as follows: 
    \begin{equation}\label{eq: noise matrix concentration}
    \mathbb{P}\left(\norm{N}_{\infty} \geq t \right) \leq 2d^2 \exp \left(
        -\frac{n t^2}{2 d^2}
    \right) .
\end{equation}
\end{lemma}

On the other hand, the matrix $S$ is a diagonal matrix with each diagonal entry belonging to the following set: $\{1-n^{-1}\norm{r_e}, 1-n^{-1}\norm{r_h} \} $.  Hence, $\norm{S}_{\infty} = 1-n^{-1}\norm{r_h} \leq 1.$ Observe that the normalized spectral gap $ \nu(n^{-1}R_y)$ of the low-rank signal matrix 
$n^{-1}R_y$ given in equation \eqref{eq: spectral gap} is of the order of $O(1)$.  These observations along with Lemma \ref{lemma: noise matrix concentration} show that the perturbation matrix $S+N$ to the signal $n^{-1}R_y$ in equation \eqref{eq: decomposition of T} is small in the infinity norm sense, compared to the spectral gap $d \nu(n^{-1}R_y)$ of the matrix $n^{-1}R_y$ for sufficiently large $n.$ This motivates us to use a matrix perturbation result for a low-rank signal matrix derived in \citet{fan2018eigenvector} (see Theorem 3 in \citet{fan2018eigenvector}) to show that the principal eigenvector of the task-similarity matrix is a perturbed version of $v(n^{-1}R_y)$ in the following sense. 
See subsection \ref{appendix: l_inty concentration} for the proof.
\begin{lemma} \label{lemma :l inf concentration}
   Under the assumption \ref{assumption: not collinear}, if $\lambda_2(n^{-1}R_y)$ satisfies : $ C_2 (\min(\alpha,1-\alpha))^2 \nu(n^{-1}R_y)d - 1 >0,$ then,  for every $0 < \epsilon <( C_2 (\min(\alpha,1-\alpha))^2 \nu (n^{-1}R)d - 1)$,
    \begin{align}\label{eq: l inf concentration}
        & \mathbb{P}\left(
\min_{\theta \in \{-1, +1\}} \lVert  \theta \hat{v} - v(n^{-1}R_y) \rVert_{\infty} \geq 
    C_3 \frac{(\epsilon  + 1)}{(\min (\alpha,1-\alpha))^2\nu(n^{-1}R_y) d\sqrt{d}} 
        \right) \leq 2d^2 \exp \left(-n\frac{\epsilon^2}{2d^2}\right) ,
    % \min_{\theta \in \{-1, +1\}} \lVert v(\mathbb{E}[T]) - \theta \hat{v}\rVert_\infty \leq 2^{3/2} \left(\lambda_1 (n^{-1} R_y) - \lambda_2 (n^{-1} R_y)\right)^{-1} + 3^{2} \cdot 2^{-2} \frac{\lambda_2 (n^{-1} R_y)}{\nu \sqrt{d}} .
    \end{align}
       
    where $C_2$ and $C_3$ are universal constants. 
\end{lemma}

\begin{comment}
    Next, we show in Proposition \ref{prop: perfect clustering events} the event of perfect clustering by Algorithm \ref{alg:cluster} is guaranteed when the $l_{\infty}$ gap between $\hat{v}$ and $v(n^{-1}R_y)$ is sufficiently small.  

\begin{proposition}\label{prop: perfect clustering events}
   Define the following event based on the concentration of $\hat{v}$ around $v(n^{-1}R_y)$ in the $l_{\infty}$ norm sense :
\begin{equation}
    E_{L_\infty} = \left\{
        \min_{\theta \in \{-1, +1\}} \lVert v(n^{-1}R_y) - \theta \hat{v} \rVert_{\infty}  < \frac{1}{2} \min \left\{ m_e(n^{-1}R_y), m_h(n^{-1}R_y)\right\} 
    \right\}  ,
\end{equation}
Under the event $E_{L_\infty}$, Algorithm \ref{alg:cluster} achieves perfect clusterin, that is $\eta = 0$
\end{proposition}

From Lemma \ref{lemma :l inf concentration}, We can see that the events $E_{L_\infty}$ occur with high probability when we choose $\epsilon$ in the equation \eqref{eq: l inf concentration} such the following inequality is satisfied 
 \begin{equation}\label{eq: condition pc}
     C_3 \frac{(\epsilon d + 1)}{4 (\max (\alpha,1-\alpha))^2\nu(n^{-1}R_y) d\sqrt{d}} \leq \frac{1}{2} \min \left\{ m_e(n^{-1}R_y), m_h(n^{-1}R_y)\right\}
 \end{equation}
 We observe that $ \frac{1}{2} \min \left\{ m_e(n^{-1}R_y), m_h(n^{-1}R_y)\right\}$ is a order $O\left(\frac{1}{\sqrt{d}}\right)$ quantity and $\mu{n^{-1}R_y}$ is a order $O(1)$ quantity. Hence, the condition \eqref{eq: condition pc} is satisfied for sufficiently small $\epsilon.$
\end{comment}

Putting all the above lemmas together yields the main result in Theorem \ref{thm:cluster_crowdsourcing}. The proof of this theorem essentially is based on the following fact: using the Lemma \ref{lemma :l inf concentration} we show that each element of $v(n^{-1}R_y)$ is perturbed by an amount at most $\frac{1}{2}\min(\mu_h(n^{-1}R_y),\mu_e(n^{-1}R_y)).$ The detailed proof is given in Appendix \ref{Appendix: proof of thm: perfect clustering}.

\begin{theorem}\label{thm:cluster_crowdsourcing}
 Under the stated assumptions,
     if the number of tasks $d$ satisfies
    \begin{equation}\label{eq:cluster_d_requirement_main}
               d \geq \frac{C_4}{\sqrt{D(r_e,r_h, \alpha,d)}},
    \end{equation}
     then, Algorithm \ref{alg:cluster} satisfies
    \begin{equation}\label{eq:perfect clustering prob}
        P(\eta=0)\geq 1- 2d^2\exp\left(-C_{5}nD(r_e,r_h, \alpha,d)\right),
    \end{equation}
    where the problem-dependent quantity $D(r_e,r_h, \alpha,d)$ characterizing the error exponent and the requirement on $d$ is given by
    
    \begin{equation}\label{eq:D}
        D(r_e,r_h, \alpha,d) = \begin{cases}
         \left((\min(\alpha,1-\alpha))^3\frac{\nu(n^{-1}R_y)||s|-1|}{\sqrt{s^2+1}}\right)^2 & \text{when, } r_e^{\top}r_h \neq 0,\\
         \left((\min(\alpha,1-\alpha))^3\nu(n^{-1}R_y)\right)^2 & \text{when, } r_e^{\top}r_h = 0,
    \end{cases}
    \end{equation}
    and $C_4$ and $C_5$ are universal constants, independent of the problem parameters. 
\end{theorem}

\subsection{Label Estimation for Hard-Easy Tasks}
After having divided the tasks into two clusters, we are now set to estimate the true labels $y \in \{-1, +1\}^d$ of the tasks. In practice, one can simply apply a DS algorithm, such as TE, to each task type separately. However, analyzing such an algorithm is difficult because the clustering step and label estimation steps are correlated due to the fact that we use the same dataset for both. Therefore, as is common in the literature (see \citet{OBIWAN}, for example), we split the $n$ workers into two disjoint groups and use the responses of one group for clustering and the other group for label estimation.
We present these details next.

For the following analysis, let $\mathcal{N}_{cl}$ be the set of workers used for clustering, and define $\mathcal{N}_{rl} = [n] - \mathcal{N}_{cl}$ to be the set of workers that will be used for reliability estimation as well as label estimation. Let the responses of the workers in the set $\mathcal{N}_{cl}$ be denoted by 
\begin{align*}
    \mathrm{X}_{cl} := (X_{ij}: (i, j) \in \mathcal{N}_{cl} \times [d]),
\end{align*}
and the worker responses of the set $\mathcal{N}_{rl}$ be 
\begin{align*}
    \mathrm{X}_{rl} := (X_{ij}: (i, j) \in \mathcal{N}_{rl} \times [d]).
\end{align*}
We use Algorithm~\ref{alg:cluster} to cluster the tasks in
$\mathrm{X}_{cl}$ using Algorithm \ref{alg:cluster} (with $X = \mathrm{X}_{cl}$) resulting in the following type assignment for all task $j \in [d]$:
\begin{align*}
    \mathcal{T}_k = \left\{
        j \in [d]: \hat{k}_j = k
    \right\} , k \in \{e, h\} .
\end{align*}
We then use the TE algorithm to estimate reliabilities $\mathrm{\hat{r}}_k = (\hat{r}_{ki}: i \in \mathcal{N}_{rl})$ from the responses $\left(X_{ij}: (i, j) \in \mathcal{N}_{rl} \times \mathcal{T}_k\right)$ for each $k$.
Lastly, the labels $y_j$ are estimated using the NP decision rule
\begin{equation}\label{eq:ultimate_label_estimation}
    \hat{y}^{TE}_j = \mathrm{sgn}\left(\sum_{i\in \mathcal{N}_{rl}} \log \frac{1 + \hat{\mathrm{r}}_{\hat{k}_ji}}{1 - \hat{\mathrm{r}}_{\hat{k}_ji}} X_{ij} \right) .
\end{equation}

Now we are ready to present the theorem characterizing the accuracy of our combined clustering and label estimation algorithm. 
Let $n_{cl}$ and $n_{rl}$ be the number of workers in the sets $\mathcal{N}_{cl}$ and $\mathcal{N}_{rl},$ respectively. Let $r_{k}(\mathcal{N}_{cl})$ and $r_{k}(\mathcal{N}_{rl})$ be the reliability vector associated with each task type $k$ for the set of workers $\mathcal{N}_{cl}$ and $\mathcal{N}_{rl},$ respectively.

\begin{theorem}\label{thm:labeling easy} 
    Suppose $\left(n_{rl}, d_e,d_h,r_k(\mathrm{N}_{rl})\right)$ satisfy the conditions stated in Proposition \ref{prop:achievable_perfect} and 
    $\left(n_{cl}, d_e,d_h,r_k(\mathrm{N}_{cl})\right)$ satisfy the conditions from Theorem \ref{thm:cluster_crowdsourcing}.
    Then, for the hard-easy crowdsourcing model under the stated assumptions, the labels $\hat{y}$ estimated using \eqref{eq:ultimate_label_estimation} satisfy
    % Then, the expected label estimation error averaged over all tasks is 
\begin{align*}
         & \mathbb{E}\left(\frac{1}{d} \sum_j {\mathbf 1}\left(\hat{y}_j \neq y_j\right)\right) \\
         & \leq  
        3\left[\sum_{k\in \{e,h\}}\frac{d_k}{d}\exp \left(-n_{rl} \Phi_{k,\mathcal{N}_{rl}}\right)
       \right] +
        2 d^2 \exp \left(- C_{5} n_{cl} D(r_e(\mathcal{N}_{cl}),r_h(\mathcal{N}_{cl}),\alpha,d)\right)
\end{align*}
 where $\Phi_{k,\mathcal{N}_{rl}} \coloneqq \Phi(r_{k}(\mathcal{N}_{rl}))$ and $D(r_e(\mathcal{N}_{cl}),r_h(\mathcal{N}_{cl}),\alpha,d)$ is defined similarly to $D(r_e,r_h, \alpha,d)$ in equation \eqref{eq:D}, with the obvious changes to account for the fact that we are only using the reduced dataset $\mathrm{X}_{cl}$ for clustering.
\end{theorem}
The proof of the Theorem \ref{thm:labeling easy} is an immediate application of the Theorem \ref{thm:cluster_crowdsourcing} and is provided in Appendix \ref{appendix:proof of labeling}. 

\section{Discussion} \label{sec:discussion}

In the previous sections, we proposed the clustering Algorithm \ref{alg:cluster} and showed that for a hard-easy model, we can cluster tasks by type with a reasonable number of workers and tasks. Recall that Theorem \ref{thm:cluster_crowdsourcing}, which characterizes the performance of Algorithm \ref{alg:cluster}, requires the reliability vectors $r_e$ and $r_h$ to be not collinear (Assumption \ref{assumption: not collinear}). Even though this is a zero measure set, an interesting technical question is whether this assumption is needed for our algorithm to work. In what follows, we show that a large fraction of tasks will be clustered correctly with high probability even when this assumption is not satisfied. 

We notice from Lemma \ref{lemma: spectral properties of R} that even when $r_e$ and $r_h$ are collinear, the principal eigenvector $v(n^{-1}R_y)$ retains the structure that reveals the type information for each task. A similar structure is observed for the principal eigenvector of 
$\mathbb{E}[T]$ too. Specifically, we can cluster tasks by type by clustering the magnitude of entries of $v(\mathbb{E}[T])$ into two groups even when $r_e$ and $r_h$ are collinear. The properties of $v(\mathbb{E}[T])$ is established in the appendix \ref{appendix: spectral properties of ET}. We exploit this property to show in the following Theorem \ref{thm:imperfect_cluster} that the Algorithm \ref{alg:cluster} achieves arbitrarily small clustering error $\eta$ with high probability even without the assumption \ref{assumption: not collinear}. 

 Let $\lambda_1(\mathbb{E}[T])$ and $\lambda_2(\mathbb{E}[T])$ be the first and second largest eigen values of $\mathbb{E}[T]$ and the normalized Eigen-gap of the expected task similarity matrix $\mathbb{E}(T)$ be $\sigma(\mathbb{E}[T]) = d^{-1}(\lambda_1(\mathbb{E}[T])-\lambda_1(\mathbb{E}[T]))$. Similar to the case of $n^{-1}R_y$ in Lemma \ref{appendix: l_inty concentration}, let us denote the ratio between the entries of the principal eigenvector of $\mathbb{E}[T]$ corresponding to easy and hard tasks by $\gamma$.
 %The normalized eigengap $\sigma(\mathbb{E}[T])$ of the matrix $\mathbb{E}[T]$ can be expressed in terms of the problem parameters as :
%\begin{equation}
    %\sigma(\mathbb{E}[T]) = \frac{1}{n} \sqrt{[ (d_e-1) \lVert r_e \rVert_2^2 - (d_h-1) \lVert r_h \rVert_2^2]^2 + 4 d_e d_h (r_e^T r_h)^2}.
%\end{equation}
\begin{theorem}[Imperfect Clustering]\label{thm:imperfect_cluster} Assume $\min(d_e,d_h) \geq 2.$
     Then, for the Hard-Easy crowdsourcing model, under Assumptions \ref{assumption:easyhard}, \ref{assumption:rho}, \ref{assumption:V}, \ref{assumption:number_of_tasks_per_k} , and if the following is satisfied :  when $d_e \neq d_h$, 
    \begin{align*}
        \lvert r_e^T r_h \rvert \neq \left\lvert\frac{ (d_e-1) \lVert r_e \rVert_2^2 - (d_h-1) \lVert r_h\rVert_2^2 }{d_e - d_h}\right\rvert ,
    \end{align*}
      Algorithm \ref{alg:cluster} returns cluster membership with the following confidence on the clustering error: we have for every $t \in [0,1)$
    \begin{equation}
        \mathbb{P}(\eta > t) \leq \begin{cases}
            4d \exp \left(
            - C_6\left(\sigma(\mathbb{E}[T]) \min \left\{\alpha,1-\alpha \right\} \frac{||\gamma|-1|}{\sqrt{\gamma^2+1}}\right)^2 nt
        \right) & \text{when } r_e^{\top}r_h \neq 0\\
        4d \exp \left(
            - C_6\left(\sigma(\mathbb{E}[T]) \min \left\{\alpha,1-\alpha \right\} \right)^2 nt
        \right) & \text{when } r_e^{\top}r_h = 0
        \end{cases} 
        % 4d\exp\left(-\tilde{C}_5nt\right)
    \end{equation}
    where $C_6$ is an absolute constant.
\end{theorem}
The proof of the above theorem is given in the Appendix section \ref{appendix: proof of imperfect cluster}. A proof sketch is given here:
\begin{enumerate}[leftmargin=*]
    \item We show that the entries principal eigenvector of $\mathbb{E}[T]$ contain task type information (see Appendix \ref{appendix: spectral properties of ET} for the proof).
    \item Unlike the proof of Theorem \ref{thm:cluster_crowdsourcing}, we treat $\mathbb{E}[T]$ as the signal matrix for this proof. We apply Davis-Kahan perturbation result (\citet{DavisKahan}) to show the principal eigenvector of $T$, $\hat{v}$ is a small perturbation of the eigenvector of the signal $\mathbb{E}[T]$ in the $l_2$-norm sense. A concentration of the noise matrix $N = T - \mathbb{E}[T]$ in the $l_2$ norm is used for this purpose (see Appendix  \ref{appendix: proof of imperfect cluster} ).
    \item Finally, we relate the event $\eta \leq t$ for some $t \in [0,1]$ to the concentration of principal eigenvectors in the $l_2$-norm sense (again see Appendix  \ref{appendix: proof of imperfect cluster} ).
\end{enumerate}

While the above result shows that a large fraction of tasks will be clustered correctly even without the non-collinearity assumption on $r_e$ and $r_h$, the result does not show that perfect clustering is possible with high probability. As a result, one cannot use the above theorem to establish a result like Theorem \ref{thm:cluster_crowdsourcing}. Nevertheless, the above theorem raises the interesting possibility that the non-collinearity assumption may not be necessary for good performance. Establishing such a result would be an interesting direction for future work.

\section{Proof of Theorem \ref{thm:cluster_crowdsourcing}: Perfect Clustering}\label{appendix: perfect clustering}
This section proves the clustering Theorem \ref{thm:cluster_crowdsourcing}. As discussed in the proof sketch of the Theorem, the first step is to show that the principal eigenvector $v(n^{-1}R_y)$ of the signal matrix $n^{-1}R_y$ reveals the type information for each task. This is discussed in detail in Lemma \ref{lemma: spectral properties of R} and proved in Appendix \ref{Appendix: proof of lemma: spectral properties of R}. Building upon the Lemma \ref{lemma: spectral properties of R}, the rest of the proof of Theorem \ref{thm:cluster_crowdsourcing} is given in this section as enlisted below.

\begin{enumerate}[leftmargin=*]
    \item First, we prove the Lemma \ref{lemma: noise matrix concentration} in the subsection \ref{appendix: proof of concentration of N}. 
    \item Then we Show that the principal eigenvector $\hat{v}$ of the task-similarity matrix $T$ is a small perturbation of $v(n^{-1}R_y)$ in the $l_{\infty}$ sense. This is stated in Lemma \ref{lemma :l inf concentration} and proved in the following subsection \ref{appendix: l_inty concentration}.
    \item Next, we relate the event of perfect clustering, that is $\{\eta = 0\}$ with a sufficient condition on the concentration of $\hat{v}$ with respect to $v(n^{-1}R_y)$. (see Proposition \ref{prop:conditions for perfect clustering} in the subsection \ref{appendix: conditions for perfect clustering}).
    \item Finally, we prove that the condition described in The Proposition \ref{prop:conditions for perfect clustering} is satisfied with high probability. See section \ref{Appendix: proof of thm: perfect clustering} for this final step.
\end{enumerate}
%Let $\theta \in \{-1,1\}$ is a binary variable used to resolve the sign ambiguity of eigenvectors in the following analysis whenever necessary.

\subsection{Concentration of the Noise Matrix \texorpdfstring{$N$}{N}}\label{appendix: proof of concentration of N}
  The proof of the Lemma \ref{lemma: noise matrix concentration} stating the concentration of $N$ is given as: 
\begin{align*}
    \mathbb{P}\left(\|N\|_{\infty}\geq \epsilon\right) & =  \mathbb{P}\left(\max_{i\in [d]}\sum_{j=1}^d\left|T_{ij}-\mathbb{E}[T_{ij}]\right| \geq \epsilon \right) \underbrace{\leq}_{(a)} \sum_{i=1}^d \mathbb{P}\left(\sum_{j=1}^d\left|T_{ij}-\mathbb{E}[T_{ij}]\right| \geq \epsilon \right)\nonumber\\ & \leq \sum_{i=1}^d \mathbb{P}\left(\max_{j\in [d]}\left|T_{ij}-\mathbb{E}[T_{ij}]\right| \geq \frac{\epsilon}{d} \right) \underbrace{\leq}_{(b)} 
    \sum_{i=1}^d \sum_{j=1}^d  \mathbb{P}\left(\left|T_{ij}-\mathbb{E}[T_{ij}]\right| \geq \frac{\epsilon}{d}\right)\nonumber\\& = \sum_{i=1}^d \sum_{j=1}^d  \mathbb{P}\left(\left|\frac{1}{n}\sum_{l=1}^n \left(X_{li}X_{lj}-\mathbb{E}[X_{li}X_{lj}]\right)\right| \geq \frac{\epsilon}{d}\right)\\ 
    & \underbrace{\leq}_{(c)} 2d^2 \exp \left(-n\frac{\epsilon^2}{2d^2}\right) \numberthis \label{eq:Ninfty_concentration}
\end{align*}
In $(a)$ and $(b)$ we use the union bound, and in $(c)$ we employ Hoeffiding's inequality for the independent bounded random variables $X_{li}X_{lj}\in \{\pm 1\}$.

\subsection{\texorpdfstring{$l_\infty$}{} Concentration of the Principal Eigenvector}  \label{appendix: l_inty concentration}
We prove the Lemma \ref{lemma :l inf concentration} here.
    
First, we need to define a quantity called the coherence of the signal matrix $n^{-1}R_y$. Writing the modal matrix of $n^{-1}R_y$ which is of size $d \times \ell$ as $V$ so that its columns correspond to the unit-norm eigenvectors of $n^{-1}R_y$, the coherence $M$ of matrix $n^{-1} R_y$ is defined as
\begin{equation}\label{eq:coherence}
    M = \frac{d}{\ell} \max_{i \in [d]} \sum_{j=1}^\ell V_{ij}^2 .
\end{equation}

Recall the low-rank decomposition $T = n^{-1} R_y + S + N $
in \eqref{eq:ET_decomposition_main_part}, where $N = T - \mathbb{E}[T]$ and $S=- \frac{1}{n}\text{diag}\left([\|r_e\|_2^2 1_{1\times d_e}, \|r_h\|_2^2 1_{1\times d_h}]^T\right) + I_d$. Here we are interested in the distance between $\hat{v}$ which is the principal eigenvector of $T$ and $v(n^{-1}R_y)$ induced by the infinity norm.
We utilize the following result by \citet{fan2018eigenvector}, cf. Theorem 3.\footnote{The theorem in \citet{fan2018eigenvector} is for a matrix of rank $\ell$ where $\ell$ can take any finite value, we simplified it for our purpose when $\ell = 2.$} 
\begin{lemma}\label{lemma:fan2018}
    Consider a rank-2 symmetric matrix $A$ and its eigen-decomposition
    \begin{equation}
        A = \sum_{g=1}^2 \lambda_g (A) v_g (A) v_g(A)^T .
    \end{equation}
    Denote by $\nu(A) = d^{-1}\min(\lambda_1(A)-\lambda_2(A),\lambda_2(A)), M(A), \nu(A)$ to be the normalized spectral gap, coherence, and principal eigenvector of $A$.
    For a symmetric matrix $\hat{A}$, if the second eigenvalue of $A$ satisfies
\begin{equation}\label{eq:eigval_condition_infty}
    \lvert \lambda_2 (A) \rvert \geq \max \left\{
        3, 8 M(A) ( 1 + 4 \sqrt{2 M(A)}, 2^8 (1 + 2 M(A)) M(A) )
    \right\}  \lVert A - \hat{A}\rVert_\infty ,
\end{equation}
and the normalized spectral gap $\nu(A)$ satisfies 
\begin{equation}\label{eq:nu condition fan}
   \nu(A) > \norm{A-\hat{A}}_{\infty} 
\end{equation}
    then 
    \begin{equation}
        \min_{\theta \in \{-1, +1\}} \lVert v_1(A) - \theta v_1(\hat{A})\rVert \leq 3 \cdot 2^7 \frac{(1 + 2M(A)) M(A) \lVert A - \hat{A}\rVert_\infty}{\lambda_2 (A) \sqrt{d}} + 
        2^{7/2} \frac{\sqrt{M(A)} \lVert A - \hat{A}\rVert_2}{\nu(A) d\sqrt{d}} ,
    \end{equation}
    where, $v_1(\hat{A})$ denotes the principal eigenvector of the matrix $\hat{A}.$
\end{lemma}
The coherence $M$ of $n^{-1} R_y$ by definition is necessarily $M \geq 2^{-1}$, and so
\begin{align*}
    \max \left\{ 
        3, 8 M (1 + 4 \sqrt{2M}), 2^8 (1 + 2M)M 
    \right\} \leq 2^{10} M^2 .
\end{align*}
To apply the above result, we substitute the matrix $A$ with $n^{-1}R_y$ and the perturbation $\hat{A}-A$ with $S+N.$ 

Next, we define an event $E_N$ on the random noise matrix $N$ to ensure the conditions \eqref{eq:eigval_condition_infty} and \eqref{eq:nu condition fan}. Specifically, the conditions \eqref{eq:eigval_condition_infty} and \eqref{eq:nu condition fan} are satisfied when we have 
\begin{equation}
     \nu(n^{-1}R_y)d  \geq 2^{10} M^2 \lVert T - n^{-1} R_y\rVert_\infty = 2^{10} M^2 \lVert S + N \rVert_{\infty} ,
\end{equation}
or equivalently
\begin{align*}
        \lVert S + N \rVert_\infty \leq 2^{-10} M^{-2} \nu(n^{-1}R_y)d 
\end{align*}
Define the event $E_{N}$ as:
\begin{align*}
    E_N := \left\{
        \lVert N \rVert_{\infty} \leq C_2 \frac{\nu(n^{-1}R_y)d}{4M^2} - 1
    \right\} 
\end{align*}
where $C_2 = 2^{-8}$. Clearly, on the event $E_N$ the conditions \eqref{eq:eigval_condition_infty} and \eqref{eq:nu condition fan} are satisfied by the use of the triangle inequality with the fact that $\lVert S \rVert_\infty = 1 - n^{-1} \lVert r_h \rVert_2^2$ for the diagonal matrix $S$.

Now conditioning on the event $E_N$ we can use the Lemma \ref{lemma:fan2018} as: 
\begin{align*}
    \min_{\theta \in \{-1, +1\}} \|v(n^{-1} R_y) - \theta \hat{v} \|_\infty & \leq \frac{3\cdot2^7(1+2M)M\|S+N\|_{\infty}}{\lambda_2(n^{-1}R_y)\sqrt{d}} + \frac{2^{\frac{7}{2}}\sqrt{M}\|S+N\|_2}{\nu(n^{-1}R_y) d\sqrt{d}}\\
    & \underbrace{\leq}_{(a)} \frac{3\cdot2^7(1+2M)M\|S+N\|_{\infty}}{\nu(n^{-1}R_y) d\sqrt{d}} + \frac{2^{\frac{7}{2}}\sqrt{M}\|S+N\|_{\infty}}{\nu(n^{-1}R_y)  d\sqrt{d}}\\
    & \underbrace{\leq}_{(b)}
    \left[\frac{3\cdot2^{10}M^2}{\nu(n^{-1}R_y) d\sqrt{d}}\right]\left[\|N\|_{\infty}+1\right] \numberthis \label{eq:dist_Ry_T}
    % \\ 
    % & \leq 3^2 \cdot 2^{-2}
    % \frac{\lambda_2 (n^{-1} R_y) }{\nu\sqrt{d}} \numberthis \label{eq:dist_Ry_T}
\end{align*}
In (a), we use the fact that $\nu(n^{-1}R_y) d \leq \lambda_2(n^{-1}R_y)$, in (b), we use $\|S\|_{\infty} \leq 1$ and $M \geq \frac{1}{2}$.

We are interested the event $E_N \cap \left\{\lVert N\rVert_{\infty} \leq \epsilon \right\}$ for some $\epsilon$ such that, $0 < \epsilon \leq 2^{-2}C_2 M^{-2} \lambda_2 (n^{-1} R_y) - 1$.
On the event  $E_N \cap \left\{\lVert N\rVert_{\infty} \leq \epsilon \right\}$, the following is satisfied using \eqref{eq:dist_Ry_T}:  
\begin{equation}\label{eq:l_inf temp}
    \min_{\theta \in \{-1, +1\}} \lVert v(n^{-1}R_y) - \theta \hat{v}\rVert_\infty \leq
    C_3 \frac{4M^2}{\nu(n^{-1}R_y) d\sqrt{d}}\left(\epsilon + 1\right) 
    .
\end{equation}
where $C_3 = 3\cdot2^{8}$.  
It remains to show that the event $E_N \cap \left\{\lVert N\rVert_{\infty} \leq \epsilon \right\}$ for some $\epsilon$ in the range $( 0,2^{-2}C_2 M^{-2} \lambda_2 (n^{-1} R_y) - 1]$  occurs with high probability: 
\begin{align*}
    \mathbb{P}\left(E_N \cap \left\{\lVert N\rVert_{\infty} \leq \epsilon \right\}\right) \underbrace{=}_{(c)} 1 - \mathbb{P}(\left\{\lVert N\rVert_{\infty} \leq \epsilon \right\}^c) \geq 1 - 
    2d^2 \exp \left(\frac{-n\epsilon^2}{2d^2}\right)
    % 2d^2 \exp \left(-n C_3 \left(\frac{\lambda_2 (n^{-1}R_y)^2 - M^2}{d M^2} \right)^2\right)
\end{align*}
where in $(c)$ we use the fact that the event $\left\{\lVert N\rVert_{\infty} \leq \epsilon \right\}$ is a subset of the event $E_{N}.$

Lastly, we want to give an upper bound on the coherence parameter $M$ to arrive at the final form as in Lemma \ref{lemma :l inf concentration}. For non-collinear $r_e$ and $r_h$, the two non-zero eigenvectors for the signal matrix $n^{-1}R_y$ can be written as $\mathrm{diag}(y)[s 1_{1\times d_e}, 1_{1\times d_h}]^T$ and $\mathrm{diag}(y)[\overline{s} 1_{1\times d_e}, 1_{1\times d_h}]^T$  where $s$ and $\overline{s}$ takes the following values:
\begin{equation}
   s,\overline{s} =  \frac{d_e\|r_e\|_2^2-d_h\|r_h\|_2^2\pm \sqrt{\left[d_e\|r_e\|_2^2-d_h\|r_h\|_2^2\right]^2+4d_ed_h(r_e^Tr_h)^2}}{2d_er_e^Tr_h} .
\end{equation}
The above statement is proved in the Appendix \ref{Appendix: proof of lemma: spectral properties of R}.
From Lemma \ref{lemma: spectral properties of R}, the elements of $v(n^{-1}R_y)$ corresponding to easy and hard tasks are as $\frac{s^2}{d_e s^2+d_h}$ and $\frac{1}{d_e \overline{s}^2+d_h}$, respectively. Similarly, the corresponding entries of the second eigenvector of $n^{-1}R_y$ would be $\frac{\overline{s}^2}{d_e \overline{s}^2+d_h}$ and $\frac{1}{d_e \overline{s}^2+d_h}$. From the expressions obtained above, we can write the coherence defined in the equation \eqref{eq:coherence} as
\begin{align*}
    M = \frac{d}{\ell} \max_{i \in [d]} \sum_{j=1}^\ell V_{ij}^2 = \frac{d}{2} \max \left\{
        \frac{s^2}{d_e s^2 + d_h} + \frac{\overline{s}^2}{d_e \overline{s}^2 + d_h}, \frac{1}{d_e s^2 + d_h} + \frac{1}{d_e \overline{s}^2 + d_h}
    \right\}
\end{align*}
Hence, we can upper bound the coherence term $M$ as:
\begin{align*}
     M 
     \leq \frac{1}{2} \left( \frac{ds^2+d}{d_es^2+d_h}+\frac{d\overline{s}^2+d}{d_e\overline{s}^2+d_h}\right)
     \underbrace{\leq}_{(e)} \frac{1}{2} \left( \frac{s^2+1}{\alpha s^2+(1-\alpha)}+\frac{\overline{s}^2+1}{\alpha \overline{s}^2+(1-\alpha)}\right)
    % & \leq \frac{1}{2\min(\alpha,1-\alpha)}
    \leq  \frac{1}{\min(\alpha,1-\alpha)} 
\end{align*}
where in $(e)$, we use the Assumption \ref{assumption:number_of_tasks_per_k}.
Using this upper bound in the equation \eqref{eq:l_inf temp} proves the Lemma \ref{lemma :l inf concentration}.

\subsection{Sufficient Condition for Perfect Clustering}\label{appendix: conditions for perfect clustering}
Here, we relate the event of perfect clustering with the concentration of the principal eigenvector $\hat{v}$ with respect to $v(n^{-1}R_y)$. 

\begin{proposition}\label{prop:conditions for perfect clustering}
    Under the stated assumptions, Algorithm \ref{alg:cluster} achieves perfect clustering, that is $\eta=0$ when the following event occurs : 
    \begin{equation}
    E_{l_\infty} = \left\{
        \min_{\theta \in \{-1, +1\}} \lVert v(n^{-1}R_y) - \theta \hat{v} \rVert_{\infty} < \frac{1}{2 } \min \left\{m_e(n^{-1}R_y) , m_h(n^{-1}R_y) \right\}
    \right\}  ,
\end{equation}
\end{proposition}
The proof of the above proposition is given in Appendix \ref{appendix:perfect clustering condition proof}

\subsection{Proof of Theorem \ref{thm:cluster_crowdsourcing}: Perfect Clustering}\label{Appendix: proof of thm: perfect clustering}
Now we complete the proof of the clustering Theorem \ref{thm:cluster_crowdsourcing}.
From Proposition \ref{prop:conditions for perfect clustering}, we know that, 
\begin{equation}
    \mathbb{P}\left(\eta = 0 \right) \geq \mathbb{P}\left(
        \min_{\theta \in \{-1, +1\}} \lVert v(n^{-1}R_y) - \theta \hat{v} \rVert_{\infty} < \frac{1}{2 } \min \left\{m_e(n^{-1}R_y) , m_h(n^{-1}R_y) \right\}
    \right)   .
\end{equation}
Now we show that right hand side of the above equation is close to 1 for large values of $n$ using Lemma \ref{lemma :l inf concentration}. We also derive the corresponding necessary conditions on the problem parameters $n$ and $d$.

One requirement of Lemma \ref{lemma :l inf concentration} is that $ C_2 (\min(\alpha,1-\alpha))^2 \nu(n^{-1}R_y)d - 1 >0$. This leads to the following requirement on $d$:
\begin{equation}\label{eq: 1st requirement on d}
    d > \frac{1}{C_2 (\min(\alpha,1-\alpha))^2 \nu(n^{-1}R_y)}
\end{equation}
Under \eqref{eq: 1st requirement on d}, we have from Lemma \ref{lemma :l inf concentration}, for every $0 < \epsilon <( C_2 (\min(\alpha,1-\alpha))^2 \nu (n^{-1}R)d - 1)$,
    \begin{align}
        & \mathbb{P}\left(
\min_{\theta \in \{-1, +1\}} \lVert  \theta \hat{v} - v(n^{-1}R_y) \rVert_{\infty} \geq 
    C_3 \frac{(\epsilon  + 1)}{(\min (\alpha,1-\alpha))^2\nu(n^{-1}R_y) d\sqrt{d}} 
        \right) \leq 2d^2 \exp \left(-n\frac{\epsilon^2}{2d^2}\right) ,
    % \min_{\theta \in \{-1, +1\}} \lVert v(\mathbb{E}[T]) - \theta \hat{v}\rVert_\infty \leq 2^{3/2} \left(\lambda_1 (n^{-1} R_y) - \lambda_2 (n^{-1} R_y)\right)^{-1} + 3^{2} \cdot 2^{-2} \frac{\lambda_2 (n^{-1} R_y)}{\nu \sqrt{d}} .
    \end{align}
Next, we choose $\epsilon$ with  $0 < \epsilon <( C_2 (\min(\alpha,1-\alpha))^2 \nu (n^{-1}R)d - 1)$ such that 
\begin{equation*}
    C_3 \frac{(\epsilon  + 1)}{(\min (\alpha,1-\alpha))^2\nu(n^{-1}R_y) d\sqrt{d}}  \leq \frac{1}{2 } \min \left\{m_e(n^{-1}R_y) , m_h(n^{-1}R_y) \right\}
\end{equation*}
As, $C_2 = 2^{-8}$ and $C_3 = 3 \cdot 2^8$, the following value of $\epsilon$ satisfies the above requirement : 
\begin{equation*}
    \epsilon = \frac{1}{4C_3}(\min(\alpha,1-\alpha))^2 \nu(n^{-1}R_y)d \min(m_e(n^{-1}R_y)d^{\frac{1}{2}},m_h(n^{-1}R_y) d^{\frac{1}{2}},1)
\end{equation*}
when we impose : 
\begin{equation}\label{eq: 2nd requirement on d}
    d > \frac{4C_3}{ (\min(\alpha,1-\alpha))^2 \nu(n^{-1}R_y)\min \left\{m_e(n^{-1}R_y) d^{1/2}, m_h(n^{-1}R_y) d^{1/2}, 1 \right\}}
\end{equation}
Notice that the requirement on $d$ in equation \eqref{eq: 2nd requirement on d} is stronger than the requirement in equation \eqref{eq: 1st requirement on d}.
Putting it together, we get, when $d$ satisfies equation \eqref{eq: 2nd requirement on d} the perfect clustering is guaranteed as
\begin{align*}
   & \mathbb{P}\left(\eta = 0 \right) \geq \mathbb{P}\left(
        \min_{\theta \in \{-1, +1\}} \lVert v(n^{-1}R_y) - \theta \hat{v} \rVert_{\infty} < \frac{1}{2 } \min \left\{m_e(n^{-1}R_y) , m_h(n^{-1}R_y) \right\}
    \right) \\
    & \geq
     1-  4 d^2 \exp \left(
        - C_5 n \left(
            (\min(\alpha,1-\alpha))^2\nu(n^{-1} R_y)  \min \left\{m_e(n^{-1}R_y) d^{1/2}, m_h(n^{-1}R_y) d^{1/2}, 1 \right\} 
        \right)^2
    \right) 
\end{align*}
where, $C_4 = 4C_3$ and $C_5 = 2^{-4} C_3^{-2}$

When $r_e^{\top}r_h = 0$, we have $\min \left\{m_e(n^{-1}R_y) d^{1/2}, m_h(n^{-1}R_y) d^{1/2}, 1 \right\} \geq \min(\alpha,1-\alpha)$ from the analysis of Appendix \ref{Appendix: proof of lemma: spectral properties of R}. On the other hand when, $r_e^{\top}r_h \neq 0$, it is convenient to express the absolute margins $m_e(n^{-1}R_y)$ and $m_h(n^{-1}R_y)$ as a function of the ratio $s = \mu_e(n^{-1}R_y)/\mu_h(n^{-1}R_y)$ between the easy and hard magnitudes $\mu_e(n^{-1}R_y), \mu_h(n^{-1}R_y)$ so that
\begin{align}
    m_e(n^{-1}R_y)
    % =|v_e - \mu| 
    = \mu_e(n^{-1}R_y) - \mu(n^{-1}R_y)
    = \frac{d_h}{d}(\mu_e(n^{-1}R_y) - \mu_h(n^{-1}R_y)) = \frac{d_h}{d}\frac{||s| - 1|}{\sqrt{d_e s^2 + d_h}} \label{eq:m_e(n^{-1}R_y)}
\\
    m_h(n^{-1}R_y)
    % =|\mu - v_h| 
    = \mu(n^{-1}R_y) - \mu_h(n^{-1}R_y) = 
    \frac{d_e}{d}(\mu_e(n^{-1}R_y) - \mu_h(n^{-1}R_y))  = \frac{d_e}{d}\frac{||s| - 1|}{\sqrt{d_e s^2 + d_h}} \label{eq:m_h(n^{-1}R_y)} .
\end{align}
Hence, we can lower bound the term $\min  \{m_e(n^{-1}R_y) d^{1/2}, m_h(n^{-1}R_y) d^{1/2}, 1\}$ as follows:
\begin{align*}
    & \min  \{m_e(n^{-1}R_y) d^{1/2}, m_h(n^{-1}R_y) d^{1/2}, 1\}
     = \min  \left\{\frac{d_h}{d}\frac{||s|-1|}{\sqrt{d_e s^2+d_h}} d^{1/2}, \frac{d_e}{d}\frac{||s|-1|}{\sqrt{d_es^2+d_h}} d^{1/2}, 1\right\} \\
    & = \min \left\{\frac{\alpha ||s|-1|}{\sqrt{\alpha s^2+(1-\alpha)}},\frac{(1-\alpha) ||s|-1|}{\sqrt{\alpha s^2+(1-\alpha)}} ,1\right\} 
     \underbrace{\geq}_{(a)} \min \left\{\alpha,1-\alpha \right\} \frac{||s|-1|}{\sqrt{s^2+1}}
\end{align*}
where in $(a)$, we use the fact that $\min \left\{\alpha,1-\alpha \right\} \frac{||s|-1|}{\sqrt{s^2+1}} \leq 1$.
From the above bounds on $M$ and $\min  \{m_e(n^{-1}R_y) d^{1/2}, m_h(n^{-1}R_y) d^{1/2}, 1\}$, we can write the sufficient number of tasks required for perfect clustering as:
\begin{equation}
   d \geq \frac{C_4}{\sqrt{D(r_e,r_h, \alpha,d)}}
\end{equation}
 and the probability guarantee on perfect clustering as 
\begin{equation}
        \mathbb{P}(\eta=0)\geq 1- 2d^2\exp\left(-C_{5}nD(r_e,r_h, \alpha,d)\right),
    \end{equation}
    where the problem-dependent quantity $D(r_e,r_h, \alpha,d)$ characterizing the error exponent and the requirement on $d$ is given by    
    \begin{equation}
        D(r_e,r_h, \alpha,d) = \begin{cases}
         \left((\min(\alpha,1-\alpha))^3\frac{\nu(n^{-1}R_y)||s|-1|}{\sqrt{s^2+1}}\right)^2 & \text{when, } r_e^{\top}r_h \neq 0,\\
         \left((\min(\alpha,1-\alpha))^3\nu(n^{-1}R_y)\right)^2 & \text{when, } r_e^{\top}r_h = 0,
    \end{cases}
    \end{equation}

\section{Experiments}
% In this section, we investigate how the proposed clustering algorithm enhances the label estimation accuracy when crowdsourced labels behave according to the easy-hard model.
In this paper, we present experiments with real-world datasets, psuedo-real datasets and synthetic datasets to supplement the theory presented in the previous sections. By pseudo-real datasets, we mean the following: some real-world sets do not contain all the information we need to run our experiments and therefore, we generate some of the data we need using the available data in the datasets. In such cases, we will explain how we filled in the required data. Our experiments are presented in three subsections, with each subsection focusing on a different aspect of our model and theory:
\begin{enumerate}[leftmargin=*]
\item First, we compare our two-step algorithm (clustering tasks and then applying a DS algorithm to each type of task) with a single-step DS algorithm (i.e., applying a DS algorithm to all the tasks). Our experiments clearly show the benefit of clustering. Although our theoretical analysis primarily employs TE followed by WMV with NP-weights as the Dawid-Skene (DS) algorithm, we also compare DS algorithms with and without clustering across various other DS algorithms to demonstrate the benefits of clustering.
    \item Next, we compare our algorithm with other algorithms that also consider tasks of different types. We demonstrate that our algorithm performs better on most of the datasets considered.
\end{enumerate}
The datasets we used for our experiments are the following:
\begin{enumerate}[leftmargin=*]
    \item Two of the real-world datasets we used called the ``Bluebird'' (\citet{Bluebird}) and ``HC-TREC''(\citet{Buckley2010OverviewOT}) are complete datasets
     i.e., they have all the information we need such as worker-task responses for all worker-task pairs and the responses are binary-valued. So we used the data in these two datasets without any modification. 
    \item Four other real-world datasets, ``Dog''(\citet{Dog}), ``Duck''(\citet{Bluebird}), ``Temp''(\citet{snow08}) and ``RTE'' (\citet{snow08})) are sparse datasets which do not provide responses corresponding to all worker-task pairs as in our motivating example in the introduction. To handle this, for the ``Dog'' dataset that contains 4 classes, we 
    converted it to binary groups $\{0, 2\}$ vs. $\{1, 3\}$ following \citet{BonaldCombes}. Then 
    we calculate the fraction of correct labels (given by workers) for each task based on the ground truth and the available responses and classify half of them (the half with the most accurate worker responses) as easy tasks and the rest as hard tasks. Then, we estimate the empirical reliabilities of the workers for each type of task and use this to generate synthetic entries for the missing worker-task pairs in the response matrix. Similar treatments for the no-response entries is done for ''Duck'', ``RTE'' and ``Temp'', each of which contains binary truth values. Thus, the actual datasets that we use for ``Dog'', ``temp'' and ``RTE'' are pseudo-real datasets. The number of workers and tasks for all six datasets (``Bluebird'', ``HC-TREC'', ``Dog'', ``Temp'', ``Duck'', and ``RTE'') are provided in Table \ref{tab:combined}.
    \item Obtaining real world crowdsourcing datasets for healthcare examples that we mention in the Introduction is difficult due to privacy reasons. With the limited information available from a radiology dataset, we created a synthetic dataset and we report the results from the dataset in the appendix \ref{appendix:additional experiments}
\end{enumerate}
\subsection{Comparison with Traditional DS Algorithms}
Since the literature on crowdsourcing is vast, there are many algorithms available in the literature for the original DS model. From these algorithms, we select a few algorithms which we choose as the baseline for our experiments.

\textbf{Baseline Algorithms:} 
We consider the following Dawid-Skene algorithms in our experiments: unweighted majority vote (MV), ratio of eigenvectors (ER, \citealt{Dalvi}), TE (\citet{BonaldCombes}), and Plug-in gradient descent (PGD, \citealt{PluginGD}).
Then we compare their performances when applied separately to each task type clustered using Algorithm \ref{alg:cluster} to demonstrate the importance of separating tasks by type. A large number of algorithms have been proposed for crowdsourcing including Spectral-EM (\citet{SpectralEM}), and message-passing (\citet{KOS}) to name just a few. 
Exhaustively comparing with all the algorithms is difficult, so we have chosen to compare our algorithms to ER, TE, and PGD for the following reason: many algorithms have been compared in \citet{Dalvi}, \citet{BonaldCombes} and \citet{PluginGD}, where it was shown that ER, TE and PGD consistently out-perform other algorithms.

\begin{table}[th]
    \centering
    \begin{tabular}{cc}
        \begin{subtable}[t]{0.45\textwidth}
            \centering
            \begin{tabular}{lcc}
                \toprule
                Dataset & \# Workers & \# Tasks \\
                \midrule
                Bluebird & 39 & 108 \\
                Dog & 78 & 807 \\
                RTE & 164 & 800 \\
                Ducks & 53 & 240 \\
                HC-TREC & 10 & 1000 \\
                Temp & 76 & 462 \\
                \bottomrule
            \end{tabular}
            %\caption{Dataset }
            \label{tab:sparse_description}
        \end{subtable}
        &
        \begin{subtable}[t]{0.55\textwidth}
            \centering
            \begin{tabular}{lcccc}
                \toprule
                Dataset & MV & ER & TE & PGD \\
                \midrule
                Bluebird-TA & 24.07 & 27.78 & 17.59 & 25.93 \\
                Bluebird-C & 24.07 & 11.11 & 12.96 & 12.96 \\
                Gain & 0.00 & 16.67 & 4.63 & 12.97 \\
                \hline
                Dog-TA & 26.15 & 19.85 & 13.64 & 19.01 \\ 
                Dog-C & 26.15 & 0.78 & 12.23 & 20.56 \\
                Gain & 0.00 & 19.07 & 1.41 & -1.64 \\
                \hline
                Duck-TA & 32.58 & 59.37 & 41.04 & 38.96 \\
                Duck-C & 32.58 & 24.33 & 41.67 & 32.58 \\
                Gain & 0.00 & 35.04 & -0.6 & 6.38 \\
                \hline
                HC-TREC-TA & 33.70 & 68.80 & 67.30 & 30.80 \\
                HC-TREC-C & 33.70 & 40.90 & 30.60 & 40.80 \\
                Gain & 0.00 & 27.90 & 36.6 & 0.00 \\
                \bottomrule
            \end{tabular}
            %\caption{Label estimation errors for different crowdsourced datasets.
            %``TA'' and ``C'' indicate that labels were estimated without (task-agnostic) or with clustering.}
            \label{tab:sparse_results}
        \end{subtable}
    \end{tabular}
    \caption{Dataset Descriptions and Label Estimation Errors}
    \label{tab:combined}
\end{table}
\begin{comment}
    The performance of crowdsourcing algorithms with and without our clustering algorithm is shown in Table \ref{tab:jsrt_results} for the medical imaging dataset.
As shown, separation consistently increases accuracy over Dawid-Skene algorithms.
Because experts labeled the JSRT dataset, we observe a high accuracy using the simple majority vote.
However, failing to identify nodules can be consequential and even a small gain in accuracy is critical.
\end{comment}

Label estimation errors for the datasets considered are shown in Table \ref{tab:combined}.
We observe that clustering improves performance in most cases. In the case of RTE and Temp datasets, with or without clustering, the accuracy of label estimation is 100\%, that is why we did not include them in the Table \ref{tab:combined}. Hence, our results show that clustering does not hurt the accuracy even in cases where it may not be required.
\begin{figure*}[t]
    \centering
    \begin{subfigure}[b]{0.3\textwidth}
        \includegraphics[width=\textwidth]{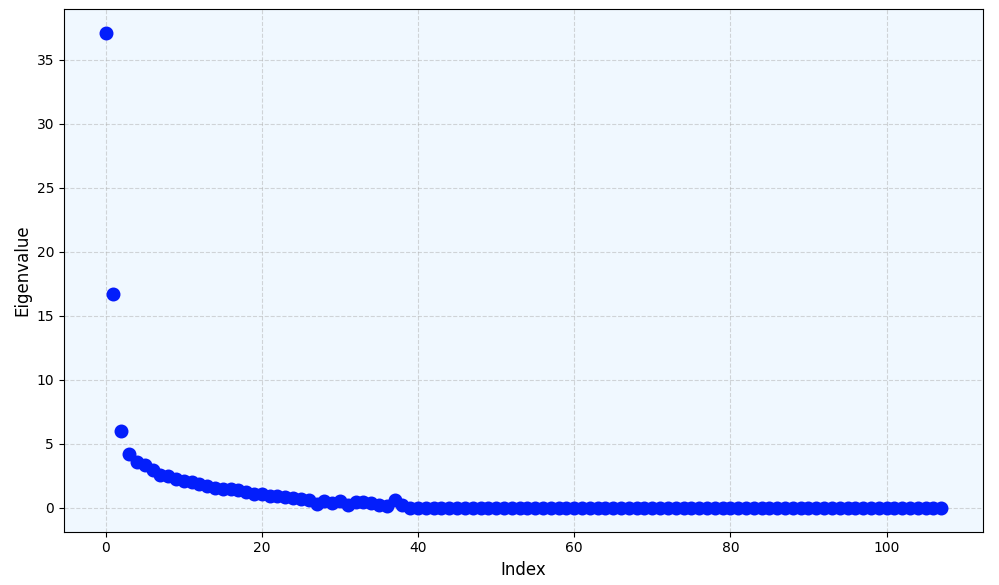}
        \caption{Bluebird Eigenspectrum}
        \label{fig:bluebird}
    \end{subfigure}
    \hfill
    \begin{subfigure}[b]{0.3\textwidth}
        \includegraphics[width=\textwidth]{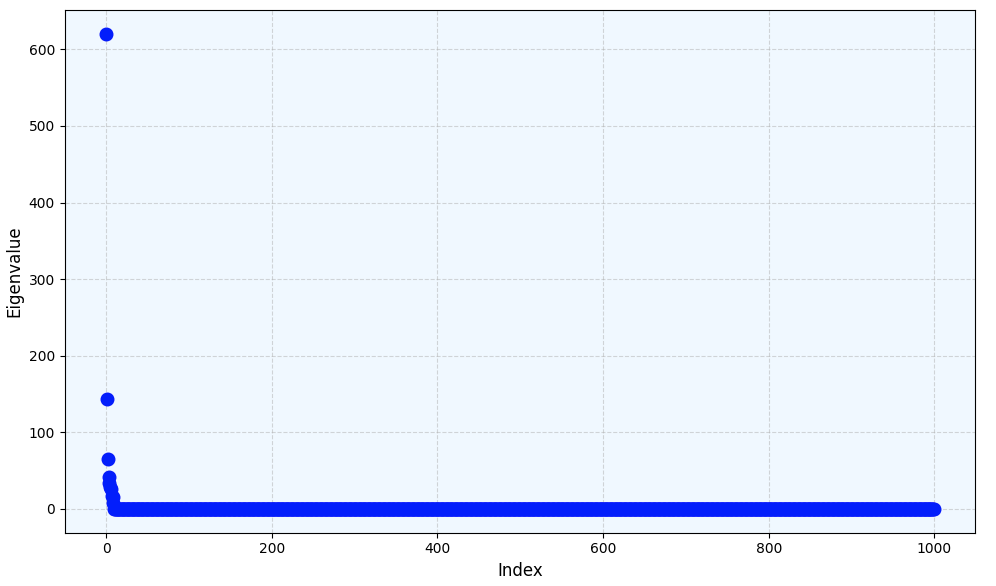}
        \caption{TREC Eigenspectrum}
        \label{fig:trec}
    \end{subfigure}
    \hfill
    \begin{subfigure}[b]{0.3\textwidth}
        \includegraphics[width=\textwidth]{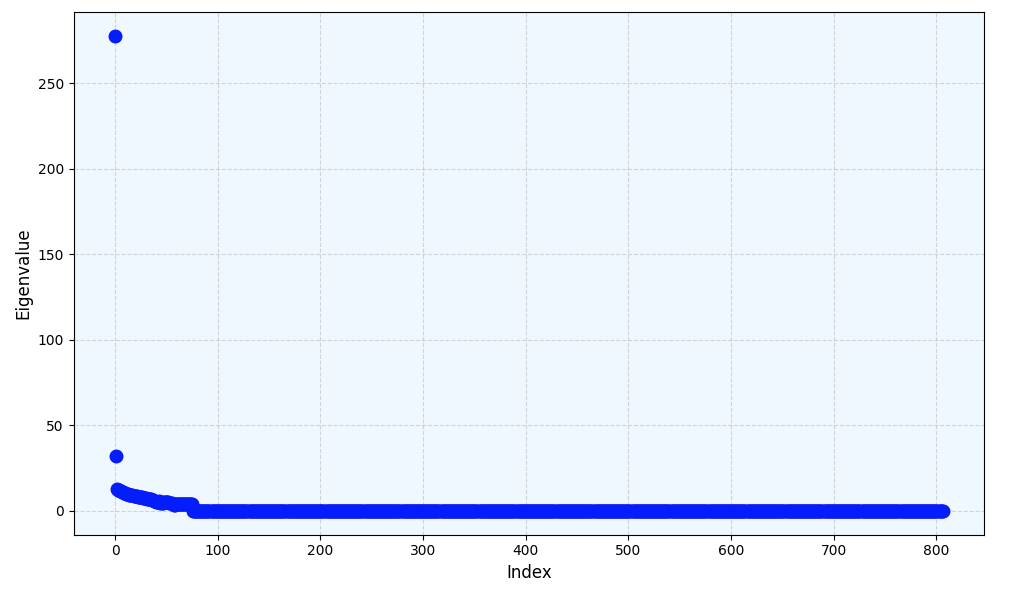}
        \caption{Dog Eigenspectrum}
        \label{fig:dog}
    \end{subfigure}

    \begin{subfigure}[b]{0.3\textwidth}
        \includegraphics[width=\textwidth]{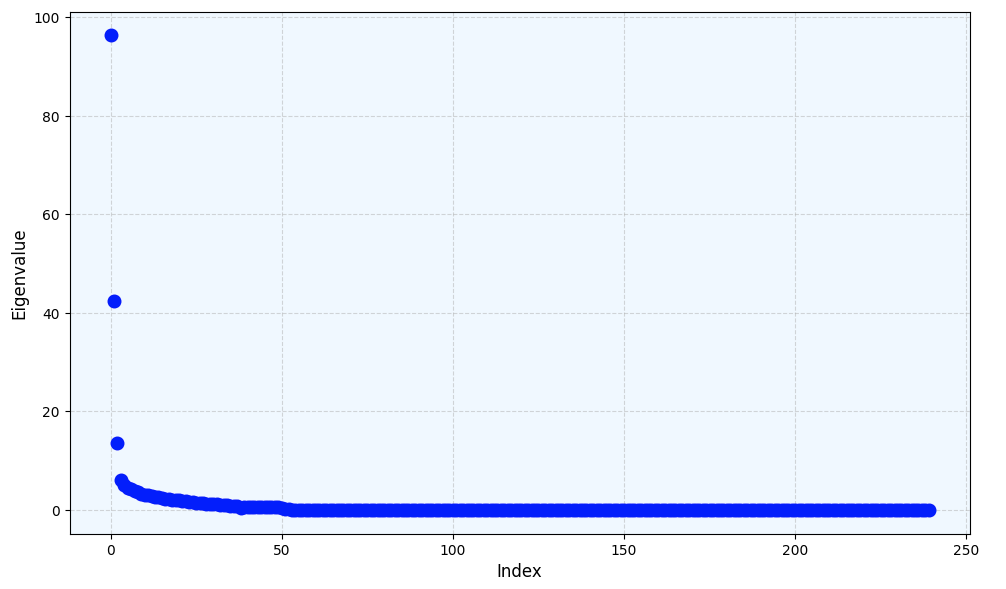}
        \caption{Duck Eigenspectrum}
        \label{fig:ducks}
    \end{subfigure}
    \hfill
    \begin{subfigure}[b]{0.3\textwidth}
        \includegraphics[width=\textwidth]{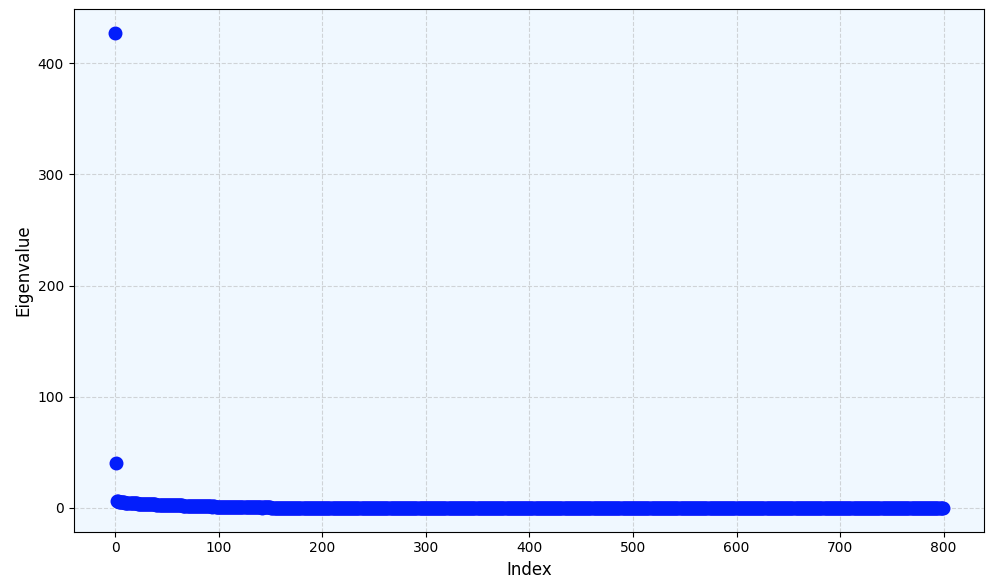}
        \caption{RTE Eigenspectrum}
        \label{fig:rte}
    \end{subfigure}
    \hfill
    \begin{subfigure}[b]{0.3\textwidth}
        \includegraphics[width=\textwidth]{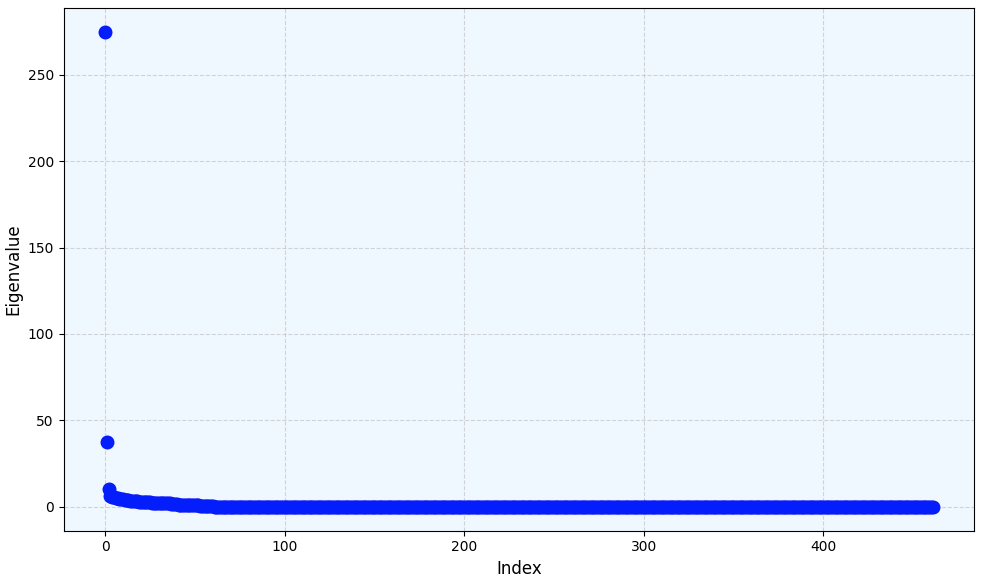}
        \caption{Temp Eigenspectrum}
        \label{fig:temp}
    \end{subfigure}

    \caption{Eigenspectrum of $T$ for different datasets: (a) Bluebird, (b) TREC, (c) Dog, (d) Duck, (e) RTE, and (f) Temp. For each plot, the y-axis represents the eigenvalues, and the x-axis represents the corresponding index of each eigenvalue. }
    \Description{Eigenspectrum of $T$ for different datasets}
    \label{fig:eigenvalues}
\end{figure*}
An interesting is question to understand is why clustering performs better than directly applying a DS algorithm to each of the datasets. To answer this question, we plotted the eigenspectrum of the matrix $T$ in Figure \ref{fig:eigenvalues}. As we can see, all the datasets exhibit at least two eigenvalues which are larger than the rest of them which are close to zero, thus indicating that there is more than one type of task. Therefore, clustering helps to separate tasks by their reliabilities.

\begin{comment}
     \textbf{Eigenspectrum of the Task-similarity matrix in the Real-world Datasets}\\
{\color{red}(modify this)} In this paper, we consider the hard-easy model, where the task-similarity matrix $T = n^{-1}O^{\top}O$ is a small perturbation of a rank-2 matrix.  In contrast, the Dawid-Skene (DS) model's task-similarity matrix is a small perturbation of a rank-1 matrix. Figure \ref{fig:eigenvalues} shows the eigenvalue of $T$ for four datasets: Bluebird, Duck, Temp, RC-TRCE. We observe that each dataset has more than one significant eigenvalue. This suggests that clustering tasks into different types before estimating the labels for each type is likely to be more beneficial than applying Dawid-Skene algorithms without clustering on these datasets. 
\end{comment}

\subsection{Comparison with Task-specific Reliability Models}
As discussed in the related work section, several previous papers address models with multiple types of tasks and use different task-specific reliability models to infer task labels. Notable works include \citet{KhetanOh}, \citet{OBIWAN}, \citet{Graphon}, \citet{Kaist} to name a few. The model in \citet{KhetanOh} assumes that $E(T)$ is a rank-1 matrix. Clearly, this is not true in all the datasets we have considered as shown in Figure \ref{fig:eigenvalues}. The algorithm in \citet{OBIWAN} involves a large number of parameters, leading to very poor performance on the datasets we used, therefore we are not comparing it with our model. Thus, we restrict the comparison of our algorithm to those in \citet{Graphon} and \citet{Kaist}.  

\begin{table}[th]
\centering
        \begin{tabular}{lccccccc}
        % \begin{tabular}{llccccc}
        \toprule
        Dataset & TE-C & SDP-2 & SDP-3 & SDP-4 & SS-2 & SS-3 & SS-4  
        % $\norm{r_1}^2 - \norm{r_2}^2$ & $r_1^T r_2$
        \\
        \midrule
        Bluebird & 12.96 & 24.81 & 44.07 & 40.25 & 25.68 & 27.9 & 22.62 \\
        Dog & 12.23 & 34.56 & 39.41 & 39.18 & 51.70 & 74.85 & 74.65  \\
        % min_reports = 10.
        Duck & 41.67 & 19.88 & 22.17 & 23.08 & 56.58 & 68.67 & 75.12  \\
        TREC & 30.6 & 48.94 & 42.95 & 38.22 & 49.39 & 66.49 & 75.7 \\
        Temp & 0 & 1.93 & 19.96 & 31.28 & 50.35 & 67.67 & 74.26
        % min_reports = 20
        % Dog & 69 & 807 & 15.3 
        % 78	807	0.126251072	46.65737727	1.480151111
        \\
        \bottomrule
        \end{tabular}  
     \caption{Comparison of our approach with Task-specific Reliability Models. `TE-C' is our two step approach - clustering followed by TE-WMV; `SDP-g' and `SS-g' are the algorithms from \citet{Kaist} and \citet{Graphon} considering g-type specialization model.}
     \label{tab:task-specific comparison}
\end{table}
Table \ref{tab:task-specific comparison} shows the comparison of our algorithm with the SDP-based algorithm in \citet{Kaist}(alg3 in that paper) and the Subset selection algorithm(SS) in \citet{Graphon}. Both the SPD and SS algorithm requires an input: the number of specializations considered in their g-type specialization model.
In the Table \ref{tab:combined}, the column `TE-C', `SDP-g' and `SS-g' correspond to our two-step approach, SDP-based algorithm in \citet{Kaist} with g number of specializations and SS algorithm from \citet{Graphon} with g number of specializations, respectively. We used the MATLAB code provided by the authors in \citet{Kaist} for running different `SDP' and `SS' algorithms. We see that our algorithm outperforms SDP and SS in the following datasets: ``Bluebird'', ``TREC'', ``Dog'', ``temp''. The "ducks" dataset is an exception: here, `SDP-2' beats `TE-C'. We believe that the reason for this is the following: 'SDP-2' uses majority voting among the matched workers instead of weighted majority voting as in `TE-C'. It is widely reported in the prior literature that weighted majority voting is better than simple majority voting for most crowdsourcing applications. However, there are datasets for which this is not true, the ``Duck'' dataset is one such example. as can be seen from the `Duck' entry in Table \ref{tab:combined}.  

\section{Conclusion}
We considered a crowdsourcing model which is more appropriate than the Dawid-Skene model when there are tasks that require different levels of skill sets.
Then we described a spectral clustering algorithm that clusters tasks by difficulty and analyzed its performance in clustering and its performance in label estimation when combined with TE and NP-WMV for each task type separately. Experiments with real-life datasets demonstrate the benefits of our algorithm.

% \label{fig:phase_transition}
 % based on crowdsourced radiology data

% \section{Proof Sketch}
% To derive the clustering error in Theorem \ref{thm:cluster}, we prove the following lemma.
% \begin{lemma}
%     Let $\hat{v}$ be the principal eigenvector of the task-similarity matrix $T = \frac{1}{n} Y^T Y$, and define $\mu = \frac{1}{d} \sum_j \abs{v}_j$ to be the average over entries of the principal eigenvector corresponding to the expected task-similarity matrix ${\mathbb E} T$.
%     With probability $\geq 1- 1/d$,
%     \begin{equation}
%         \frac{1}{d}\sum_j \indicator\left\{ \abs{\hat{v}_j - \mu} = \mathcal{O}\left(
%             \frac{1}{n} \sqrt{\frac{\log d}{\Delta d}}
%             \right\}
%         \right)
%     \end{equation}
%     where $\Delta$ is the problem-parameter defined in Eq. \eqref{eq:Delta}.
% \end{lemma}

% {
% \color{red}
% Also see if 
% }

%\begin{acks}
%To Robert, for the bagels and explaining CMYK and color spaces.
%\end{acks}

\bibliographystyle{ACM-Reference-Format}
\bibliography{example_paper}

\appendix

\section{Additional Experiments : Synthetically Generated Radiology Data}\label{appendix:additional experiments}
Obtaining real-world datasets for healthcare examples mentioned in the Introduction is difficult. Due to privacy reasons, such datasets do not contain much of the information we require, including ground truths and responses. Nevertheless, we considered one radiology dataset: the Japanese Society of Radiological Technology (JSRT) Database and its report \citep{JSRT} to conduct a synthetic experiment. These datasets only contain information about the reliability of the doctors who looked at the data. In other words, this dataset only provides a range of realistic reliabilities, but we had to generate synthetic ground truths and response matrices.
\subsection{Setup}
In this subsection, we describe how we generate our synthetic datasets from JSRT report in \citet{JSRT}.
\begin{table}[th]
        \centering
        \begin{tabular}{lcccccc}
        \toprule
        % Degree of Subtlety & Obvious & Relatively obvious & Subtle & Very subtle & Extremely subtle \\
        Subtlety & 0 & 1 & 2 & 3 & 4 & 5 \\
        \midrule
        Count & 93 & 25 & 29 & 50 & 38 & 12 \\
        Size & 
        % 23.0 $\pm$ 9.03 & 17.9 $\pm$ 6.05 & 17.2 $\pm$ 8.69 & 16.4 $\pm$ 5.89 & 14.6 $\pm$ 6.69
        0.0 & 23.0 & 17.9 & 17.2 & 16.4 & 14.6 \\
        Mean sensitivity(accuracy) of experts & 80.9 & 99.6 & 92.6 & 75.7 & 54.7 & 29.6
        \\
        \bottomrule
        \end{tabular}
       % \caption{First Week}
  \caption{JSRT dataset. Size is in millimeters, and a subtlety of 0 indicates that a nodular pattern is absent.}
   \label{tab:jsrt_summary}
\end{table}
The JSRT report contains the performance of 20 radiologists for identifying solitary pulmonary nodules in chest radiographs.
Its dataset statistics are summarized in Table \ref{tab:jsrt_summary}.
Expert performances are reported for various levels of subtlety defined by the size of nodular patterns.
It is clear that detecting nodular patterns becomes significantly more difficult as the size is decreased, demonstrating a multi-type phenomenon with varying levels of task difficulty.
Our setup for the JSRT experiments are as follows.
There is a total of 6 types according to the mean sensitivity reported across all radiologists for 6 different subtlety levels. 
These values are used as the accuracy for each type as described next. 
\begin{enumerate}[leftmargin=*]
    \item For the JSTR-6 data, we use the reported means and standard deviations of sensitivities of a type $k \in [6]$: $(\tilde{r}_k,\sigma_k)$ as: for each type, we sample the probability parameter for each worker $i$, $p_{ki}$ as a sample from the uniform distribution with support $\tilde{r}_k \pm \sigma_k$. Then we set $r_{ki}= \frac{p_{ki}+1}{2}$. 
    \item To get an easy-hard model from this, we generate the dataset JSRT-2. Here, we combine the higher and lower 3 accuracy parameters as: for the easy type, the sensitivity is estimated as having mean of $\frac{1}{3}\sum_{k=1}^{3}\tilde{r}_k$ and standard deviation as the root mean square of the standard deviation of the first three subtlety levels. The parameters for the hard types are generated similarly from the next 3 subtlety levels.
\end{enumerate}

% , where $\bar{r}_k, \sigma_k$ are the reported means and standard deviations for each type $k$.
Each truth value $y_j$ is drawn randomly from its class-distribution defined by the sample mean of positive (presence of nodules) cases.
We then sample the crowd's response following the number of tasks per type in Table \ref{tab:jsrt_summary}.

\subsection{Results}
\begin{table}[th]
\begin{center}
\begin{sc}
% \begin{tabular}{lccccc}
\begin{tabular}{lcccc}
\toprule
Dataset & MV & ER & TE & PGD 
\\
\midrule
JSRT-2-TA & 5.65 & 5.65 & 4.74 & 5.06 \\
JSRT-2-C & 5.65 & 4.39 & 3.16 & 3.81 \\
Gain     & 0.00 & 1.26 & 1.58 & 1.25 \\ 
\hline
% \midrule
JSRT-6-TA & 10.30 & 10.30 & 9.96 & 9.72 \\
JSRT-6-C & 10.30 & 10.02 & 9.84 & 9.76 \\
Gain     & 0.00  & 0.28  & 0.12 & -0.04
% \hline 
% \midrule
% Bird-U & 24.07 & x & 28.70 & 17.59 & 25.00 \\
% Bird-S & 24.07 & x & 11.11 & 12.96 & 19.44 \\
% Gain & 0.00 & x & 16.59 & 4.63 & 5.56 
\\
\bottomrule
\end{tabular}
\end{sc}
\end{center}
\caption{Label estimation errors (\%) for the JSRT experiments. 
    ``TA'' and ``C'' after dataset names indicate whether label estimation was performed without(type agnostic) or with clustering, respectively.
}
\label{tab:jstr_results}
\end{table}
The performance of crowdsourcing algorithms with and without our clustering algorithm on the JSRT-6 and JSRT-2 datasets is shown in Table \ref{tab:jstr_results}.
As shown, separation consistently increases accuracy over Dawid-Skene algorithms.
Because experts labeled the JSRT dataset, we observe a high accuracy using the simple majority vote.
However, failing to identify nodules can be consequential and even a small gain in accuracy is critical.

\begin{comment}
\subsection{Numerical Analysis of Clustering Error}

We proved in Theorem \ref{thm:cluster_crowdsourcing} that the error due to clustering depends on the product of $\nu(n^{-1}R_y)$ and $\frac{|s-1|}{\sqrt{s^2+1}}$ .
Figure \ref{fig:cluster} shows through numerical experiments that the error bound is tight for $d_e = d_h$ in these problem-dependent parameters.

\begin{figure}[h]
\vskip 0.2in
\begin{center}
% \centerline{\includegraphics[width=\columnwidth, height=4.5cm]{icml_numpapers}}
\centerline{\includegraphics[width=0.45\textwidth, height=4.5cm]{figs/cluster_error_plot.png}}
\caption{A line of best fit comparing Monte Carlo simulations of the clustering error $(n=50, d=200)$ with our error bound in \eqref{eq:perfect clustering prob} adjusted by a constant multiplicative factor. This figure demonstrates that our theoretical analysis is tight in $s$ and $\nu(n^{-1}R_y)$.
}
\label{fig:cluster}
\end{center}
\vskip -0.2in
\end{figure}

\end{comment}

\section{Proof of Proposition \ref{prop:WMV UB}: Error Rate of Type-Agnostic Weighted Majority Vote} \label{proof: prop: WMV UB}
Recall the label estimation error $P_{av}(w) =  \frac{1}{d} \sum_j {\mathbb P}\left(\hat{y}_j^{WMV}(w) \neq y_j\right)$ defined in equation \eqref{eq:wmv_error}. 
We drop the task subscript and write $k_j = k$. We also drop the superscript `WMV' in this section from $\hat{y}_j^{WMV}(w)$.
For each worker $i$ and task $j$, let $G_{ij}$ be a random variable which takes the value $+1$ if worker $i$ correctly labels task $j$ and is $-1$ otherwise. In other words, $G_{ij}=y_jX_{ij}$, which is $+1$ with probability $\frac{1}{2} (1 + r_{ki})$.
\begin{equation*}
\mathbb{P}_k\left(\hat{y}_j(w) \neq y_j\right) \leq \mathbb{P}_k\left(\sum_{i=1}^n w_i G_{ij} \leq 0\right).
\end{equation*}
For any $t > 0$, we bound the error probability using Markov's inequality on the moment generating function as
\begin{equation}
    \mathbb{P}_k\left(\sum_{i=1}^n w_i G_{ij} \leq 0\right) \leq 
    \mathbb{P}_k\left(\exp\left(-t\sum_{i=1}^n w_i G_{ij}\right) \geq 1\right)
    \leq \mathbb{E}\left( \exp\left(-t\sum_{i=1}^n w_i G_{ij}\right) \right)
\end{equation}
Then for $t\geq 0$, we can write:
\begin{equation*}
    \mathbb{P}_k\left(\sum_{i=1}^n w_i G_{ij} < 0\right) \leq \min_{t \geq 0}\mathbb{E}\left( \exp\left(-t\sum_{i=1}^n w_i G_{ij}\right) \right)
\end{equation*}
Because workers respond independently of each other, $G_{1j}, \dots, G_{nj}$ are mutually independent and
% From the mutual independence of the responses of different workers, 
\begin{align*}
    \min_{t \geq 0}\mathbb{E}\left( \exp\left(-t\sum_{i=1}^n w_i G_{ij}\right) \right) & = \min_{t \geq 0}\prod_{i=1}^n \mathbb{E}\left( \exp(-t w_iG_{ij})\right)\\
    & = \min_{t \leq 0}\prod_{i=1}^n \left( \frac{1}{2}e^{t w_i}(1-r_{ki})+\frac{1}{2}e^{-t w_i }(1+r_{ki})\right)\\
    & = \exp(-n \varphi(w,r_k)) ,
\end{align*}
where $\varphi(w,r_k)$ is defined in equation \eqref{eq:varphi}, 
% \begin{equation*}
%     \varphi(w,r_k) = -\min_{\lambda > 0} \frac{1}{n} \sum_{i=1}^n log\left( \frac{1}{2}e^{\lambda w_i}(1-r_{ki})+\frac{1}{2}e^{-\lambda w_i} (1+r_{ki})\right)   
% \end{equation*}
% For k = 2, this leads to 
Therefore,
\begin{align*}
    P_{av}(w) & \leq \frac{1}{d} \left(\sum_{j:k_j=e} \exp\left(-n\varphi(w,r_e)\right) + \sum_{j:k_j=h} \exp\left(-n\varphi(w,r_h)\right) \right) \\
    & = \frac{1}{d} \left(d_e \exp\left(-n\varphi(w,r_e)\right) +d_h \exp\left(-n\varphi(w,r_h)\right) \right)\\
    & \leq \exp\left(-n \min_{k} \varphi(w,r_k)\right)
\end{align*}

\section{Proof of Proposition \ref{thm:WMV}: Error Rate Lower Bound of Task Agnostic Weighted Majority Vote}
\label{appendix:proof thm:WMV }
Let us first fix a task index $j$ and let the type of that task be $k_j=k$ for some $k\in \{e,h\}$. Defining $G_{ij}$ as in Appendix \ref{proof: prop: WMV UB},
% From the probabilistic model of workers labeling j-th task, we can write,
\begin{equation*}
\mathbb{P}_k\left(\hat{y}_j(w) \neq y_j\right) \geq \mathbb{P}_k\left(\sum_{i=1}^n w_i G_{ij} < 0\right),
\end{equation*}
where we drop the superscript `WMV' in this section from $\hat{y}_j^{WMV}(w)$.
We notice that $\sum_{i=1}^n w_i G_{ij}$ can only take finitely many values with the maximum being lower bounded by $\sum_{i}w_i$. 
Consider the set $\mathbb{S}= \{s : s = \sum_{i}g_i, g_i \in \{-w_i,w_i\}\}$. 
For any positive value of $S_k$ with $0<S_k \leq \sum_{i}|w_i| $,
% We define the following random variables : $H(i) = \lambda w(i)T(i)$, $S_n = \sum_{i=1}^nH(i)$ for some $\lambda > 0$. Then,
\begin{align}
    % \mathbb{P}(\sum_{i=1}^n w(i)T(i) > 0) = \mathbb{P}(S_n > 0) \geq \sum_{ 0 < s < L} \mathbb{P}(S_n = s) \geq \sum_{0 < s < L} \sum_{h_i : \sum_{h_i}=s} \prod_{i=1}^{n}\mathbb{P}(H_i = h_i)
    \mathbb{P}_k\left(\sum_{i=1}^n w_i G_{ij} < 0\right) & = \mathbb{P}_k\left(
        - \sum_i w_i G_{ij} > 0 
    \right)
     \geq 
    \sum_{s\in \mathbb{S}: 0<s<S_k} {\mathbb P}_k\left(- \sum_i w_i G_{ij} = s\right) \\
    \label{eq:weightedG}
    & = 
    \sum_{s\in \mathbb{S}: 0<s<S_k} \sum_{\sum_i g_i = s,g_i \in \{-w_i,w_i\}} \prod_{i=1}^n {\mathbb {P}_k}\left(- w_i G_{ij} = g_i\right) 
\end{align}
holds by the independence of workers.
Now, we use a change of measure of the underline random variable. Lets say a new random variable corresponding to each $i$ is $\tilde{G}_{ij}$ given by the following mass distribution for some $t_n(k) \geq 0$
\begin{align*}
    Q_{k} \left(\tilde{G}_{ij} = 1\right) = \frac{(1 + r_{ki})e^{- t_n(k)w_i}}{(1 + r_{ki}) e^{-t_n(k)w_i} + (1 - r_{ki}) e^{t_n(k)w_i}} , \\
    Q_{k} \left(\tilde{G}_{ij} = -1\right) = \frac{(1 - r_{ki}) e^{t_n(k)w_i}}{(1 + r_{ki}) e^{-t_n(k)w_i} + (1 - r_{ki}) e^{t_n(k)w_i}}  .
\end{align*}
Its joint distribution over workers $i \in [n]$ is written as $Q_k$, and Eq. \eqref{eq:weightedG} is expressed as
\begin{align*}
    \sum_{s\in \mathbb{S}: 0<s<S_k} \sum_{\sum_i g_i = s,g_i \in \{-w_i,w_i\}} \prod_{i=1}^n {\mathbb {P}}\left(- w_i G_{ij} = g_i\right) \\ 
    \geq 
    Q_k \left(0 < - \sum_i w_i \tilde{G}_{ij} < S_k\right)  \frac{\prod_{i=1}^n\left((1 + r_{1i}) e^{-t_n(k)w_i} + (1 - r_{1i}) e^{t_n(k)w_i}\right)}{2 e^{S_k}}
\end{align*}
% Using the definition in Eq. \eqref{eq:varphi} 
where, to obtain the last step above, we have multiplied and divided each term in the product by $\frac{2 e^{g_i}}{(1 + r_{1i}) e^{-t_n(k)w_i} + (1 - r_{1i}) e^{t_n(k)w_i}}$ and used the bound $ \sum |w_i| \leq S_k.$
Recall the expression
\begin{equation*}
    \varphi_n (w, r_k) = - \min_{t \geq 0} \frac{1}{n} \sum_i \log \left(\frac{1}{2} \left( 
        (1 + r_{ki}) e^{-t w_i} + (1 - r_{ki}) e^{tw_i}\right)\right),
\end{equation*}
Define $t_n(k) = t_n^*(k)$ to be a minimizing argument of $\varphi_n (w, r_k)$ in the domain $t_n(k) \geq 0$.
Now, putting the minimizing argument $t_n^*(k)$ in the place of $t_n(k)$ we obtain a lower bound for type $k$ as
\begin{equation}
    {\mathbb{P}_k} \left(\hat{y}_j \neq y_j \right)
    \geq 
    Q_k\left(0 < -\sum_i w_i \tilde{G}_{ij} < S\right) e^{- n \varphi_n (w, r_k) - S_k}.
\end{equation}
Noting that the distribution $Q_k(\tilde{G}_{i,j})$ is invariant to task index $j$, we drop the index $j$ in the following bound on the error rate for positive values $S_k, \forall k \in \{e,h\}$ (note that the following holds for all $y$):
\begin{equation}\label{eq:LBQ}
     {\mathbb P}_{av}(w) \geq \sum_{k \in \{e,h\}}\frac{d_k}{d}Q_k\left(0 < - \sum_i w_i \tilde{G}_{i} < S_k\right) e^{- n \varphi_n (w, r_k) - S_k} .
\end{equation}
To analyze this further, use the following Lemma on the distribution of $- \sum_i w_i \tilde{G}_{i}$, an extension to the asymptotic analysis of majority voting in \citet{Gao16}.

Recall our definition $\rho_k \leq \min_{i} \frac{1 + r_{ki}}{2} \leq 1 - \rho_k, \forall k \in \{1,2\}$.
The following lemma is similar to Lemma 6.3 in \citet{Gao16}. The proof is given next to it for completeness. 
\begin{lemma}\label{lem:63}
Let $\rho\leq \frac{1\pm r_{ki}}{2}\leq 1-\rho, \forall  i\in [n], k\in \{e,h\}$, for some  $\rho \in (0, 1/2)$. 
    \begin{enumerate}
        \item Let $t^*_n(k)$ be the minimizer of $\varphi_n(w,r_k)$ defined in equation
        \eqref{eq:varphi}. Then,\\ $0 \leq t_n^*(k) \leq - \frac{n}{\|w\|_1}\log \rho,  k\in \{e, h\}, \forall n\geq 1$. 
        \item For any $y\in \{\pm 1\}$ and any $t_n(k) \geq 0$,
                \begin{equation*}  
        \frac{\sum_{i=1}^n \left(-w_iG_i-\mathbb{E}_{Q_k}[-w_i\tilde{G}_i]\right)}{\sqrt{ \mathrm{Var}_{Q_k} (-\sum_{i=1}^nw_i\tilde{G}_i)}}  \xrightarrow[n\rightarrow \infty]{d}\mathcal{N}\left(0, 1\right) ,\ \ \ \ \ \ \text{under the measure}\ \ Q_k.
    \end{equation*}
Moreover, at $t_n(k)=t_n^*(k)$,
\begin{equation*}  
        \frac{-\sum_{i=1}^n w_i\tilde{G}_i}{\sqrt{ \mathrm{Var}_{Q_k} (-\sum_{i=1}^nw_i\tilde{G}_i)}}  \xrightarrow[n\rightarrow \infty]{d}\mathcal{N}\left(0, 1\right) ,\ \ \ \ \ \ \text{under the measure}\ \ Q_k.
    \end{equation*}   
     \end{enumerate}

\end{lemma} 
\begin{proof}
\begin{enumerate}[leftmargin=*]
   \item  Let 
    \begin{equation*}
        \beta_k(t_n(k)) = \prod_{i=1}^n\frac{1}{2}
\left[(1 + r_{ki}) e^{-t_n(k)w_i} + (1 - r_{ki}) e^{t_n(k)w_i}\right]. 
\end{equation*}
Then $\beta_k(0)=1$ and $\forall t_n(k) \geq -\frac{n}{\|w\|_1}\log\rho$, we have that $\beta_k(t_n(k)) \geq \prod_{i=1}^n\left(\rho e^{t|w_i|}\right)\geq 1$. Therefore,  $t^*_n(k)\in \left[ 0,- \frac{n}{\|w\|_1}\log \rho\right]$.

\item  For the second part, we use Lindeberg's condition for the Central Limit Theorem for the expression $\sum_{i=1}^n -w_iG_i$. The Lindeberg's condition in this context corresponds to
\begin{align*}
& \lim_{n\rightarrow \infty}\frac{\sum_{i=1}^n \mathbb{E}_{Q_k}\left[\left(-w_i\tilde{G}_i-\mathbb{E}_{Q_k}[-w_i\tilde{G}_i]\right)^2 \indicator\left\{\left|-w_i\tilde{G}_i-\mathbb{E}_{Q_k}[-w_i\tilde{G}_i]\right|>\epsilon\sqrt{\mathrm{Var}_{Q_k}\left(\sum_{i=1}^n -w_i\tilde{G}_i\right)}\right\}\right]}{\mathrm{Var}_{Q_k}(\sum_{i=1}^n -w_i\tilde{G}_i)}\\
& =0, \forall \epsilon>0.
\end{align*}
A direct calculation gives 
\begin{align*}
\mathbb{E}_{Q_k}[-w_i\tilde{G}_i] & \underbrace{=}_{(a)} w_i \frac{(1-p_{ki}) e^{t_n(k)w_i} -  p_{ki} e^{-t_n(k)w_i}}{(1-p_{ki}) e^{t_n(k)w_i} +  p_{ki} e^{-t_n(k)w_i}}\\
& = \frac{\frac{d}{dt_n(k)}\left[(1-p_{ki}) e^{t_n(k)w_i} + p_{ki} e^{-t_n(k)w_i}\right]}{(1-p_{ki}) e^{t_n(k)w_i} +  p_{ki} e^{-t_n(k)w_i}}\\
& = \frac{d}{dt_n(k)}\log\left((1-p_{ki}) e^{t_n(k)w_i} + p_{ki} e^{-t_n(k)w_i}\right),
\end{align*}
where in $(a)$, we used the following relation : $p_{ki} = \frac{1+r_{ki}}{2}, \forall k \in {e,h} , i \in [n].$

The last two equalities imply: at $t_n(k)=t_n^*(k)$, $\mathbb{E}_{Q_k}\left[\sum_{i=1}^n-w_i\tilde{G}_i\right]=0$. Moreover, $\mathbb{E}_{Q_k}[(-w_i\tilde{G}_i)^2]=w_i^2$.
Therefore, 
\begin{align*}
\mathrm{Var}_{Q_k}(-w_i\tilde{G}_i)&=w_i^2\left[1- \frac{[(1-p_{ki}) e^{t^*_n(k)w_i} -  p_{ki} e^{-t^*_n(k)w_i}]^2}{[(1-p_{ki}) e^{t^*_n(k)w_i} +  p_{ki} e^{-t^*_n(k)w_i}]^2}\right]\\
& =w_i^2\frac{4p_{ki}(1-p_{ki})}{[(1-p_{ki}) e^{t^*_n(k)w_i} +  p_{ki} e^{-t^*_n(k)w_i}]^2}\\&\geq \frac{4 w_i^2 \rho^2}{(1-\rho)^2[e^{t^*_n(k)w_i}+e^{-t^*_n(k)w_i}]}\geq \frac{2 w_i^2 \rho^2}{(1-\rho)^2e^{t^*_n(k)|w_i|}}\geq \frac{2 w_l^2 \rho^2}{(1-\rho)^2e^{t^*_n(k)w_u}},
\end{align*}
and hence, $\mathrm{Var}_{Q_k}(\sum_{i=1}^n -w_i\tilde{G}_i)\geq n  \frac{2 w_l^2 \rho^2}{(1-\rho)^2e^{t^*_n(k)w_u}}\rightarrow \infty$ as $n\rightarrow \infty$. 

Additionally, $\left|-w_i\tilde{G}_i-\mathbb{E}_{Q_k}[-w_i\tilde{G}_i]\right|\leq 2|w_i|\leq 2w_u$ almost surely (and therefore, $\mathrm{Var}_{Q_k}(-w_i\tilde{G}_i)\leq 4 w_u^2$). Thus, for every $\epsilon>0$ we have that 
\begin{equation*}
\indicator\left\{\left|w_i\tilde{G}_i-\mathbb{E}_{Q_k}[-w_i\tilde{G}_i]\right|>\epsilon\sqrt{\mathrm{Var}_{Q_k}\left(\sum_{i=1}^n -w_i\tilde{G}_i\right)}\right\}=0, \ \ \text{almost surely}
\end{equation*}
for $n>\frac{2w_u^2 (1-\rho)^2 e^{t^*_n(k)w_u}}{\epsilon^2 w_l^2 \rho^2}$. Lindeberg's condition now follows.
\begin{remark}
    We can see that $\mathrm{Var}_{Q_k}(-w_i\tilde{G}_i) > 0$ as $t^*_n(k) \leq -\frac{n}{\norm{w}_1}\log \rho$
\end{remark}
\end{enumerate}

\end{proof}

Now, let us go back to proving the lower bound. Setting $S_k = \sqrt{\mathrm{Var}_{Q_k} (\sum_i -w_i \tilde{G}_{ij})}$, we write the following 
\begin{align*}
    & Q_k\left(0 < - \sum_i w_i \tilde{G}_{i} < S_k\right) \\
    & =   Q_k\left(0 < - \sum_i w_i \tilde{G}_{i} < \sqrt{\mathrm{Var}_{Q_k} \left(\sum_i -w_i \tilde{G}_{ij}\right)}\right)\\
    & \underbrace{=}_b  Q_k\left(0 < \frac{- \sum_i w_i \tilde{G}_{i}}{ \sqrt{\mathrm{Var}_{Q_k} (\sum_i -w_i \tilde{G}_{ij})}} < 1\right).
\end{align*}
In $(b)$, we use the following fact from Lemma \ref{lem:63}: $\sqrt{\mathrm{Var}_{Q_k} \left(\sum_i -w_i \tilde{G}_{ij}\right)} > 0$ at $t_n(k) = t^*_n(k)$. 
Also, 
\begin{align*}
    \exp \left(- n \varphi_n (w, r_k) - S_k\right)  & = \exp \left(-n  \varphi_n (w, r_k) - \sqrt{\mathrm{Var}_{Q_k} \left(\sum_i -w_i \tilde{G}_{i}\right)}
    \right) .
\end{align*}
Evaluating $\mathrm{Var}_Q (\sum_i -w_i \tilde{G}_{i}) \leq \sum_i w_i^2$ and using the following bounds on the entries of $w$ : $w_l \leq w_i \leq w_u , \forall i \in [n]$ ,
\begin{align}
    \exp \left(-n  \varphi_n (w, r_k) - t_n^*(k) \sqrt{\mathrm{Var}_{Q_k} \left(\sum_i -w_i \tilde{G}_{i}\right)}
    \right) & \geq 
    \exp\left( - n\norm{w}_2
    - n \varphi_n(w, r_k)
    \right) \\
    & \geq \exp\left( - \sqrt{n}w_u
    - n \varphi_n(w, r_k)
    \right)
\end{align}
Putting it all together, We can write from equation \eqref{eq:LBQ},
\begin{align*}
    & {\mathbb P}_{av}(w) \\
    & \geq \sum_{k \in \{e,h\}}\frac{d_k}{d}  Q_k\left(0 < \frac{- \sum_i w_i \tilde{G}_{i}}{\sqrt{\mathrm{Var}_{Q_k} (\sum_i -w_i \tilde{G}_{ij})}} < 1\right) \exp\left( - \sqrt{n}w_u
    - n \varphi_n(w, r_k)
    \right)\\
    & \geq \min_k\frac{d_k}{d} \exp\left( -  \sqrt{n}w_u
    - n \min_k\varphi_n(w, r_k)
    \right)\min_k Q_k\left(0 < \frac{ \sum_i -w_i \tilde{G}_{i}}{ \sqrt{\mathrm{Var}_{Q_k} (\sum_i- w_i \tilde{G}_{ij})}} < 1\right)
\end{align*}
By first taking a minimum over weight vector $w$ and then taking the $\lim\inf$ as $n \to \infty$ and using Lemma \ref{lem:63},
\begin{equation*}
    \liminf_{n \to \infty} 
            \frac{1}{n} \log \min_{w} P_{av} (w) 
        \geq -\limsup_{n \to \infty}\max_{w}\min_{k}\varphi_n(w,r_k)  .
\end{equation*}

\section{Proof of Proposition \ref{prop:achievable_perfect}}
The statement is obtained by appropriately modifying Theorem 4.1 in \citet{Gao16} and Theorem 2 in \citet{BonaldCombes}. 
For the known type case, we separate the tasks according to their type and each type is dealt separately as two Dawid-Skene problem instances. 
Because task types are known for this setting, the TE algorithm is applied separately to each type independently of the other to estimate each type's reliability vectors.
Labels are estimated using the corresponding NP-WMV.

From Theorem 2 \footnote{According to the TE algorithm described in the section \ref{sec:DS algo}, the
 estimated reliabilities are projected onto the set $\rho \leq \frac{1+\hat{r}_{ki}}{2} \leq 1-\rho$. This step was not included in the original TE algorithm proposed by \citet{BonaldCombes}. Nevertheless, the concentration of the reliability estimate derived from Theorem 2 of \citet{BonaldCombes} in the max-norm sense also holds under this projection, as it acts as a contraction operator.} in \citet{BonaldCombes}\footnote{One difference between our model and the model considered in \citet{BonaldCombes} is that we consider the true labels as deterministic quantity and \citet{BonaldCombes} considers them to be random variables. The TE algorithm uses the worker-similarity matrix and we can easily show that the worker-similarity matrix is independent of the true labels and thus the performance bound on the TE algorithm in Theorem 2 in \citet{BonaldCombes} is valid for deterministic labels }, we have that if the number of workers $n$ satisfies
\begin{equation}\label{eq:number_of_workers_known_type}
    n^2 \geq \frac{3\rho}{\overline{r}}
\end{equation}
and the number of tasks $d_k$ per type $k \in \{e, h\}$ is
\begin{equation}\label{eq:number_of_tasks_known_type}
    d_{k} \geq \max\left(120 \times 24^2 \frac{n^2}{V_{k}^4\rho^2}(n\Phi(r_k)+\log(6n^2)), 30 \times 8^2 \frac{n}{V_{k}^2 \bar{r}^2}(n\Phi(r_k)+\log(4n^2))\right),
\end{equation}
then
\begin{equation}\label{eq:reliability_concentration}
    \mathbb{P}\left(\|\hat{r}_{k}-r_{k}\|_{\infty} \geq \frac{\rho}{n}\right) \leq \exp(-n\Phi(r_k))    .
\end{equation} 
The sufficient condition of $d$ can also be written as \begin{equation*}
    d_k \geq
            C_1\frac{n^2}{V_k^4 \min(\rho^2,\bar{r}^2)}\left(n \Phi(r_k)  + \log (6 n^2)\right)
\end{equation*} with $C_1 = 15 \times 2^9$.

Using the inequality $|\log x-\log y|\leq \frac{\lvert x-y\rvert}{\min\{x,y\}}, \forall x,y>0$ implied by $\log x\leq x-1$ for positive $x$, we have that when $d_k$ satisfies \eqref{eq:number_of_tasks_known_type}, 
\begin{equation}
    \sum_i \max \left\{ 
        \left|\log \frac{1 + \hat{r}_{ki}}{1 + r_{ki}}\right|, \left|\log \frac{1 - \hat{r}_{ki}}{1 - r_{ki}}\right|
    \right\} \leq \frac{1}{2}
\end{equation}
with probability $\geq 1-\exp(-n\Phi(r_k))$.  
Now define the event
\begin{equation*}
    E_{k} \coloneqq \left\{\sum_i \max \left( 
            \left|\log \frac{1 + \hat{r}_{k i}}{1 + r_{k i}}\right|, \left|\log \frac{1 - \hat{r}_{k i}}{1 - r_{k i}}\right|
        \right) \leq \frac{1}{2}\right\} .
\end{equation*}
Under this event, the weights used by the NP estimate is approximately equal to the maximum likelihood weights. 
Applying \eqref{eq:reliability_concentration}, 
\begin{align*}
    \mathbb{P}\left(E_{k}^c\right) \leq \exp \left(-n \Phi(r_k) \right) .
\end{align*}

Without loss of generality, consider $y_j = 1$ so that $\hat{y}_j \neq y_j$ implies $\hat{y}_j = -1$.
\begin{align*}
    \mathbb{P}(\hat{y}_j \neq y_j) 
    & \leq \mathbb{P}(\{\hat{y}_j \neq y_j \}\cap E_{k_j}) + \mathbb{P}(E_{k_j}^c)\\
    & = \mathbb{P}\left( \left\{ 
        \sum_{i}\left(\log \frac{1+\hat{r}_{k_j i}}{1-\hat{r}_{k_j i}}X_{ij}\right) < 0 \right\}
    \cap E_{k_j}\right) + \mathbb{P}(E_{k_j}^c)\\
    & = \mathbb{P}\left(\left\{\prod_{i}\left(\frac{1-\hat{r}_{k_j i}}{1+\hat{r}_{k_j i}}\right)^{\mathbf{1}_{X_{ij} = 1}} \left(\frac{1+\hat{r}_{k_j i}}{1-\hat{r}_{k_j i}}\right)^{\mathbf{1}_{X_{ij} = -1}}\geq 1 \right\}\cap E_{k_j}\right ) +  \mathbb{P}(E_{k_j}^c)
    \numberthis \label{eq:label_estimation_decomposition1}
\end{align*}

% {\color{blue}Move this to a later part when it is used.}
% Using the inequality $|\log x-\log y|\leq \frac{\lvert x-y\rvert}{\min\{x,y\}}, \forall x,y>0$ implied by $\log x\leq x-1, \forall x>0$, we have 
% \begin{equation}
%     \frac{1}{2} \min \left \{1+\hat{r}_{ki},1-\hat{r}_{ki}, 1+r_{ki},1-r_{ki}\right\} \leq \rho ,
% \end{equation}
% and therefore
% \begin{align*}
%      \left\{\begin{array}{c}
%      \left|\frac{1+\hat{r}_{k i}}{2}-\frac{1+r_{k i}}{2}\right|<\frac{\hat{r}_{k}-r_{k}}{2}\\
%      \left|\frac{1-\hat{r}_{k i}}{2}-\frac{1-r_{k i}}{2}\right|<\frac{\hat{r}_{k}-r_{k}}{2} 
%      \end{array}\right., \ \ \ \forall i\in \mathcal{N}, \ \ k\in\{e,h\}.
% \end{align*} 

Define the two random variables
\begin{align*}    
    A_1 &= \prod_{i}\left(\frac{1-{r}_{k_j i}}{1+{r}_{k_j i}}\right)^{\mathbf{1}_{X_{ij} = 1}} \left(\frac{1+{r}_{k_j i}}{1-{r}_{k_j i}}\right)^{\mathbf{1}_{X_{ij} = -1}}
    \\ 
    A_2 &= \prod_{i}\left(\frac{(1-\hat{r}_{k_ji})(1+{r}_{k_ji})}{(1-{r}_{k_ji})(1+\hat{r}_{k_ji})}\right)^{\mathbf{1}_{X_{ij} = 1}}  \left(\frac{(1+\hat{r}_{k_ji})(1-{r}_{k_ji})}{(1+{r}_{k_ji})(1-\hat{r}_{k_ji})}\right)^{\mathbf{1}_{X_{ij} = -1}} .
\end{align*}
Then, the first event in the above probability is given by the product of $A_1$ and $A_2$.
On the event $E_{k_j}$, 
\begin{equation*}
A_2 \leq \exp\left(2\sum_i \max \left( 
            \left|\log \frac{1 + \hat{r}_{k_j i}}{1 + r_{k_j i}}\right|, \left|\log \frac{1 - \hat{r}_{k_j i}}{1 - r_{k_j i}}\right|
        \right)\right) \leq \exp(1) .
\end{equation*}
Therefore,
\begin{align}
    \mathbb{P}\left( \left\{ A_1 A_2 \geq 1 \right\} \cap E_{k_j} \right) & \leq \mathbb{P}\left(\left\{ A_1 \geq \exp \left(-1\right) \right\} \cap E_{k_j}\right)\\
    & \underbrace{\leq}{(a)}  
    \mathbb{P}\left(\left\{ A_1^{\frac{1}{2}} \geq \exp \left(-\frac{1}{2}\right) \right\} \cap E_{k_j}\right) \\
    & \leq \mathbb{P}\left(\left\{ A_1^{\frac{1}{2}} \geq \exp \left(-\frac{1}{2}\right) \right\} \right)\\
    & \underbrace{\leq}_{(b)} \exp \left(\frac{1}{2}\right) \mathbb{E}[A_1^{1/2}] ,
\end{align}
where in $(a)$ we used the observation that $A_1 > 0$ and in $(b)$ we used Markov's inequality on the random variable $A_1^{1/2} > 0$.
Evaluating the expectation,
\begin{align*}
    \mathbb{E}[A_1^{\frac{1}{2}}] & = \prod_{i}\left(\frac{1-{r}_{k_j i}}{1+{r}_{k_j i}}\right)^{\frac{1}{2}}\frac{1+r_{k_j,i}}{2} +\left(\frac{1+{r}_{k_j i}}{1-{r}_{k_j i}}\right)^{\frac{1}{2}}\frac{1-r_{k_j,i}}{2}\\
    & = \exp \left( \frac{1}{2} \sum_i \log \left(1 - r_{ki}^2\right)\right) = \exp(-n \Phi(r_k))  .
\end{align*}
Returning to \eqref{eq:label_estimation_decomposition1}, we have that for a task $j$ with type $k$,
\begin{align*}
    \mathbb{P}_k\left(\hat{y}_j \neq y_j \right) \leq 
    \exp \left(\frac{1}{2}\right) \mathbb{E}[A_1^{1/2}] + \mathbb{P}\left(E_{2, k}\right) 
    \leq \left(\exp \left(\frac{1}{2}\right) + 1\right) \exp \left(-n \Phi(r_k)\right) .
\end{align*}
Averaging, we get the error rate for known types as
\begin{equation}
    \mathbb{P}_{av}(y) = \frac{1}{d} \sum_{j=1}^d \mathbb{P}\left(\hat{y}_j \neq y_j \right) \leq 3 \sum_{k \in \{e, h\}}
        \frac{d_k}{d} \exp \left(-n \Phi(r_k)\right) .
\end{equation}

\section{Spectral Properties}\label{appendix:spectral_properties}
The content of this section goes as follows:
\begin{enumerate}[leftmargin=*]
    \item We establish the spectral properties of the signal matrix $n^{-1}R_y$ and give a proof to Lemma \ref{lemma: spectral properties of R}. 
    \item Next, we discuss the two largest eigenvalues and the corresponding eigenvectors of the matrix $\mathbb{E}[T].$ This discussion about the spectral properties of $\mathbb{E}[T]$ serves as a basis of the proof of Theorem \ref{thm:imperfect_cluster} given in \ref{proof:thm:imperfect_cluster}.
\end{enumerate}

Given the ordered response matrix $X$ we consider in Section \ref{sec:clustering}, where the easy and hard tasks are listed consecutively in the columns, the true response matrix with arbitrary task ordering is obtained by a column permutation of $X$.
It is easy to see that the ordered task similarity matrix $T = n^{-1} X^T X$ is then related to the true task similarity matrix with type-permutations by a similarity transform.
All eigenvalues and eigen-spectrums are therefore related by the same permutations, and as long as the algorithm does not utilize an unknown prior on the ordering of these types, its analysis still pertains to the un-ordered case.

Recall the decomposition of the expected task similarity matrix $\mathbb{E}[T]$ into 
\begin{align}\label{eq:ET}
   \mathbb{E}[T]&= \underbrace{n^{-1} \mathrm{diag}(y)\begin{pmatrix}
        \|r_e\|_2^2 1_{d_e \times d_e} & r_e^T r_h 1_{d_e \times d_h} \\ r_h^T r_e 1_{d_h \times d_e} & \|r_h\|_2^2 1_{d_h \times d_h} 
    \end{pmatrix}\mathrm{diag}(y)}_{n^{-1}R_y} 
   \underbrace{- n^{-1}\text{diag}\left([\|r_e\|_2^2 1_{1\times d_e}, \|r_h\|_2^2 1_{1\times d_h}]^T\right) + I_d}_{S}\nonumber
   \\&= n^{-1}R_y + S,   
   \end{align}
where $S$ is a diagonal matrix.
\subsection{Proof of Lemma \ref{lemma: spectral properties of R} : Spectral Properties of \texorpdfstring{$n^{-1}R_y$}{}}\label{Appendix: proof of lemma: spectral properties of R}
Recall, $n^{-1}R_y$ is defined as,
$n^{-1}R_y = n^{-1} \mathrm{diag}(y)\begin{pmatrix}
        \|r_e\|_2^2 1_{d_e \times d_e} & r_e^T r_h 1_{d_e \times d_h} \\ r_h^T r_e 1_{d_h \times d_e} & \|r_h\|_2^2 1_{d_h \times d_h} 
    \end{pmatrix}\mathrm{diag}(y).$
Clearly, the $n^{-1}R_y$ is a rank-$\ell$ matrix with $\ell \leq 2.$ Specifically,
\begin{equation*}
    \ell = \begin{cases}
        1, & \text{when $r_e$ and $r_h$ are colliner}\\
        2, & \text{else}
    \end{cases}
\end{equation*}
Next, we calculate the eigenspectram of $n^{-1}R_y$. First consider the case when $r_e^{\top}r_h \neq 0$.\\
\textbf{Case 1: When $r_e^{\top}r_h \neq 0:$}\\
Consider a generic vector $q$ of the form $\mathrm{diag}(y)[\overline{s}1_{1\times d_e}, 1_{1\times d_h}]^T$ for some $\overline{s}$ as the ratio of the magnitude between the entries of the vector corresponding to different types of tasks.. 
A normalization of $q$ serves as a candidate eigenvector for the matrix $n^{-1}R_y$, where
\begin{equation}
    \frac{q}{\|q\|_2}=\mathrm{diag}(y)\left[\begin{array}{c}
    \frac{\overline{s}}{\sqrt{d_e\overline{s}^2+d_h}}1_{d_e\times 1}\\
    \frac{1}{\sqrt{d_e\overline{s}^2+d_h}} 1_{d_h\times 1}
    \end{array}\right] .
\end{equation}
The eigen-pair equation for the candidate eigenvector above is calculated to be:
\begin{align}\label{eq:eigvecbarR_2}
    \frac{1}{n} R_y q 
&
    =\mathrm{diag}(y)\left[\begin{array}{c}
    \left[\frac{1}{n}(\overline{s} d_e \|r_e\|_2^2+d_h r_e^Tr_h)\right] 1_{d_e\times 1}\\
    \\
    \left[\frac{1}{n} (\overline{s} d_e r_e^Tr_h+d_h \|r_h\|_2^2) \right]1_{d_h\times 1}\end{array}\right]=\left[\frac{1}{n} (\overline{s} d_e r_e^Tr_h+d_h \|r_h\|_2^2) \right] q
\end{align}
Now as $\overline{s}$ is the ratio between the quantity $\frac{1}{n}(\overline{s} d_e \|r_e\|_2^2+d_h r_e^Tr_h)$ and $\frac{1}{n} (\overline{s} d_e r_e^Tr_h+d_h \|r_h\|_2^2)$, we can write: 
\begin{equation}\label{eq:condition_bar_s}
    \overline{s}  \left[\frac{1}{n} (\overline{s} d_e r_e^Tr_h+d_h \|r_h\|_2^2) \right] =\frac{1}{n}(\overline{s} d_e \|r_e\|_2^2+d_h r_e^Tr_h).
 \end{equation}   
The solutions to this quadratic equation are given by
\begin{equation}\label{eq:bar_s_12}
    s,\overline{s} = \frac{d_e\|r_e\|_2^2-d_h\|r_h\|_2^2\pm \sqrt{\left[d_e\|r_e\|_2^2-d_h\|r_h\|_2^2\right]^2+4d_ed_h(r_e^Tr_h)^2}}{2d_er_e^Tr_h}. 
\end{equation}
The eigenvalues $n^{-1} (\overline{s} d_e r_e^Tr_h+d_h \|r_h\|_2^2)$ of $n^{-1} R_y$ corresponding to each solution of $\overline{s}$ are
% \begin{align*}
%     \frac{d_e\|r_e\|_2^2+d_h\|r_h\|_2^2\pm \sqrt{\left[d_e\|r_e\|_2^2-d_h\|r_h\|_2^2\right]^2+4d_ed_h(r_e^Tr_h)^2}}{2n} .
% \end{align*} 
%  for $\bar{s}=\bar{\overline{\gamma}}$ and $\bar{s}=\bar{\overline{\gamma}}_2$. Therefore, 
 \begin{equation*}
 \lambda_1(n^{-1} R_y)=\frac{d_e\|r_e\|_2^2+d_h\|r_h\|_2^2+\sqrt{\left[d_e\|r_e\|_2^2-d_h\|r_h\|_2^2\right]^2+4d_ed_h(r_e^Tr_h)^2}}{2n}
  \end{equation*} 
 and
 \begin{equation}\label{eq:lambda_2_app_3_1}
 \lambda_2(n^{-1} R_y)=\frac{d_e\|r_e\|_2^2+d_h\|r_h\|_2^2-\sqrt{\left[d_e\|r_e\|_2^2-d_h\|r_h\|_2^2\right]^2+4d_ed_h(r_e^Tr_h)^2}}{2n} ,
  \end{equation} 
where $\lambda_1(n^{-1}R_y) \geq \lambda_2(n^{-1}R_y).$
  By Assumption \ref{assumption:easyhard}, we have $\norm{r_e} > 0$. Hence, for $d_e \geq 1$ and $d_h \geq 1,$ we can write, $\lambda_1 (n^{-1} R_y) > 0$ and $\lambda_2 (n^{-1} R_y) \geq 0.$ When $r_e$ and $r_h$ are co-linear, $\lambda_2(n^{-1}R_y) = 0.$\\
\textbf{Case 2: when $r_e^{\top}r_h = 0:$}\\
When the reliability vectors are orthogonal, we can write 
\begin{equation}\label{eq:direct_sum}
    n^{-1}R_y =    n^{-1} \|r_e\|_2^2 \mathrm{diag}(y_{1:d_e})1_{d_e\times d_e }\mathrm{diag}(y_{1:d_e}) 
    \oplus \|r_h\|_2^2 \mathrm{diag}(y_{d_e + 1: d})1_{d_h\times d_h}\mathrm{diag}(y_{d_e + 1: d}) ,
\end{equation}
where $y_{1:d_e}$ and $y_{d_e+1:d}$ are the ground truth vectors corresponding to type easy tasks and type hard tasks respectively and $\oplus$ is the notation for a direct sum. From the expression \eqref{eq:direct_sum}, it is clear that $\mathrm{rank}(n^{-1}R_y)=2$ when $\norm{r_h}_2 \neq0$ with the following eigenvalues:
 \begin{equation}
     \lambda_1(n^{-1}R_y)= n^{-1}\max_{k\in \{e,h\}}d_k\|r_k\|_2^2\geq \lambda_2(n^{-1}R_y)=n^{-1}\min_{k\in \{e,h\}}d_k \|r_k\|_2^2 ,
 \end{equation}
 with $\lambda_2 (n^{-1} R_y) \geq \lambda_j(n^{-1}R_y)=0$ for all $j=3,\ldots, d$. 
  
Also, the eigenvectors of $n^{-1} R_y$ corresponding to the eigenvalues $n^{-1}d_e \norm{r_e}_2$ and $n^{-1}d_h \norm{r_h}_2$ are respectively 
\begin{align*}
    \mathrm{diag}(y) \left[\begin{array}{c}
    \frac{1}{\sqrt{d_e}}1_{d_e\times 1}\\
    0_{d_h\times 1}\end{array}\right]\ \ \text{and}\ \ \mathrm{diag}(y) \left[\begin{array}{c}
    0_{d_e\times 1}\\
    \frac{1}{\sqrt{d_h}}1_{d_h\times 1}\\
\end{array}\right]
\end{align*}
\subsection{Spectral Properties of \texorpdfstring{$\mathbb{E}[T]$}{E[T]}}\label{appendix: spectral properties of ET}
The calculation of the spectral properties of $\mathbb{E}[R]$ is similar to the case of the matrix $n^{-1}R_y.$ We observe that the structure of the principal eigenvectors in both the matrices are similar with a difference in the ratio between the magnitude of the entries in the vector. 
Lets consider the case $r_e^{\top}r_h \neq 0$ first.\\
\textbf{Case 1: $r_e^{\top}r_h \neq 0$}\\
Consider a generic vector $q$ of the form $\mathrm{diag}(y)[\overline{\gamma}1_{1\times d_e}, 1_{1\times d_h}]^T$ for some $\overline{\gamma}$.
A normalization of $q$ serves as a candidate eigenvector for the matrix $\mathbb{E}[T]$, where
\begin{equation}\label{eq:genericEigenvector_ET}
    \frac{q}{\|q\|_2}=\mathrm{diag}(y)\left[\begin{array}{c}
    \frac{\overline{\gamma}}{\sqrt{d_eS^2+d_h}}1_{d_e\times 1}\\
    \frac{1}{\sqrt{d_e\overline{\gamma}^2+d_h}} 1_{d_h\times 1}
    \end{array}\right] .
\end{equation}
The eigen-pair equation for the candidate eigenvector above is calculated to be
\begin{align}\label{eq:eigvecbarR}
    \mathbb{E}[T] q
    &=\mathrm{diag}(y)\left[\begin{array}{c}
    \left[\frac{1}{n}(\overline{\gamma} d_e \|r_e\|_2^2+d_h r_e^Tr_h)-\frac{1}{n}\overline{\gamma}\|r_e\|_2^2+\overline{\gamma}\right] 1_{d_e\times 1}\\
    \\
    \left[\frac{1}{n} (\overline{\gamma} d_e r_e^Tr_h+d_h \|r_h\|_2^2)-\frac{1}{n}\|r_h\|_2^2+1 \right]1_{d_h\times 1}\end{array}\right] 
    \\ &=\left[\frac{1}{n} (\overline{\gamma} d_e r_e^Tr_h+d_h \|r_h\|_2^2)-\frac{1}{n}\|r_h\|_2^2+1 \right] q  , \numberthis \label{eq:eig2}
\end{align}
which gives rise to the quadratic equation
\begin{equation}\label{eq:condition_s}
    \overline{\gamma} \left[\frac{1}{n} (\overline{\gamma} d_e r_e^Tr_h+d_h \|r_h\|_2^2)-\frac{1}{n}\|r_h\|_2^2+1 \right] =\frac{1}{n}(\overline{\gamma} d_e \|r_e\|_2^2+d_h r_e^Tr_h)-\frac{1}{n}\overline{\gamma}\|r_e\|_2^2+\overline{\gamma} .
 \end{equation}   
The condition on $\overline{\gamma}$ given by (\eqref{eq:condition_s}) corresponds to a quadratic equation whose solutions are given by
\begin{equation}\label{eq:s_12}
    \gamma,\overline{\gamma}= \frac{(d_e-1)\|r_e\|_2^2-(d_h-1)\|r_h\|_2^2\pm \sqrt{\left[(d_e-1)\|r_e\|_2^2-(d_h-1)\|r_h\|_2^2\right]^2+4d_ed_h(r_e^Tr_h)^2}}{2d_er_e^Tr_h}.
\end{equation}
As a result, we see that eigenvalues corresponding to eigenvectors of the form \eqref{eq:genericEigenvector_ET} are given by
\begin{align*}
    \frac{(d_e-1)\|r_e\|_2^2+(d_h-1)\|r_h\|_2^2+\sqrt{\left[(d_e-1)\|r_e\|_2^2-(d_h-1)\|r_h\|_2^2\right]^2+4d_ed_h(r_e^Tr_h)^2}}{2n}+1 \numberthis \label{eq:eigenvalues_non_orthogonal_1} \\
% \begin{equation}\label{eq:lambda_2_app_3}
    \frac{(d_e-1)\|r_e\|_2^2+(d_h-1)\|r_h\|_2^2-\sqrt{\left[(d_e-1)\|r_e\|_2^2-(d_h-1)\|r_h\|_2^2\right]^2+4d_ed_h(r_e^Tr_h)^2}}{2n}+1 . \numberthis \label{eq:eigenvalues_non_orthogonal_2}
% \end{equation} 
\end{align*}
Next we show that the expression in equation \eqref{eq:eigenvalues_non_orthogonal_1} corresponds to the largest eigenvalue of $\mathbb{E}[T]$ when $\min(d_e,d_h) \geq 2.$ Recall the decompsition $\mathbb{E}[T] = n^{-1}R_y + S$. We also know that $S$ is a diagonal matrix with its eigenvalues belonging to the set $ \{1-n^{-1}\norm{r_h}_2 , 1- n^{-1}\norm{r_e}_2\}$. Clearly with an application of Weyl's inequality we can bound the $g$-th largest eigenvalue of $\mathbb{E}[T]$ as 
\begin{equation*}
    \lambda_g(\mathbb{E}[T]) \leq \lambda_g(n^{-1}R_y) + \{1-n^{-1}\norm{r_h}_2  \leq \lambda_g(n^{-1}R_y) + 1
\end{equation*}
 We observe when $\min(d_e,d_h) \geq 2,$
 the following is satisfied : 
 \begin{align*}
      & \frac{(d_e-1)\|r_e\|_2^2+(d_h-1)\|r_h\|_2^2+\sqrt{\left[(d_e-1)\|r_e\|_2^2-(d_h-1)\|r_h\|_2^2\right]^2+4d_ed_h(r_e^Tr_h)^2}}{2n}+1 \\
      & \geq \lambda_g(n^{-1}R_y) + 1 , \forall g \geq 2
 \end{align*}
 implying, 
 \begin{align*}
     & \lambda_1(\mathbb{E}[T]) = \\
     &  \frac{(d_e-1)\|r_e\|_2^2+(d_h-1)\|r_h\|_2^2+\sqrt{\left[(d_e-1)\|r_e\|_2^2-(d_h-1)\|r_h\|_2^2\right]^2+4d_ed_h(r_e^Tr_h)^2}}{2n}+1 
 \end{align*}
\begin{comment}

For these eigenvalues to be the largest two, we must compute an upper bound on the remaining eigenvalues of $\mathbb{E}[T]$.
From the decomposition \eqref{eq:ET_decomposition_main_part}, we use Weyl's inequality to relate the remaining eigenvalues of $\mathbb{E}[T]$ to those of $n^{-1}R_y$: Given $\lVert r_h \rVert \leq \lVert r_e \rVert \neq 0$,
\begin{align*}
    \lambda_j \left(\mathbb{E}[T]\right) \leq \lambda_j (I_d + n^{-1} R_y) - \lambda_d \left(n^{-1} \mathrm{diag}([\lVert r_e \rVert_2^2 1_{1 \times d_e}, \lVert r_h \rVert_2^2 1_{1 \times d_h}])\right) = 1 - n^{-1} \lVert r_h \rVert_2^2 < 1 .
\end{align*}
When both possible eigenvalues \eqref{eq:eigenvalues_non_orthogonal} corresponding to the two possible values of $\overline{\gamma}$ exceed $1 - n^{-1} \lVert r_h \rVert_2^2$, the two are guaranteed to be the largest eigenvalues of $\mathbb{E}[T]$.
The following Assumption \ref{assumption:spectral_properties} provides such a sufficient condition.
\begin{assumption}\label{assumption:spectral_properties}
    The reliability vectors satisfy: For any $d_e \geq 1$: 
    \begin{equation}
        r_e^T r_h \leq
        \norm{r_h}^2 \left(
            \left(1 - \frac{1}{d_e}\right) \norm{r_e}^2 + \frac{1}{d_e} \norm{r_h}^2
        \right).
    \end{equation}
\end{assumption}
Therefore, we have that $\lambda_1 (\mathbb{E}[T])$ and $\lambda_2 (\mathbb{E}[T])$, where eigenvalues are indexed in non-increasing order, are given by the two expressions in \eqref{eq:eigenvalues_non_orthogonal} under Assumption \ref{assumption:spectral_properties}.
\end{comment}

\textbf{Case 2: when $r_e^{\top}r_h = 0:$}\\
In this case, $\mathbb{E}[T]$ is block-diagonal and can be expressed as the direct sum
\begin{align*}
    \mathbb{E}[T] & = \frac{1}{n} \mathrm{diag}(y_e)\|r_e\|_2^2 1_{d_e\times d_e} \mathrm{diag}(y_{e})- \frac{1}{n}\|r_e\|_2^2 I_{d_e} + I_{d_e}\nonumber\\& \oplus \frac{1}{n} \mathrm{diag}(y_{h})\|r_h\|_2^2 1_{d_h\times d_h}\mathrm{diag}(y_{h})- \frac{1}{n}\|r_h\|_2^2 I_{d_h} + I_{d_h} .
\end{align*}
The eigenvalues of $\mathbb{E}[T]$ are the eigenvalues of its blocks. 
Each block indexed by $k$ is associated with eigenvalues $1 + n^{-1} (d_k - 1) \lVert r_k \rVert_2^2 \geq 1$ and $1 - n^{-1} \lVert r_k \rVert_2^2$ with multiplicity $d_k - 1$.
Hence, the largest two eigenvalues have the following relation:
\begin{align*}
    \lambda_1 (\mathbb{E}[T]) = 1 + \max_{k \in \{e, h\}} \frac{d_k - 1}{n} \lVert r_k \rVert_2^2 \geq 1 + \min_{k \in \{e, h\}} \frac{d_k - 1}{n} \lVert r_k \rVert_2^2 = \lambda_2 (\mathbb{E}[T]) ,
\end{align*}
and $\lambda_2 (\mathbb{E}[T]) \geq 1 - n^{-1} \lVert r_h \rVert_2^2$ is the second largest eigenvalue.

By a direct calculation using the same steps as in the non-orthogonal case, the two corresponding eigenvectors are given by
\begin{align*}
    \mathrm{diag}(y) \left[\begin{array}{c}
    \frac{1}{\sqrt{d_e}}1_{d_e\times 1}\\
    0_{d_h\times 1}\end{array}\right], \quad \mathrm{diag}(y) \left[\begin{array}{c}
    0_{d_e\times 1}\\
    \frac{1}{\sqrt{d_h}}1_{d_h\times 1}\\
    \end{array}\right]
\end{align*}
corresponding to the eigenvalue associated with $1+n^{-1}(d_e - 1) \lVert r_e \rVert_2^2$ and $1+n^{-1}(d_h - 1) \lVert r_h \rVert_2^2$, respectively.
The normalized eigen-gap of $\mathbb{E}[T]$ is then given by
\begin{equation}
    \sigma(\mathbb{E}[T]) = 
       d^{-1} n^{-1} \lvert (d_e - 1) \lVert r_e\rVert_2^2 - (d_h - 1) \lVert r_h \rVert_2^2 \rvert
   .
\end{equation}

\begin{comment}

\begin{remark}
When reliability vectors are orthogonal, task types can be clustered using the principal eigenvector, and the ground-truth labels can be recovered up to sign ambiguity using the signs of each eigenvector corresponding to $\lambda_1 (\mathbb{E}[T])$ and $\lambda_2 (\mathbb{E}[T])$.
% In the case of orthogonal reliability vectors, task separation can be achieved given the dominant eigenvector of $\mathbb{E}[T]$ as it is clear from the form of this eigenvector. Therefore, task separation can be performed based on $T$ by applying Algorithm \ref{alg:cluster}. Also, the ground-truth labels  can be recovered, up to the sign ambiguity of any eigenvector, by considering the signs of the entries of \emph{both} the two dominant eigenvectors of $\mathbb{E}[T]$. Therefore, they can be estimated in practice by the two dominant eigenvectors of $T$. Here, we specifically refer to $\mathbb{E}[T]$ and $T$, since the permutation of $X$ resulting in $X$ is unknown in practice.
\end{remark}

\begin{remark}
The coincidence of the eigenvectors corresponding to the largest and second largest eigenvalues (as well as an approximate coincidence of principal eigenvalues and associated eigengaps) for $\mathbb{E}[T]$ and $n^{-1} R_y$ holds regardless of whether $r_e$ and $r_h$ are orthogonal or not.
% We note the approximate coincidence of principal eigenvectors, principal eigenvalues and associated eigengaps of $\mathbb{E}[T]$ and $n^{-1}\bar{R}$ for nonorthogonal reliability vectors. The same also holds for orthogonal reliability vectors.
\end{remark}
\end{comment}
\section{Remaining Part of Proofs for Theorem \ref{thm:cluster_crowdsourcing}}
\subsection{Relating the Event of Misclustering to Eigenvector Concentration}
Before stating the sufficient condition for the perfect clustering, we state a more general result that provides the sufficient conditions for a clustering error $\eta \leq 1-t$ for some $t \in [0,1]$ in the following proposition:
\begin{proposition}\label{prop:misclustering implication}
    Let $\theta$ be the sign that resolves the eigenvector ambiguity
    \begin{align*}
        \theta = \arg \min_{\theta \in \{-1, +1\}} \lVert v(n^{-1}R_y) - \theta \hat{v}\rVert .
    \end{align*}
   Fix any non-negative $t \leq 1$. Algorithm \ref{alg:cluster} returns cluster membership with $\eta \leq 1- t$ on the following event on the random vector $\hat{v}$ and the random variable $\hat{\mu}$:
   \begin{equation}\label{eq:event eta leq 1 - t}
       \frac{1}{d}\sum_{j=1}^d\indicator(E_{\hat{v},j}) \geq t 
   \end{equation}
   where, 
    \begin{equation}
        E_{\hat{v}, j} = \left\{
            \lvert v_j(n^{-1}R_y) - \theta \hat{v}_j \rvert 
            +\lvert \mu(n^{-1}R_y) - \hat{\mu}\rvert< \min \{m_e(n^{-1}R_y), m_h(n^{-1}R_y)\} 
        \right\} 
    \end{equation}
\end{proposition}
\begin{proof}
    Assume the event defined in equation \eqref{eq:event eta leq 1 - t} is true for a fixed $t$ such that $0 \leq t \leq 1$. Under this event we show that there exists a permutation $\pi$ from $\{e,h\}$ to $\{e,h\}$ such that $\eta \leq 1-t$.\\
    First, consider the case of $\mu_e(n^{-1}R_y) \geq \mu_h(n^{-1}R_y)$. Our candidate permutation for this case is $\pi = \{e \mapsto e ; h \mapsto h\}$.
    We claim that when event $E_{\hat{v},j}$ is true, the task $j$ is clustered into group $1$ if $k_j=e$ and into group $2$ otherwise. Under this claim, it is easy to see that on the event \eqref{eq:event eta leq 1 - t}, at least $t$ fraction of tasks are correctly clustered, that is,  $\eta \leq 1-t$. We are left to prove the claim now. Consider the case $k_j = e$ for a task $j$.
    By definition of the absolute margins $m_e(n^{-1}R_y)$ and $m_h(n^{-1}R_y)$, we have that $\min \{m_e(n^{-1}R_y), m_h(n^{-1}R_y)\} \leq \lvert v_j(n^{-1}R_y) \rvert - \mu$.
    Suppose $E_{\hat{v}, j}$ is true .
    Then,
    \begin{align*}
        \lvert \hat{v}_j \rvert - \hat{\mu} & = \lvert v_j(n^{-1}R_y) \rvert - \mu(n^{-1}R_y) + \mu(n^{-1}R_y) - \hat{\mu} + \lvert \theta \hat{v}_j \rvert - \lvert v_j(n^{-1}R_y) \rvert \\
        &
        \geq \min \{m_e(n^{-1}R_y), m_h(n^{-1}R_y) \} - \lvert v_j(n^{-1}R_y) - \theta \hat{v}_j \rvert 
            +\lvert \mu(n^{-1}R_y) - \hat{\mu}\rvert\\
            & \underbrace{>}_{(a)} \min \{m_e(n^{-1}R_y), m_h(n^{-1}R_y) \} - \min \{m_e(n^{-1}R_y), m_h(n^{-1}R_y) \} = 0 ,
    \end{align*}
    where $(a)$ is due to event $E_{\hat{v},j}$.
    This implies $\lvert \hat{v}_j \rvert > \hat{\mu}$.
    This proves that task $j$ is correctly clustered as $\hat{k}_j=e$ and $\pi(\hat{k}_j) = e$.
    By the similar arguments for $k_j = h$, we obtain that $\pi (\hat{k}_j) = h$ in the same event.\\
    Lastly, consider the case of $\mu_e(n^{-1}R_y) < \mu_h(n^{-1}R_y)$. The flow is almost identical for the case of $\mu_e(n^{-1}R_y) \geq \mu_h(n^{-1}R_y)$ but, it is given below for completeness. Our candidate permutation for this case is $\pi = \{e \mapsto h ; h \mapsto e\}$.
    We claim that when event $E_{\hat{v},j}$ is true, the task $j$ is clustered into group $1$ if $k_j=h$ and into group $2$ otherwise. Under this claim, it is easy to see that under event \eqref{eq:event eta leq 1 - t}, at least $t$ fraction of tasks are correctly clustered, that is,  $\eta \leq 1-t$. We are left to prove the claim now. Consider the case $k_j=e$ for a task $j$.
    By definition of the absolute margins $m_e(n^{-1}R_y)$ and $m_h(n^{-1}R_y)$, we have that $\min \{m_e(n^{-1}R_y), m_h(n^{-1}R_y)\} \leq \mu(n^{-1}R_y) - \lvert v_j(n^{-1}R_y) \rvert $.
    Suppose $E_{v, j}$ is true.
    Then,
    \begin{align*}
        \hat{\mu}-\lvert \hat{v}_j \rvert  & = \mu-\lvert v_j(n^{-1}R_y) \rvert   + \hat{\mu}-\mu(n^{-1}R_y) + \lvert \theta \hat{v}_j \rvert - \lvert v_j(n^{-1}R_y) \rvert \\
        & \geq \min \{m_e(n^{-1}R_y), m_h(n^{-1}R_y) \} -  \lvert v_j(n^{-1}R_y) - \theta \hat{v}_j \rvert 
            +\lvert \mu(n^{-1}R_y) - \hat{\mu}\rvert\\
        & \underbrace{>}_{(a)} \min \{m_e(n^{-1}R_y), m_h(n^{-1}R_y) \}  -  \min \{m_e(n^{-1}R_y), m_h(n^{-1}R_y) \}  = 0 ,
    \end{align*} where $(a)$ is due to the event $E_{\hat{v},j}$.
    This implies $\lvert \hat{v}_j \rvert < \hat{\mu}$.
    This proves that task $j$ is correctly clustered as $\pi(\hat{k}_j) = e$.
    Repeating the same argument for $k_j = h$, we obtain that $\pi (\hat{k}_j) = h$ in the same event.
\end{proof}

\subsubsection{Concentration of the Threshold $\hat{\mu}$}
Recall, the Algorithm \ref{alg:cluster} uses the following threshold to cluster the entris of $|\hat{v}|$ : 
\begin{equation*}
    \hat{\mu} = \frac{1}{d}\sum_{j=1}^d |\hat{v}_j|
\end{equation*}

\textbf{Fact:}
For any vectors $v, \hat{v}$ of dimension $d$ the mean absolute error $\lvert \mu(n^{-1}R_y) - \hat{\mu} \rvert$ between the average of magnitudes $\mu = d^{-1} \sum_{j=1}^d \lvert v_j \rvert$ and that of $\hat{v}$ satisfies
\begin{equation}\label{eq:mu_vs_v}
    \lvert \mu\ - \hat{\mu} \rvert \leq d^{-1/2} \min_{\theta \in \{-1, +1\}} \lVert v - \theta \hat{v} \rVert_2 \leq  \min_{\theta \in \{-1, +1\}} \norm{v-\theta \hat{v}}_{\infty}. 
\end{equation}
\begin{proof}
    \begin{align*}
        \hat{\mu} - \mu = \frac{1}{d} \sum_{j=1}^d \left(\lvert \theta \hat{v}_j \rvert - \lvert v_j \rvert \right) = \frac{1}{d} \sum_{j=1}^d \left(\lvert \hat{v}_j \rvert - \lvert v_j \rvert \right) .
    \end{align*}
    Taking the absolute value and using the triangle inequality, followed by the root mean square - arithmetic mean inequality,
    \begin{align*}
        \lvert \hat{\mu} - \mu \rvert \leq \frac{1}{d} \sum_{j=1}^d \lvert \hat{v}_j - v_j \rvert \leq d^{-1/2} \lVert \hat{v} - v\rVert \leq \min_{\theta \in \{-1, +1\}} \norm{v-\theta \hat{v}}_{\infty}.
    \end{align*}
\end{proof}
Using the above fact, we can relate the concentration of $\hat{\mu}$ with respect to \\$\mu(n^{-1}R_y) = \frac{1}{d}\sum_{j=1}^d|v_j(n^{-1}R_y)|$ as
\begin{equation}\label{eq: mu concentration means}
    \lvert \hat{\mu} - \mu(n^{-1}R_y) \rvert \leq \frac{1}{d} \sum_{j=1}^d \lvert \hat{v}_j - v_j(n^{1}R_y) \rvert \leq d^{-1/2} \lVert \hat{v} - v(n^{-1}R_y)\rVert \leq \min_{\theta \in \{-1, +1\}} \norm{v(n^{-1}R_y)-\theta \hat{v}}_{\infty}.
\end{equation}
\subsection{Proof of Proposition \ref{prop:conditions for perfect clustering}: Relating the Event of Perfect Clustering with Eigenvector Concentration}\label{appendix:perfect clustering condition proof}

    The Proposition \ref{prop:conditions for perfect clustering} is an immediate implication of \ref{prop:misclustering implication} and the equation \eqref{eq: mu concentration means}. 
    On the event $E_{l_\infty}$, using equation \eqref{eq: mu concentration means}, the following is satisfied : $|\hat{\mu} - \mu| < \frac{1}{2 } \min \left\{m_e(n^{-1}R_y) , m_h(n^{-1}R_y) \right\}.$ Hence, the event $E_{\hat{v},j}$ is satisfied for all $j \in [n]$. Hence $\eta =0 $ is achieved from Proposition \ref{prop:misclustering implication}.
\section{Label Estimation Performance:Proof of Theorem \ref{thm:labeling easy}}\label{appendix:proof of labeling}
The expected rate of labeling error using the law of total expectation can be decomposed as :
\begin{align}
    \mathbb{E}\left(\frac{1}{d} \sum_j {\mathbf 1}\left(\hat{y}_j \neq y_j\right)\right) = 
    \underbrace{\mathbb{E}\left(\frac{1}{d} \sum_j {\mathbf 1}\left(\hat{y}_j \neq y_j\right) | E_{pc}\right)\mathbb{P}(E_{pc})}_{I} +
    \underbrace{\mathbb{E}\left(\frac{1}{d} \sum_j {\mathbf 1}\left(\hat{y}_j \neq y_j\right)|E_{pc}^c\right)\mathbb{P}(E_{pc}^c)}_{II} 
\end{align}
where $E_{pc}$ is defined as the event of perfect clustering, that is $\eta = 0$.
We upper bound second term $II$ as $\mathbb{P}(E_{pc}^c)$. When the condition $\eqref{eq:cluster_d_requirement_main}$ is satisfied by $\left(n_{cl}, d, r_k(\mathrm{N}_{cl})\right)$ for each $k \in \{e,h\}$ it is characterized by Theorem \ref{thm:cluster_crowdsourcing} as :
\begin{equation*}
    II \leq
        2 d^2 \exp \left(- C_{4} n_{cl} D(r_e(\mathcal{N}_{cl}),r_h(\mathcal{N}_{cl}),\alpha,d)\right)
\end{equation*}
To upper bound the term $I$, we invoke the Proposition \ref{prop:achievable_perfect}. Recall the partition of the set of workers to mutually exhaustive sets $\mathcal{N}_{cl}$ and $\mathcal{N}_{rl}$ for clustering and the labeling steps respectively. Hence, given the event $E_{pc}$, the labeling step has perfect knowledge of each task's type and $NP-WMV$ for known type model would yield the following error rate when   $\left(n_{rl}, d_e,d_h,r_k(\mathrm{N}_{rl})\right)$ satisfy the conditions stated in Proposition \ref{prop:achievable_perfect}: 
\begin{equation*}
    I \leq 3\sum_{k\in \{1,2\}}\frac{d_k}{d}\exp \left(-n_{rl} \Phi_{k,\mathcal{N}_{rl}}\right).
\end{equation*}

\section{Imperfect Cluster: Proof of Theorem \ref{thm:imperfect_cluster}}\label{appendix: proof of imperfect cluster}
Below we give a proof sketch for theorem \ref{thm:imperfect_cluster}. The idea is to view the task-similarity $T$ matrix as a perturbation of its expectation $\mathbb{E}[T].$ Recall, in the proof of Theorem \ref{thm:cluster_crowdsourcing} given in Section \ref{appendix: perfect clustering}, we leverage the structure of the vector $v(n^{-1}R_y)$ as it has the type information for all tasks. In contrast, we leverage the structure of the principal eigenvector of $\mathbb{E}[T]$ for this proof. While the key steps are similar to the proof of clustering theorem \ref{thm:cluster_crowdsourcing}, there are a few technical differences. The proof goes as follows:
\begin{enumerate}[leftmargin=*]
    \item In Section \ref{appendix: spectral properties of ET}, we have shown the structure of the principal eigenvector $v(\mathbb{E}[T])$ of the matrix $\mathbb{E}[T]$. We observe that the tasks can be clustered into easy and hard types by clustering the magnitude of the entries of $v(\mathbb{E}[T])$. In the first part of the proof, we show that $\hat{v}$, the principal eigenvector of $T$ is a perturbation of $v(\mathbb{E}[T])$ in the $l_2$-norm sense. This is established using a Davis-Kahan perturbation result on the following matrix perturbation $T = \mathbb{E}[T] + N.$ The detail of this step is given in Section \ref{appendix: l_2 concentraion}.
    \item Next we relate the event $\eta \leq t$, for some $t \in [0,1)$ with the concentration of $\hat{v}$ around $v(\mathbb{E}[T])$ in the $l_2$-norm sense and prove that the event occurs with high probability. Recall, in contrast, the proof of Theorem \ref{thm:cluster_crowdsourcing} characterizes the event $\eta \neq 0$. This step is carried out in detail in Section \ref{proof:thm:imperfect_cluster}.
\end{enumerate}
\subsection{\texorpdfstring{$l_2$}{} Concentration of the Principal Eigenvector}\label{appendix: l_2 concentraion}
\begin{lemma}[$l_2$ norm concentration]\label{lemma:l_2 concentration}
   For every $t \geq 0$,
\begin{equation}\label{eq:eigenvec_distance_2}
    \mathbb{P}\left(\min_{\theta\in \{\pm 1\}} \|v(\mathbb{E}[T])-\theta \hat{v}\|_2\geq t\right) \leq 
    % 2d\exp\left(
    % - C_5 n \frac{\lambda_1 (\mathbb{E}[T]) - \lambda_2 (\mathbb{E}[T])}{d}t^2
    % \right) ,
         2 d \exp \left(
        -  C_7 n \frac{ \left(\lambda_1 (\mathbb{E}[T]) - \lambda_2 (\mathbb{E}[T])\right)^2}{d^2} t^2
     \right)
\end{equation}
where $C_7 = 2^{-6}$ is a universal constant.
% where $\tilde{C}_4 =  \frac{\Delta}{2^6}$ with $\Delta = \frac{\nu(\mathbb{E}[T])}{2^6}$.
% Note for the important case of $d_e\asymp d, d_h\asymp d$ and $|r_e^Tr_h|\asymp n$, we have that $\tilde{C}_4 = O(1)$  
\end{lemma}

\begin{proof}
We use the Davis-Kahan Theorem to relate the concentration of the principal eigenvector of $T$ with that of the matrix $\mathbb{E}[T]$.
Then we apply the Matrix-Hoeffding inequality to obtain a tail bound on the difference $T - \mathbb{E}[T]$ in the $l_2$-norm sense.

The Davis-Kahan Theorem \citet{DavisKahan} states that 
\begin{equation}\label{ineq:davis_kahan}
    \min_{\theta\in \{\pm 1\}} \|v(\mathbb{E}[T]) - \theta \hat{v}\|_2 \leq 2^{\frac{3}{2}} \frac{\|T-\mathbb{E}T\|_2}{\lambda_1(\mathbb{E}[T])-\lambda_2(\mathbb{E}[T])} 
\end{equation}

    % We apply the Matrix-Hoeffing inequality together with the Davis-Kahan Theorem to 
   % We start with a matrix concentration inequality on the normalized task similarity matrix.
 Next, we obtain the tail bound on the operator norm $\lVert T - \mathbb{E}[T] \rVert_2$ using Matrix-Hoeffding inequality.
A version (cf. \citet{MatrixHoeffding}, Corollary 4.2) that works for our application is stated below:
\begin{lemma}[Matrix Hoeffding]
    Consider a sequence of independent Hermitian matrices $\{Z_i\}_{i=1}^n$, each of dimension $d$.
    % Let $\{Z_i\}_{i=1}^n$ be a sequence of independent Hermitian matrices of dimension $d$.
    % Suppose for some sequence $\{A_i\}_{i=1}^n$ of Hermitian matrices that 
    Suppose for some sequence of Hermitian matrices $\{A_i\}_{i=1}^n$, each $Z_i$ satisfies
    \begin{align*}
        \mathbb{E} Z_i = 0, - A_i^2 \preceq Z_i^2 \preceq A_i^2 ,
    \end{align*}
    where $\preceq$ is the L\"{o}wner order.
    % for all $i$ with respect to the L\"{o}wner order $\preceq$.
    Then,
    \begin{align*}
        \mathbb{P}\left( \left\lVert \sum_i Z_i \right\rVert_2 \geq t\right) \leq 2 d \exp\left(-\frac{t^2}{2\sigma^2}\right)
    \end{align*}
    where $\sigma^2 = \frac{1}{2}\norm{\sum_i (A_i^2 + \mathbb{E} Z_i^2)}$.
\end{lemma}
% Moreover, we can apply this to obtain the right tail for $\lambda_{\max}(-\sum_i Z_i)$ when $-Z_i^2 \preceq A_i^2$ as well.
% When both conditions are satisfied so that $-A_i^2 \preceq Z_i^2 \preceq A_i^2$, we obtain the inequality multiplied by a factor $2$.
To apply the above result, consider the $d \times d$ matrix $n^{-1} \left(X_{i \cdot}^T X_{i \cdot} - \mathbb{E}[X_{i\cdot}^T X_{i\cdot}]\right)$ formed by the response of worker $i$ to be substituted as $Z_i$.
The matrix $Z_i$ contains entries whose magnitude is bounded by $2 n^{-1}$, and its square is therefore bounded element-wise. 
As a result, we have 
\begin{equation}
    - \frac{4 d^2}{n^2} I_d \preceq Z_i^2 \preceq \frac{4 d^2}{n^2} I_d, \quad \sigma^2 \leq \frac{4d^2}{n} .
\end{equation}
where we choose the sequence of Hermitian matrices $\{A_i\}_{i=1}^n$ for our purpose as $A_i = \frac{4 d^2}{n^2} I_d , \forall i \in [n]$.
Using the construction $\sum_{i=1}^n Z_i = T - \mathbb{E}[T]$, we directly obtain 
\begin{equation}
    \mathbb{P}\left(\left\lVert T - \mathbb{E}[T]\right\rVert_2 \geq t \right) \leq 2 d \exp \left(
        -\frac{n t^2}{8 d^2}
    \right) .
\end{equation}
Combining with Davis-Kahan's Theorem, we have that
\begin{align*}
    \mathbb{P}\left( \min_{\theta\in \{\pm 1\}} \|v - \theta \hat{v}\|_2 \geq t \right)
    & \leq 
    \mathbb{P}\left(
        \|T-\mathbb{E}T\|_2 \geq 2^{-3/2} \left(\lambda_1 (\mathbb{E}[T]) - \lambda_2 (\mathbb{E}[T])\right) t
         % 2^{\frac{3}{2}} \frac{}{\lambda_1(\mathbb{E}[T])-\lambda_2(\mathbb{E}[T])} \geq t 
     \right)
     \\ 
     &
     \leq
     2 d \exp \left(
        -  \frac{n \left(\lambda_1 (\mathbb{E}[T]) - \lambda_2 (\mathbb{E}[T])\right)^2}{2^6d^2} t^2
     \right)
\end{align*}
\end{proof}

\subsection{Proof of Theorem \ref{thm:imperfect_cluster}: Imperfect Clustering}\label{proof:thm:imperfect_cluster}

First, we need to define a few quantities based on the vector $v(\mathbb{E}[T])$. Similar to the definition of $\mu(n^{-1}R_y)$, $\mu_e(n^{-1}R_y)$, $\mu_h(n^{-1}R_y)$ $m_e(n^{-1}R_y)$ and $m_h(n^{-1}R_y)$, let us define the quantities based on the vector $v(\mathbb{E}[T])$. That is,

\begin{equation}
    \mu_e(\mathbb{E}[T]) = \begin{cases}
        \left|\frac{\gamma}{\sqrt{\gamma^2d_e+d_h}}\right| , & \text{when $r_e^{\top}r_h \neq 0$} \\ 
        \frac{1}{\sqrt{d_e}}, & \text{when $r_e^{\top}r_h = 0$}
    \end{cases}
\end{equation}
\begin{equation}
    \mu_h(\mathbb{E}[T]) = \begin{cases}
       \left| \frac{1}{\sqrt{\gamma^2d_e+d_h}}\right|, & \text{when $r_e^{\top}r_h \neq 0$} \\ 
        0, & \text{when $r_e^{\top}r_h = 0$}
    \end{cases}
\end{equation}
\begin{equation}
    \mu(n^{-1}R_y)=\frac{d_e}{d}\mu_e(n^{-1}R_y) + \frac{d_h}{d} \mu_h(n^{-1}R_y).
\end{equation}
\begin{align}
    m_e(n^{-1}R_y)
    % =|v_e - \mu| 
    = \mu_e(\mathbb{E}[T]) - \mu(\mathbb{E}[T])
    = \frac{d_h}{d}(\mu_e(\mathbb{E}[T]) - \mu_h(\mathbb{E}[T])) = \frac{d_h}{d}\frac{||\gamma| - 1|}{\sqrt{d_e \gamma^2 + d_h}} \label{eq:m_e(ET)}
\\
    m_h(n^{-1}R_y)
    % =|\mu - v_h| 
    = \mu(\mathbb{E}[T]) - \mu_h(\mathbb{E}[T]) = 
    \frac{d_e}{d}(\mu_e(\mathbb{E}[T]) - \mu_h(\mathbb{E}[T]))  = \frac{d_e}{d}\frac{||\gamma| - 1|}{\sqrt{d_e \gamma^2 + d_h}} \label{eq:m_h(ET)} .
\end{align}
Recall the Proposition \ref{prop:misclustering implication} where, we give sufficient condition for the event $\eta \leq 1-t$ for some $t \in [0,1]$ which was related to the concentration of $\hat{v}$ to $v(n^{-1}R_y)$. It turns out the similar relation holds when we replace $v(n^{-1}R_y)$, $\mu(n^{-1}R_y)$ , $m_e(n^{-1}R_y)$ and $m_h(n^{-1}R_y)$ with $v(\mathbb{E}[T])$, $\mu(\mathbb{E}[T])$, $m_e(\mathbb{E}[T])$ and $m_h(\mathbb{E}[T])$, respectively. It is shown in the following proposition.

\begin{proposition}\label{prop:misclustering implication ET}
    Let $\theta$ be the sign that resolves the eigenvector ambiguity
    \begin{align*}
        \theta = \arg \min_{\theta \in \{-1, +1\}} \lVert v(\mathbb{E}[T]) - \theta \hat{v}\rVert .
    \end{align*}
   Fix any non-negative $t \leq 1$. Algorithm \ref{alg:cluster} returns cluster membership with $\eta \leq 1- t$ when the following event on the random vector $\hat{v}$ and the random variable $\hat{\mu}$ holds true:
   \begin{equation}
       \frac{1}{d}\sum_{j=1}^d\indicator(E'_{\hat{v},j}) \geq t 
   \end{equation}
   where, 
    \begin{equation}
        E'_{\hat{v}, j} = \left\{
            \lvert v_j(\mathbb{E}[T]) - \theta \hat{v}_j \rvert 
            +\lvert \mu(\mathbb{E}[T]) - \hat{\mu}\rvert< \min \{m_e(\mathbb{E}[T]), m_h(\mathbb{E}[T])\} 
        \right\} 
    \end{equation}
\end{proposition}

The proof of the above proposition comes immediately  once we follow the proof of the proposition \ref{prop:misclustering implication} by replacing the associated spectral quantities of $n^{-1}R_y$ with that of $\mathbb{E}[T].$

Hence, using the above proposition, the tail bound on the fraction of incorrectly clustered tasks $\eta$ for any $t$ with $0 \leq t < 1$ is bounded as 
% \begin{align*}
    % \mathbb{P}\left(\eta \geq t\right) 
    % % & \leq \mathbb{P}(\text{at least $td$ tasks are incorrectly clustered} \cap E_\mu) + \mathbb{P}(E_\mu^c) \\
    % \leq \mathbb{P}(\{\eta \geq t\} \cap E_\mu) + \mathbb{P}(E_\mu^c) 
    % % \\
    % %  & 
    %  \leq \mathbb{P}\left( \left\{
    %     \frac{1}{d} \sum_j
    %  \indicator(E_{3,j}) \geq t  \right 
    %     \} \cap 
    % E_\mu \right) + \mathbb{P}(E_\mu^c)
% \end{align*}
\begin{align}
        \mathbb{P}\left(\eta > t\right) 
    % & \leq \mathbb{P}(\text{at least $td$ tasks are incorrectly clustered} \cap E_\mu) + \mathbb{P}(E_\mu^c) \\
    &
     \leq \mathbb{P}\left( 
        \frac{1}{d} \sum_j
     \indicator(E'_{\hat{v},j}) < 1- t \right)\\
     \label{eq:imperfect_cluster_general}
        & = \mathbb{P}\left( 
        \frac{1}{d} \sum_j
     \indicator({E'_{\hat{v},j}}^c) \geq t  \right).
\end{align}
Define the following event similar to the event $E_{\mu}$
\begin{equation}\label{eq:event_mu}
    E'_{\mu} = \left\{
        \lvert \hat{\mu} - \mu(\mathbb{E}[T]) \rvert < \frac{1}{2} \min \left\{ m_e(\mathbb{E}[T]), m_h(\mathbb{E}[T])\right\} 
    \right\} .
\end{equation}
We see that on the event $E'_{\mu}$, $\min \{m_e(\mathbb{E}[T]), m_h(\mathbb{E}[T])\} - \lvert \hat{\mu} - \mu(\mathbb{E}[T]) \rvert$ is non-negative. 
As a result, the intersection of events $E'_{\hat{v},j}$ and $E'_\mu$ can be re-written as
\begin{align*}
    (\theta \hat{v}_j - v_j(\mathbb{E}[T]))^2 \geq \left( \min \{m_e(\mathbb{E}[T]), m_h(\mathbb{E}[T])\} - \lvert \hat{\mu} - \mu(\mathbb{E}[T]) \rvert\right)^2  .
\end{align*}
Using this expression for the right hand side in \eqref{eq:imperfect_cluster_general} with  using the law of total probability we write,
\begin{align*}         
\mathbb{P}\left(\eta > t\right) 
& \leq \mathbb{P}\left(\left\{ 
        \frac{1}{d} \sum_j
     \indicator({E'_{\hat{v},j}}^c) \geq t \right\}\cap E'_\mu \right) + \mathbb{P}(E_\mu^c) \\
     & \leq \mathbb{P}\left( \left\{\sum_j(\theta \hat{v}_j-v_j(\mathbb{E}[T]))^2 \geq td\left(\min\{m_e(\mathbb{E}[T]),m_h(\mathbb{E}[T])\} - \abs{\hat{\mu}-\mu}\right)^2\right\} \cap E'_\mu \right) + \mathbb{P}({E'_\mu}^c) \\
     & \leq \mathbb{P}\left(\left\{\|\theta \hat{v} - v(\mathbb{E}[T])\|_2 \geq \sqrt{td} \left \lvert \min \{m_e(\mathbb{E}[T]),m_h(\mathbb{E}[T])\} - \abs{\hat{\mu}-\mu}\right\rvert \right\} \cap  E'_\mu\right)+\mathbb{P}({E'_\mu}^c)
     \\ &
     \leq \mathbb{P}\left(\left\{
        \norm{\theta \hat{v} - v(\mathbb{E}[T])}_2 \geq \sqrt{td} \frac{\min \{m_e(\mathbb{E}[T]), m_h(\mathbb{E}[T]) \}}{2} 
    \right\}\right)
     + \mathbb{P}({E'_\mu}^c).
     % \\
     % & \leq \mathbb{P}\left(\left\{\|\hat{v} - v\|_2 \geq \sqrt{td}(\min(m_e(n^{-1}R_y),m_h(n^{-1}R_y)) - \abs{\hat{\mu}-\mu}) \right\} \cap E_\mu \right)+\mathbb{P}(E_\mu^c)
\end{align*}
Next, we apply the relation \eqref{eq:mu_vs_v} to obtain an upper bound on the probability of the event ${E'_\mu}^c$,
\begin{align}\label{eq:event_mu_tail}
    \mathbb{P}(E_\mu^c) & \leq \mathbb{P}\left(
        \min_{\theta \in \{-1, +1\}} \lVert v(\mathbb{E}[T]) - \theta \hat{v}\rVert_2 \geq \sqrt{d} \frac{\min \{m_e(\mathbb{E}[T]), m_h(\mathbb{E}[T])\} }{2}
    \right)\\
    & \underbrace{\leq}_{(a)} \mathbb{P}\left(
        \min_{\theta \in \{-1, +1\}} \lVert v(\mathbb{E}[T]) - \theta \hat{v}\rVert_2 \geq \sqrt{td} \frac{\min \{m_e(\mathbb{E}[T]), m_h(\mathbb{E}[T])\} }{2}
    \right) \nonumber
\end{align}
where, $(a)$ follows due to $t \in[0,1)$.

Putting it together, after applying the $l_2$ eigenvector concentration in Lemma \ref{lemma:l_2 concentration} we arrive at the inequality:
\begin{align*}
    \mathbb{P}\left( \eta > t \right) 
    &\leq 2 \mathbb{P}\left(\min_{\theta \in \{-1, +1\}} \lVert v(\mathbb{E}[T]) - \theta \hat{v}\rVert_2 \geq \sqrt{td} \frac{\min\{m_e(\mathbb{E}[T]), m_h(\mathbb{E}[T])\}}{2}\right)
    \\
    \label{eq:temp2} & \leq 
    4d \exp \left(
        - C_6 n \frac{\left(\lambda_1 (\mathbb{E}[T]) - \lambda_2 (\mathbb{E}[T])\right)^2}{d} \min \{m_e(\mathbb{E}[T]), m_h(\mathbb{E}[T])\}^2 t      
    \right) \numberthis,
\end{align*}
where $C_6 = 2^{-8}$ .
We can simplify the above right hand side by giving a lower bound on the quantity $\min \{m_e(\mathbb{E}[T])d^{\frac{1}{2}}, m_h(\mathbb{E}[T])d^{\frac{1}{2}}$ as follows :
From the definition of $m_e(\mathbb{E}[T])$ and $m_h(\mathbb{E}[T])$ in equations \eqref{eq:m_e(ET)} and \eqref{eq:m_h(ET)},
\begin{align*}
    \min  \{m_e(\mathbb{E}[T]) d^{1/2}, m_h(\mathbb{E}[T]) d^{1/2}\}
    & = \min  \left\{\frac{d_h}{d}\frac{||\gamma|-1|}{\sqrt{d_e \gamma^2+d_h}} d^{1/2}, \frac{d_e}{d}\frac{||\gamma|-1|}{\sqrt{d_e \gamma^2+d_h}} d^{1/2}, 1\right\} \\
    & = \min \left\{\frac{\alpha ||\gamma|-1|}{\sqrt{\alpha \gamma^2+(1-\alpha)}},\frac{(1-\alpha) ||\gamma|-1|}{\sqrt{\alpha \gamma^2+(1-\alpha)}} ,1\right\}\\
    & \geq \min \left\{\alpha,1-\alpha \right\} \frac{||\gamma|-1|}{\sqrt{\gamma^2+1}}
\end{align*}
Applying the above lower bound on the equation \eqref{eq:temp2}, we get,
 \begin{equation}
        \mathbb{P}(\eta > t) \leq \begin{cases}
            4d \exp \left(
            - C_6\left(\sigma(\mathbb{E}[T]) \min \left\{\alpha,1-\alpha \right\} \frac{||\gamma|-1|}{\sqrt{\gamma^2+1}}\right)^2 nt
        \right) & \text{when } r_e^{\top}r_h \neq 0\\
        4d \exp \left(
            - C_6\left(\sigma(\mathbb{E}[T]) \min \left\{\alpha,1-\alpha \right\} \right)^2 nt
        \right) & \text{when } r_e^{\top}r_h = 0
        \end{cases} 
        % 4d\exp\left(-\tilde{C}_5nt\right)
    \end{equation}

\begin{remark}
    Recall, the perfect clustering Theorem \ref{thm:cluster_crowdsourcing} assumes the non-collinearity between $r_e$ and $r_h$, whereas, the result in Theorem \ref{thm:imperfect_cluster} works without this assumption. It is an interesting problem to find out if perfect clustering is achievable even when $r_e$ and $r_h$ are collinear.
    
    We acknowledge that the $l_2$ concentration from Lemma \ref{lemma:l_2 concentration} implies $l_\infty$ concentration in the following sense:
\begin{align*}
\mathbb{P}\left(\min_{\theta\in \{\pm 1\}} \|v(\mathbb{E}[T])-\theta \hat{v}\|_{\infty} \geq  t\right) & \leq \mathbb{P}\left(\min_{\theta\in \{\pm 1\}} \|v(\mathbb{E}[T])-\theta \hat{v}\|_{2} \geq t\right) 
    \\
    & \leq  2d\exp\left(
    % - \frac{\tilde{C}_4nt^2}{d}
    - C_6  \frac{n(\lambda_1 (\mathbb{E}[T]) - \lambda_2 (\mathbb{E}[T]))^2}{d^2} t^2
    \right) .
\end{align*}
 For an event of perfect clustering, one sufficient and straightforward condition is that $l_{\infty}$ error $\min_{\theta\in \{\pm 1\}} \|v(\mathbb{E}[T])-\theta \hat{v}\|_{\infty}$ is restricted by a order $\frac{1}{\sqrt{d}}$ quantity $\min(m_e(\mathbb{E}[T]),m_h(\mathbb{E}[T]))$ \footnote{Recall, $\mu(\mathbb{E}[T])$, $\mu_e(\mathbb{E}[T])$ and $\mu_h(\mathbb{E}[T])$ are of order $\frac{1}{\sqrt{d}}$}. But, as the eigenvalues grow linearly with $d$, the tail bound above results in a vacuous clustering performance when $d = \Omega(n)$ because of the $n/d$ exponent when $t = O\left(\frac{1}{\sqrt{d}}\right)$.
In our problem setting where $n$ workers label $d = \Omega (n)$ tasks, the above argument is insufficient to conclude that all tasks are clustered correctly for a large number of workers even when $r_e$ and $r_h$ are collinear. 
\end{remark}

\end{document}